\documentclass{article}

\PassOptionsToPackage{numbers,compress}{natbib}

\usepackage[preprint]{neurips_2026}

\usepackage[utf8]{inputenc} 
\usepackage[T1]{fontenc}    
\usepackage{hyperref}       
\usepackage{url}            
\usepackage{booktabs}       
\usepackage{graphicx}       
\usepackage{amsmath}        
\usepackage{array}          
\usepackage{subcaption}     
\usepackage{multirow}       
\usepackage{amsfonts}       
\usepackage{nicefrac}       
\usepackage{microtype}      
\usepackage{xcolor}         
\usepackage{xspace}         

\newcommand{\sys}{i\textsc{TrialSpace}\xspace}

\newcommand{\pp}{\,pp\xspace}

\definecolor{cGreen}{RGB}{34,139,34}
\definecolor{cRed}{RGB}{178,34,34}
\definecolor{cBlue}{RGB}{0,70,180}
\definecolor{cellGray}{gray}{0.93}
\newcolumntype{L}[1]{>{\raggedright\arraybackslash}p{#1}}
\newcommand{\up}[1]{\textcolor{cGreen}{{+}#1}}
\newcommand{\dn}[1]{\textcolor{cRed}{{--}#1}}

\title{\sys: Programmable Virtual Lesion Trials for Controlled Evaluation of Lung CT Models}

\author{%
  Fakrul Islam Tushar$^{1}$, Umme Hafsa Momy$^{2}$, Joseph Y. Lo$^{3}$, Geoffrey D. Rubin$^{1}$\\
  $^{1}$Department of Radiology and Imaging Sciences, University of Arizona\\
  $^{2}$Department of Biomedical Engineering, Florida International University\\
  $^{3}$Center for Virtual Imaging Trials, Department of Radiology, Duke University Medical Center
}

\begin{document}

\raggedbottom

\maketitle

\begin{center}
\small
\textbf{Hugging Face:} \href{https://huggingface.co/TusharLab/iTrialSpace}{https://huggingface.co/TusharLab/iTrialSpace}\\
\textbf{GitHub:} \href{https://github.com/tusharlabratory/itrialspace}{https://github.com/tusharlabratory/itrialspace}\\
\textbf{Dataset:} \href{https://huggingface.co/datasets/TusharLab/iTrialSpace_Lung}{https://huggingface.co/datasets/TusharLab/iTrialSpace\_Lung}
\end{center}

\begin{abstract}
We introduce \textbf{\sys{}}, a programmable evaluation framework for controlled assessment of lung CT models. Standard benchmarks are static retrospective collections that entangle lesion size, lobe prevalence, anatomy, and acquisition context, making it difficult to determine what structurally drives model accuracy. \sys{} addresses this limitation by composing real clinical CTs and lesion profiles into controlled virtual lesion trials through a four-stage pipeline: multi-dataset nodule profiling, explicit trial specification, anatomy-aware mask insertion, and ControlNet-conditioned CT synthesis. The framework is built on a unified 54-attribute nodule-profile dataset spanning 13,140 annotated nodules from seven public CT sources and instantiated as 13 trial modes.  We evaluate \sys{} in a 55,469-sample Virtual Lesion Study spanning three medical VLMs, four spatial-guidance conditions, and three clinical tasks. Across all 13 modes, the synthetic substrate remains within the real-to-real FID baseline, and synthetic performance rankings transfer strongly to real clinical data ($\rho=0.93$, $p<10^{-15}$). Controlled trial modes expose findings unavailable to fixed-distribution benchmarks, including shortcut-driven size prediction collapse under lobe-equalized sampling and host-to-donor variance ratios of $8.9\times$ and $3.3\times$ in twin-cross analysis. These results position \sys{} as an auditable evaluation infrastructure for controlled, falsifiable testing beyond static retrospective benchmarks.
\end{abstract}

\section{Introduction}
\label{sec:intro}

Evaluation of lung CT models is still dominated by fixed retrospective benchmarks~\citep{setio2017luna16,peeters2025luna25,ctbench2024,wang2025duke}. These datasets are essential for realism, but they entangle lesion size, lobe prevalence, host anatomy, and acquisition context in ways determined by data collection rather than by the evaluation question itself. As a result, standard test sets can rank models, but they often cannot explain \textbf{\emph{why}} a model succeeds or fails. In particular, they do not permit controlled tests of whether performance reflects genuine visual evidence, reliance on dataset-level priors, or sensitivity to the anatomy of the host patient.

\begin{figure}[t]
  \centering
  \includegraphics[width=\linewidth]{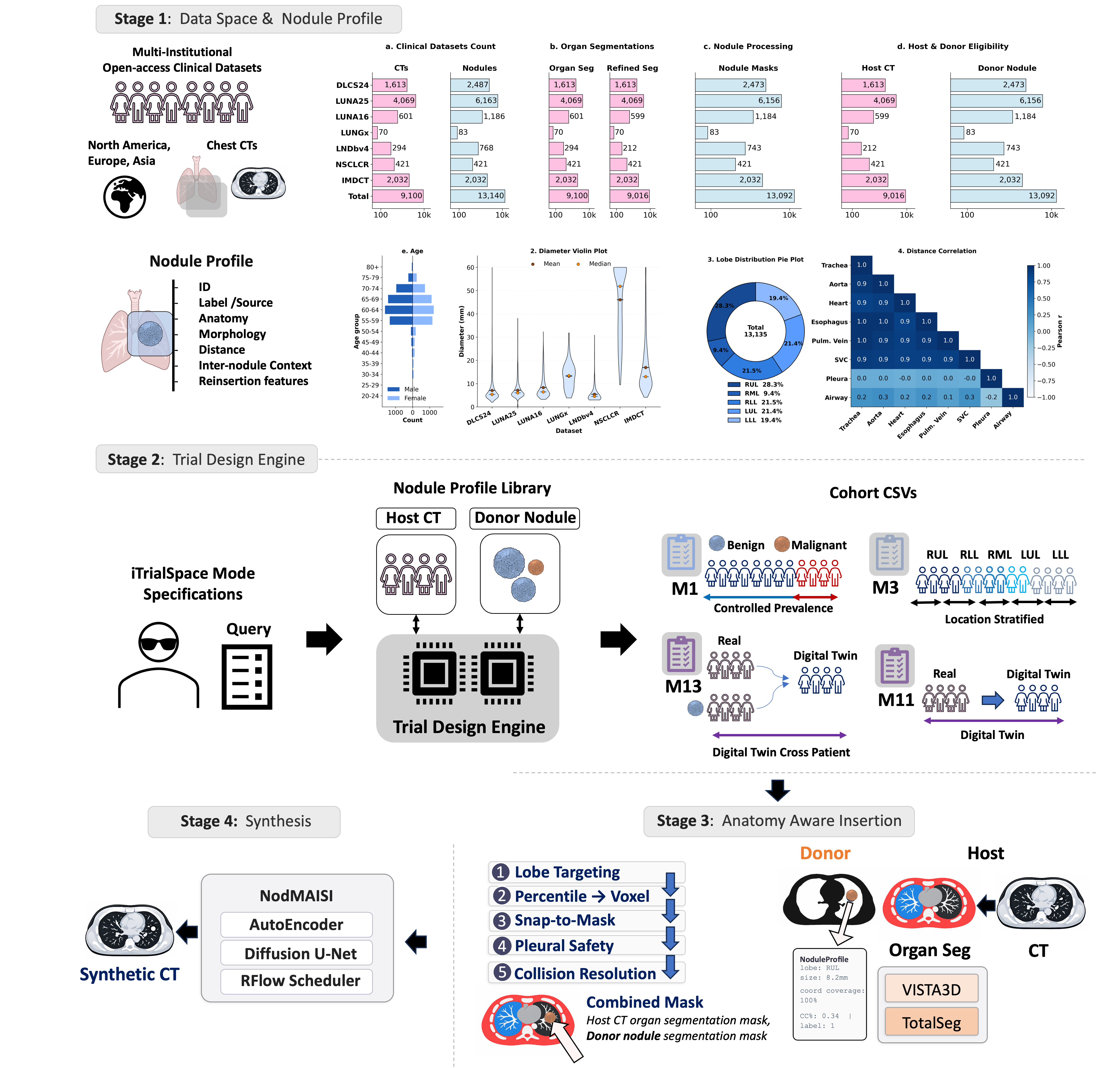}
  \vspace{-5mm}
  \caption{\textbf{\sys overview.} Real clinical CTs and anatomy segmentations are converted into structured nodule profiles (Stage~1), instantiated as explicit trial specifications and manifests (Stage~2), composed through blueprint-guided donor-to-host mask insertion (Stage~3), and synthesized into CT volumes with NodMAISI (Stage~4).}
  \label{fig:overview}
\end{figure}

Prior work addresses parts of this problem, but not the full evaluation loop. Nodule insertion~\citep{pezeshk2017poisson} and CT synthesis~\citep{chen2024maisi,tushar2025nodmaisi} methods can generate realistic synthetic data, yet they provide no explicit evaluation-design layer: they do not let the evaluator control prevalence, equalize size or lobe distributions, or systematically vary host anatomy across otherwise matched cohorts. Virtual imaging trial frameworks~\citep{badano2018victre,tushar2025vlst,tushar2026utility} demonstrate the value of controlled \emph{in-silico} experimentation, but lung CT efforts have largely operated on fully synthetic anatomy, limiting direct validation against real patient populations. What is missing is a framework that preserves real clinical CT as the evaluation substrate while making cohort construction explicit, programmable, and reproducible.

We introduce \textbf{\sys}, a programmable framework that composes real clinical CTs and lesion profiles into controlled virtual lesion trials. Rather than proposing a new benchmark, \textbf{\sys} makes evaluation cohorts explicit, reproducible, and controllable, enabling direct study of priors, anatomy, and task difficulty. Our contributions are fourfold:

\begin{itemize}
    \item \textbf{Programmable evaluation framework.}
    We introduce \textbf{\sys}, a four-stage in-silico evaluation framework that composes real clinical CT anatomy with lesion profiles to construct controlled virtual lesion trials, enabling evaluation on real patient substrates rather than fully synthetic phantoms.

    \item \textbf{Formal trial-design engine.}
    We formulate cohort construction through an explicit trial specification $\mathcal{S}$ and a deterministic Build operator that instantiate 13 trial modes, each designed to vary one evaluation factor, such as size distribution, lobe prior, prevalence, demographics, multi-nodule context, or host anatomy, while holding the remaining construction procedure fixed.

    \item \textbf{Reusable nodule-profile substrate.}
    We construct a unified 54-attribute nodule-profile representation over 13,140 annotated nodules from seven public CT sources, including precomputed reinsertion blueprints that make cohort construction deterministic and seed-reproducible across modes.

    \item \textbf{Large-scale experimental validation of evaluation utility.}
    We instantiate a 55,469-sample \textbf{Virtual Lesion Study (VLS)} spanning three medical VLMs, four spatial-guidance conditions, and three clinical tasks, and show that \textbf{\sys} produces synthetic cohorts that remain within the range of real clinical inter-dataset variability while supporting controlled evaluation beyond fixed retrospective benchmarks.
\end{itemize}

\section{Related Work}
\paragraph{Lesion insertion and CT synthesis.}
Prior work has improved the realism and controllability of synthetic lung CT data, from copy-paste lesion insertion to conditional diffusion models \citep{pezeshk2017poisson,chen2024maisi,tushar2025nodmaisi}. However, these methods do not provide an explicit trial-design layer for controlling prevalence, equalizing size or lobe distributions, or varying host anatomy across matched cohorts. In \textbf{\sys}, synthesis is used as a rendering stage within a programmable evaluation pipeline rather than as an end in itself. \textbf{Virtual imaging trials.} Virtual imaging trial frameworks have shown the value of controlled \emph{in-silico} experimentation in mammography and lung CT \citep{badano2018victre,sizikova2023msynth,tushar2025vlst,tushar2026utility,tushar2025synlungs}. Yet these systems largely rely on fully synthetic anatomy, limiting direct validation against real patient populations. By contrast, \textbf{\sys} preserves real clinical CT as the evaluation substrate while making cohort construction explicit and auditable.

\paragraph{Fixed lung CT benchmarks and medical VLM evaluation.}
Public lung CT datasets and recent CT benchmarks remain essential for realism \citep{armato2011lung,setio2017luna16,peeters2025luna25,armato2016lungx,pedrosa2021lndb,ctbench2024}, but they are static retrospective collections and therefore cannot support controlled interventions on lesion priors, host anatomy, or multi-nodule context. We evaluate \textbf{\sys} using medical VLMs from contrastive and generative families \citep{biomedclip2023,llavamed2024,medgemma2024}, but the contribution of this paper is not a new VLM; it is an evaluation framework for controlled tests unavailable to fixed-distribution benchmarks.

\section{Method}

We formulate \sys as a four-stage framework for controlled \textit{in-silico} evaluation of lung CT models (Fig.~\ref{fig:overview}). \textbf{Stage 1} constructs a structured nodule-profile space from multi-dataset clinical CTs and anatomy segmentations. \textbf{Stage 2} defines evaluation cohorts through an explicit trial specification and deterministic cohort-building operator. \textbf{Stage 3} composes donor nodules into host anatomies using anatomy-aware mask insertion. \textbf{Stage 4} synthesizes CT volumes from the composed masks using a ControlNet-conditioned model.

\subsection{Stage 1: Data Space and Nodule Profiling}
\label{sec:stage1}

\paragraph{Clinical datasets.}
\sys integrates seven publicly available CT datasets: DLCS24~\citep{wang2025duke}, LUNA25~\citep{peeters2025luna25}, LUNA16~\citep{setio2017luna16}, LUNGx~\citep{armato2016lungx}, LNDbv4~\citep{pedrosa2021lndb}, the NSCLC Radiomics (NSCLCR) dataset~\citep{aerts2014nsclcr}, and the Integrated Multiomics Dataset (IMDCT)~\citep{zhao2025integrated}. Together, they contribute \textbf{13{,}140} nodules from \textbf{7{,}069} patients across \textbf{9{,}100} CT scans, spanning screening, challenge-curated, diagnostic, and staging contexts in North America, Europe, and Asia. Of the \textbf{13{,}140} annotated nodules, \textbf{11{,}186} (85.1\%) have binary malignancy labels; the remaining cases are either unlabeled (LUNA16) or annotated only with multi-reader suspicion ratings (LNDbv4). Fig.~\ref{fig:overview} summarizes the \sys workflow and key corpus statistics, while Appendix~\ref{app:datasets} provides detailed per-dataset demographic and dataset-level summaries. \sys evaluation modes explicitly track dataset provenance via $\mathcal{D}_{\mathrm{excl}}$ and the \emph{dataset source} profile field.

\paragraph{Segmentation infrastructure.}
For each patient $p$, four segmentation assets are computed:
\textbf{(i)} Organ anatomy mask $\mathbf{m}_p^{\mathrm{anat}} \in \{0,\ldots,127\}^{H \times W \times S}$ produced by \textbf{VISTA3D}~\citep{he2024vista3d}, covering 127 anatomical structures.
\textbf{(ii)} Refined lung-lobe segmentation from \textbf{TotalSegmentator}~\citep{wasserthal2023totalseg}, which provides explicit five-lobe labels. \textbf{(iii)} A body mask obtained by thresholding at $-300$~HU, retaining the largest connected component, and applying 3D hole filling. Voxels inside the body but outside labelled organs are assigned a dedicated body label. \textbf{(iv)} Nodule masks $\{\mathbf{m}_{p,j}^{\mathrm{nod}}\}_{j=1}^{N_p}$, taken from dataset-provided annotations when available or, for unannotated cases, generated by \textbf{PiNS}~\citep{pins2025} following the NodMAISI protocol~\citep{tushar2025nodmaisi}.

\paragraph{Nodule profiling.}
Controlled cohort construction requires a reusable structured lesion representation rather than raw annotations alone. We therefore distinguish between a compact clinical descriptor and the full profile record used by the trial engine. For each nodule $(p,j)$, we define the compact feature vector as:
\begin{equation}
  \mathbf{v}_{p,j} = \bigl(d_{p,j},\; \ell_{p,j},\; y_{p,j},\; z_{p,j},\; s_{p,j},\; c_{p,j},\; \rho_{p,j}\bigr),
  \label{eq:nodule_vec}
\end{equation}
where $d \in \mathbb{R}^+$ is effective diameter (mm), $\ell \in \{\text{RUL, RML, RLL, LUL, LLL}\}$ is lung lobe (Right Upper, Middle, Lower; Left Upper, Lower), $y$ is the binary malignancy label when available, $z$ is lung zone, $s$ is laterality, $c$ is centrality (central/peripheral), and $\rho \in \mathbb{R}^+$ is pleural distance.
The full nodule profile record, denoted $P_{p,j}$, contains $\mathbf{v}_{p,j}$ together with identity, bounding-box, anatomy, positional, distance, inter-nodule, reinsertion-blueprint, and dataset-provenance fields. Across the seven source datasets this yields a unified \textbf{54-attribute Nodule Profile Dataset}. The reinsertion blueprint attributes are the key enabler of deterministic, seed-reproducible trial generation: rather than computing insertion coordinates at runtime, the trial engine reads precomputed targets directly from $P_{p,j}$. Complete schema specification is in Appendix~\ref{app:schema}.

The unified 54-attribute nodule-profile captures morphological, anatomical, spatial, and relational information for all 13{,}140 annotated nodules. Nodule volume spans 225--809{,}875\,mm$^3$ and follows the expected diameter power law ($R^2 \approx 0.85$). Nodules are predominantly peripheral (91--99\%), with NSCLCR showing the highest central fraction (8.8\%). Pleural distance is the smallest structural distance (median 5.8\,mm) and SVC distance the largest (median 172.8\,mm), while mediastinal distances are strongly correlated ($r > 0.85$). LUNA25 and LNDbv4 show the highest multi-nodule burden (4.86 and 5.21 nodules/patient), with nearest inter-nodule distances of 80--117\,mm. The 13 reinsertion attributes achieve $\geq 97\%$ coverage. Further descriptive analyses are deferred to Appendix~\ref{app:schema}.

\textbf{Donor and host eligibility criteria.} Donor eligibility required a valid nodule mask, yielding 13{,}092 eligible nodules (99.6\%), whereas host eligibility required refined organ segmentation, yielding 9{,}016 eligible CTs (99.1\%). Among host-eligible CTs, 9{,}003 also contained at least one donor-eligible nodule and therefore supported M12; Fig.~\ref{fig:overview} summarizes per-dataset availability, and additional details are deferred to Appendix~\ref{app:schema}.


\subsection{Stage 2: Trial Design Engine}
\label{sec:stage2}

Rather than treating evaluation cohorts as fixed artifacts of data collection, \sys treats them as designed objects specified before synthesis. This allows the evaluator to vary one distributional factor while holding others fixed, making comparisons auditable, repeatable, and explicitly tied to a stated evaluation question.

\paragraph{Trial specification.}
A \emph{trial specification} $\mathcal{S}$ is the fundamental unit of
cohort design:
\begin{equation}
  \mathcal{S} = \bigl(
    n,\;\pi,\;T,\;
    \phi_{\mathrm{nod}},\;\phi_{\mathrm{ins}},\;\phi_{\mathrm{demo}},\;
    \sigma,\;B,\;\mathcal{D}_{\mathrm{excl}}
  \bigr).
\end{equation}
The population fields $(n, \pi, T)$ define cohort size, malignancy
prevalence $\pi \in [0,1]$, and a clinical trial template
$T = (\pi_T, \mathbf{w}_T, \boldsymbol{\lambda}_T, \boldsymbol{\mu}_T)$
encoding the size-bucket prior $\mathbf{w}_T \in \Delta^6$, lobe prior
$\boldsymbol{\lambda}_T \in \Delta^5$, and demographic parameters from a
published screening trial (NLST~\citep{aberle2011nlst},
NELSON~\citep{de_koning2020nelson}; full parameter tables in
Appendix~\ref{app:modes}).
The filter fields $(\phi_{\mathrm{nod}}, \phi_{\mathrm{ins}},
\phi_{\mathrm{demo}})$ are predicates rather than literal sets:
$\phi_{\mathrm{nod}}(P){=}1$ denotes donor-profile eligibility,
$\phi_{\mathrm{ins}}(P,q){=}1$ denotes insertion feasibility for donor
profile $P$ in host $q$ under the geometric constraints
($\alpha_{\max}{=}1.5$, $\rho_{\min}{=}2\,$mm), and
$\phi_{\mathrm{demo}}(q){=}1$ denotes host eligibility by demographic or
dataset constraint. The distinguished predicate $\top$ means ``no additional
constraint.''
The execution fields $(\sigma, B, \mathcal{D}_{\mathrm{excl}})$ provide
the random seed for exact reproducibility, the bootstrap replicate count, and
a dataset exclusion set to prevent training-data leakage.

\paragraph{Cohort construction (\textbf{Build}).}
The operator $\mathrm{Build}: \mathcal{S} \to \mathcal{M}$ produces a
manifest $\mathcal{M}$, one row per (donor nodule, host CT, label)
triple in four steps: \textbf{Pool.} First assemble eligible donors:
        $\Omega = \{(p,j) : \mathrm{src}(P_{p,j}) \notin
        \mathcal{D}_{\mathrm{excl}},\; \phi_{\mathrm{nod}}(P_{p,j}){=}1\}$;
        for prevalence-controlled modes, partition the labeled subset into
        $\Omega_{\mathrm{mal}}$ and $\Omega_{\mathrm{ben}}$ according to the
        binary label in $\mathbf{v}_{p,j}$.
\textbf{Sample.} Second draw $\lfloor n\pi \rfloor$ donors from
        $\Omega_{\mathrm{mal}}$ and $n{-}\lfloor n\pi \rfloor$ from
        $\Omega_{\mathrm{ben}}$, weighted by template priors
        $(\mathbf{w}_T, \boldsymbol{\lambda}_T)$ under seed $\sigma$.
\textbf{Host assignment.} Third for each sampled $(p,j)$, select host
        $q \neq p$ satisfying $\phi_{\mathrm{demo}}(q){=}1$ and
        $\phi_{\mathrm{ins}}(P_{p,j},q){=}1$.
\textbf{Placement.} Fourth read the nodule's pre--computed reinsertion
        blueprint from $P_{p,j}$ to obtain target coordinates
        $(i,j,k)$ in host $q$'s anatomy. Fixing $\sigma$ makes Build fully deterministic: the same $\mathcal{S}$
always produces an identical manifest, isolating any downstream variability
to the stochastic Stage~4 synthesis.

\paragraph{Mode families.}
All 13 trial modes share the same trial specification $\mathcal{S}$ and $\mathrm{Build}$ operator, differing only in which component of $\mathcal{S}$ is varied. We group them into two families. Synthetic cohort modes (M1--M10) compose donor nodules and host anatomies from different patients ($q \neq p$) to isolate one factor at a time: baseline prevalence control (M1), size-stratified cohorts (M2), lobe-specific
location isolation under matched size constraints (M3), demographic stratification (M4),
counterfactual prevalence sweeps with shared host pools (M5), cross-dataset transfer by donor-source
restriction (M6), bootstrap resampling for confidence intervals (M7), fixed-manifest algorithm
comparison (M8), multi-round screening simulation with decaying prevalence (M9), and single-
versus multi-nodule context (M10). \textbf{Digital twin modes (M11--M13)} construct patient-specific scenarios by re-inserting lesions into native anatomy or transplanting them across hosts. \textbf{M11} isolates lesion-level readout in native context, \textbf{M12} reconstructs the full native lesion configuration, and \textbf{M13} disentangles lesion effects from host anatomy through cross-patient transplantation. Thus, a trial mode is not a separate pipeline, but a controlled intervention on the same specification space. Full mode definitions are provided in Appendix~\ref{app:modes_defi}.

\subsection{Stage 3: Anatomy-Aware Mask Insertion}
\label{sec:stage3}
Anatomy-aware insertion ensures that donor nodules are placed into host scans in a way that respects lobe identity, local parenchymal boundaries, and pleural safety constraints. Given a manifest row pairing donor nodule $(p,j)$ with host $q$, insertion
proceeds in five constrained steps.
\textbf{(1)~Lobe targeting:} the donor's reinsertion blueprint specifies the
target lobe in host $q$'s TotalSegmentator lobe mask.
\textbf{(2)~Voxel resampling:} the donor mask is resampled to host spacing
with isotropic scale $\alpha \in [\alpha_{\min}, \alpha_{\max}]$ where
required.
\textbf{(3)~Blueprint placement:} the reinsertion percentile blueprint maps
to candidate coordinates $(x, y, z)$ within the target lobe.
\textbf{(4)~Snap correction:} violations of lung-parenchyma boundaries
trigger a shift to the nearest valid position.
\textbf{(5)~Validation:} pleural distance $\rho \ge \rho_{\min}$ and overlap
constraints are enforced; failures trigger re-placement or case rejection.
Pipeline Stages~1--3 are fully deterministic given $\sigma$: re-running the
pipeline with the same specification $\mathcal{S}$ reproduces the identical
manifest and composed mask $\hat{\mathbf{m}}$, isolating any downstream
variability to the stochastic Stage~4 synthesis.
Detailed placement statistics are in Appendix~\ref{app:generation}.

\subsection{Stage 4: ControlNet-Conditioned CT Synthesis}
\label{sec:stage4} 
For CT synthesis, we adopt \textbf{NodMAISI}~\citep{tushar2025nodmaisi}. Each accepted insertion mask $\mathbf{m}^{\mathrm{ins}}$ is passed to
\textbf{NodMAISI}, a ControlNet-conditioned rectified flow model to generate synthetic CT volume
$\hat{\mathbf{x}} \in \mathbb{R}^{H \times W \times S}$ conditioned on the
full anatomy+nodule mask:
\begin{equation}
  \mathrm{NodMAISI}: \mathbf{m}^{\mathrm{ins}} \longrightarrow \hat{\mathbf{x}},
\end{equation}

NodMAISI was fine-tuned for lung nodule synthesis on the clinical datasets ~\citep{tushar2025nodmaisi}.
All distributional-fidelity and downstream demonstrations in this paper are
therefore \emph{generator-conditional}: they characterise the
\sys\,+\,NodMAISI release as a coupled system, and properties of the
synthetic substrate may differ under alternative conditioned generators.
Automated quality control extracts axial, coronal, and sagittal slices
through the nodule centroid for downstream evaluation.
Generation statistics, quality metrics, and failure analysis are reported in
Appendices~\ref{app:quality} and \ref{app:generation}.

Taken together, these four stages transform real clinical CTs, lesion profiles, and controlled insertion policies into programmable evaluation cohorts within \sys.


\section{Experiments and Results}
\label{sec:experiments}
\begin{figure}[t]
  \centering
  \includegraphics[width=\linewidth]{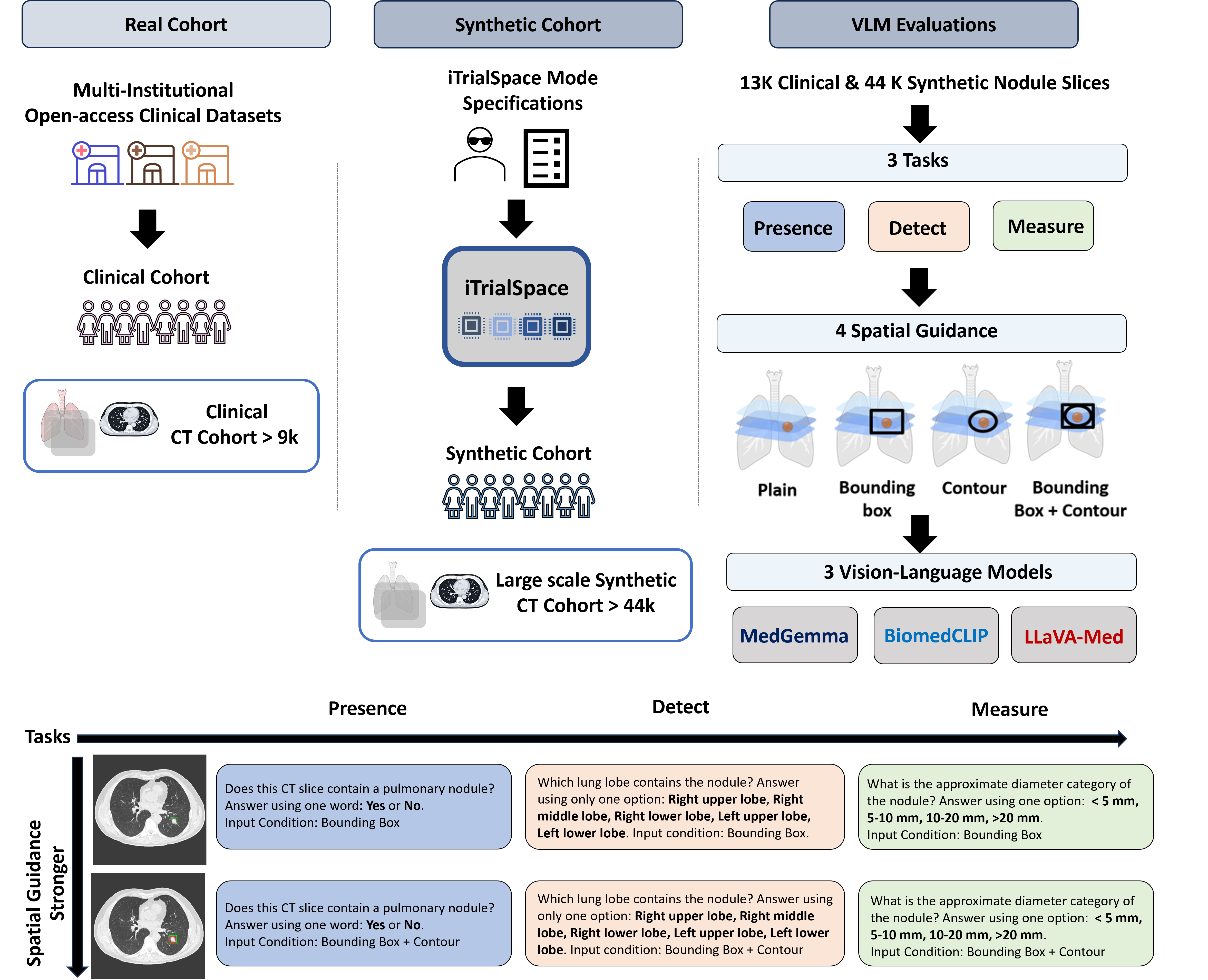}
  \vspace{-5mm}
\caption{\textbf{Virtual Lesion Study (VLS) overview.}
\sys-generated synthetic CTs are combined with real clinical CTs to form the evaluation corpus used in the VLS. The study compares three medical VLMs across three clinically motivated tasks, presence detection, lobe localisation, and size classification, and four spatial-guidance conditions: plain, bounding box, contour, and bounding box+contour. Representative prompts illustrate the controlled visual-guidance setup used for evaluation.}
  \label{fig:vls_overview}
\end{figure}

To evaluate whether \textbf{\sys} functions as a useful evaluation substrate rather than merely a synthetic-data pipeline, we conducted a Virtual Lesion Study (VLS) using \textbf{\sys} (Fig.~\ref{fig:vls_overview}). The VLS examines four questions: (i) whether the framework operates at practical scale and reproducibly constructs multi-mode cohorts, (ii) whether the synthetic substrate remains within the range of natural clinical inter-dataset variability, (iii) whether controlled cohorts expose architecture-specific behaviour under spatial guidance, and (iv) whether conclusions drawn on synthetic cohorts transfer to real clinical data.

\subsection{Virtual Lesion Study Setup}
\label{sec:vls_setup}
The VLS evaluates three medical VLMs spanning complementary architectural families: \textbf{BiomedCLIP}~\citep{biomedclip2023}
(${\sim}$400\,M parameters, contrastive image--text alignment),
\textbf{LLaVA-Med}~\citep{llavamed2024} (7\,B, generative instruction-tuned),
and \textbf{MedGemma}~\citep{medgemma2024} (4\,B, generative multi-slice). Each model is tested on three clinically motivated tasks: \textbf{presence detection, lobe localization}, and \textbf{size classification}; under four spatial-guidance conditions: \textbf{plain, bounding box, contour}, and \textbf{bounding box+contour}. The final evaluation corpus contains \textbf{55{,}469} nodule samples drawn from \textbf{51{,}385} CT volumes, comprising \textbf{42{,}382} synthetic samples across 13 controlled trial modes and \textbf{13{,}087} real samples from seven clinical datasets, for a total of \textbf{1{,}996{,}884} inference calls.

\subsection{Corpus construction and synthetic--substrate validity}

Across 13 modes, \sys{} generated 45{,}024 trial specifications, 49{,}024 insertions, and 44{,}176 synthetic CT volumes, demonstrating practical multi-mode scale. The standard synthetic-cohort modes achieved near-perfect reliability overall; across the full 13-mode study, insertion-stage attrition was concentrated in M13, while synthesis-stage attrition was dominated by M11, with additional non-trivial losses in M13 and only minor losses in M10. Table~\ref{tab:mode_summary} and Appendices~\ref{app:modes} and~\ref{app:generation} provide details. Because cohort construction is deterministic given the declared specification and seed, the same specification yields the same manifest on re-run.

\paragraph{Synthetic Validity.}
Scale alone is insufficient; the synthetic substrate must also remain within the range of natural clinical variability. We therefore evaluate synthetic fidelity against a real-to-real (R2R) baseline rather than in isolation. Across 15 pairwise real--dataset comparisons, the R2R Fréchet Inception Distance (FID)~\citep{Heusel2017GANs} baseline has median \textbf{1.57} and interquartile range \textbf{[1.05, 2.72]}. All 13 \sys modes fall within this interval, with FID$_{\mathrm{avg}}$ ranging from \textbf{1.29} to \textbf{2.22}. Mode ordering by FID is consistent with cohort-level histogram similarity, indicating agreement between learned feature-space and intensity-distribution metrics (Appendix~\ref{app:quality}, Table~\ref{tab:quality_full}). 

\paragraph{Intensity fidelity.}
Per-case intensity fidelity supports the same conclusion. Histogram intersection across 607 synthetic--host pairs ranges from 0.816 to 0.847 (grand mean \textbf{0.838}), remaining above \textbf{0.80} for all modes. Overall, these results indicate that \sys produces synthetic cohorts that are not identical to real data, but remain within the natural span of inter-dataset variability and are suitable as an evaluation substrate (Appendix~\ref{app:quality}).

\begin{table*}[t]
\caption{%
  \textbf{Pipeline summary across all 13 trial modes.}
  \emph{Key formulation}: the distinguishing parameter or operator for each
  mode; all modes share the Build operator defined in \S\ref{sec:stage2}.
  \emph{Insertions} and \emph{Synthetic CTs}: counts generated under each
  specification. FID$_\text{avg}$ denotes FID averaged over the XY, YZ, and ZX
  planes (2.5D, 1\,mm$^3$, 100 volumes vs.\ LUNA25). Full quality metrics are
  reported in Appendix~\ref{app:quality} (Table~\ref{tab:quality_full}).}
\label{tab:mode_summary}
\centering\footnotesize
\setlength{\tabcolsep}{1.5pt}
\begin{tabular}{@{}ccp{4.8cm}crcc@{}}
\toprule
\textbf{Mode} & \textbf{Sub-cohorts}
  & \textbf{Key Formulation}
  & \textbf{Trial Specs}
  & \textbf{Insertions} & \textbf{Synthetic CTs} & \textbf{FID$_\text{avg}$}$\downarrow$ \\
\midrule
M1  &  1 & NLST prior ($\pi = \pi_0$)
  &  1{,}000 &  1{,}000 &    999  & 1.77 \\
M2  &  6 & Size bin $s_i$, $i=1,\ldots,6$
  &    600   &    600   &    600  & 2.03 \\
M3  &  5 & Lobe $\ell$; $d \in [d_{\min}, d_{\max}]$
  &    500   &    500   &    500  & 2.07 \\
M4  &  4 & Sex $\times$ age stratum
  &    800   &    800  &    799  & 1.68 \\
M5  &  5 & Prevalence $\pi$ (anatomy fixed via $\sigma$)
  &  2{,}500 &  2{,}500 &  2{,}496 & 1.60\\
M6  &  5 & Leave-one-dataset-out $D_k$
  &  1{,}500 &  1{,}500 &  1{,}499 & 2.13\\
M7  & 20 & Bootstrap seed $b = 1,\ldots,20$
  &  4{,}000 &  4{,}000 &  3{,}984 & 1.31\\
M8  &  1 & Fixed seed; shared exclusion set
  &    500   &    500   &    499  & 1.29 \\
M9  &  3 & Prevalence decay $\pi_r = \pi_0\gamma^r$
  &  1{,}500 &  1{,}500 &  1{,}495 & 1.66\\
M10 &  2 & Mix $\alpha$ single / $(1{-}\alpha)$ multi
  &    531   &    530    &    498  & 1.96 \\
\cmidrule{1-7}
M11 &  7 & Twin isolate ($q = p$)
  & 13{,}092 & 13{,}087 & 12{,}448 & 1.74\\
M12 &  7 & Twin complete ($q = p$, all $N_p$)
  &  9{,}003$^*$ & 13{,}087 &  8{,}981 & 1.82\\
M13 & 42 & Twin cross ($q \neq p$)
  &  9{,}498 &  9{,}420  &  9{,}378 & 2.22\\
\cmidrule{1-7}
    \textbf{Total} &  &
  & \textbf{45{,}024} & \textbf{49{,}024} & \textbf{44{,}176} & \textbf{1.57$^{\dagger}$ [1.05, 2.72]} \\
\bottomrule
\end{tabular}\\[2pt]
{\footnotesize $^*$M12 specifications reflect the subset of 9,016 host-eligible
patient CTs that also contain at least one donor-eligible nodule; this yields 9,003 CTs
embedding 13,092 nodules simultaneously. $^{\dagger}$R2R: real-to-real baseline across 15 cross-dataset pairs (FID
median = 1.57, IQR [1.05, 2.72]; HI = 0.927 for LUNA25 $\leftrightarrow$ DLCS24).}
\end{table*}


\subsection{Controlled VLM behavior under spatial guidance}
\label{sec:vlm_eval}

\begin{table*}[t]
\caption{\textbf{VLM results on real and synthetic data.}
Accuracy (\%) under plain input and best spatial guidance condition
(in parentheses: B=BBox, C=Contour, B+C=Both). $\Delta$ denotes guidance lift
(best guided $-$ plain). Real: $n{=}13{,}087$. Synthetic: $n{=}42{,}382$ across 13 trial modes.
Chance levels are 50\% for Presence, 20\% for Lobe, and 25\% for Size.
$^\ast$Degenerate: predicts \emph{present} for $>$99.9\% of cases regardless of guidance.}
\label{tab:vlm_real_syn}
\centering\small
\setlength{\tabcolsep}{4pt}
\begin{tabular}{@{}ll ccc ccc ccc@{}}
\toprule
\multirow{2}{*}{\textbf{Source}} & \multirow{2}{*}{\textbf{Model}} 
& \multicolumn{3}{c}{\textbf{Presence Detection}} 
& \multicolumn{3}{c}{\textbf{Lobe Localisation}} 
& \multicolumn{3}{c}{\textbf{Size Classification}} \\
\cmidrule(lr){3-5}\cmidrule(lr){6-8}\cmidrule(lr){9-11}
& & Plain & Best & $\Delta$ & Plain & Best & $\Delta$ & Plain & Best & $\Delta$ \\
\midrule
\multirow{3}{*}{Real}
& BiomedCLIP & 13.0 & 44.0\,(C)    & \up{31.0} & 52.2 & 78.5\,(C)   & \up{26.3} & 25.7 & 27.2\,(B+C) & \up{1.5}  \\
& LLaVA-Med  & 100$^\ast$ & 100$^\ast$\,(B) & 0.0  & 22.1 & 11.4\,(B)   & \dn{10.7} & 24.9 & 25.0\,(B+C) & \up{0.1}  \\
& MedGemma   & 22.2 & 94.8\,(B+C)  & \up{72.6} & 44.4 & 53.0\,(B)   & \up{8.6}  & 43.0 & 53.3\,(B+C) & \up{10.3} \\
\midrule
\multirow{3}{*}{Synthetic}
& BiomedCLIP & 17.1 & 63.0\,(C)    & \up{45.9} & 38.0 & 67.4\,(C)   & \up{29.4} & 28.6 & 31.1\,(B+C) & \up{2.5}  \\
& LLaVA-Med  & 99.9$^\ast$ & 100.0$^\ast$\,(B) & 0.0 & 20.4 & 8.2\,(B) & \dn{12.2} & 28.0 & 28.5\,(B+C) & \up{0.5}  \\
& MedGemma   & 17.9 & 88.6\,(B)    & \up{70.7} & 47.4 & 62.7\,(B)   & \up{15.3} & 41.1 & 47.0\,(B+C) & \up{5.9}  \\
\bottomrule
\end{tabular}
\end{table*}

\begin{figure}[t]
  \centering
  \includegraphics[width=\linewidth]{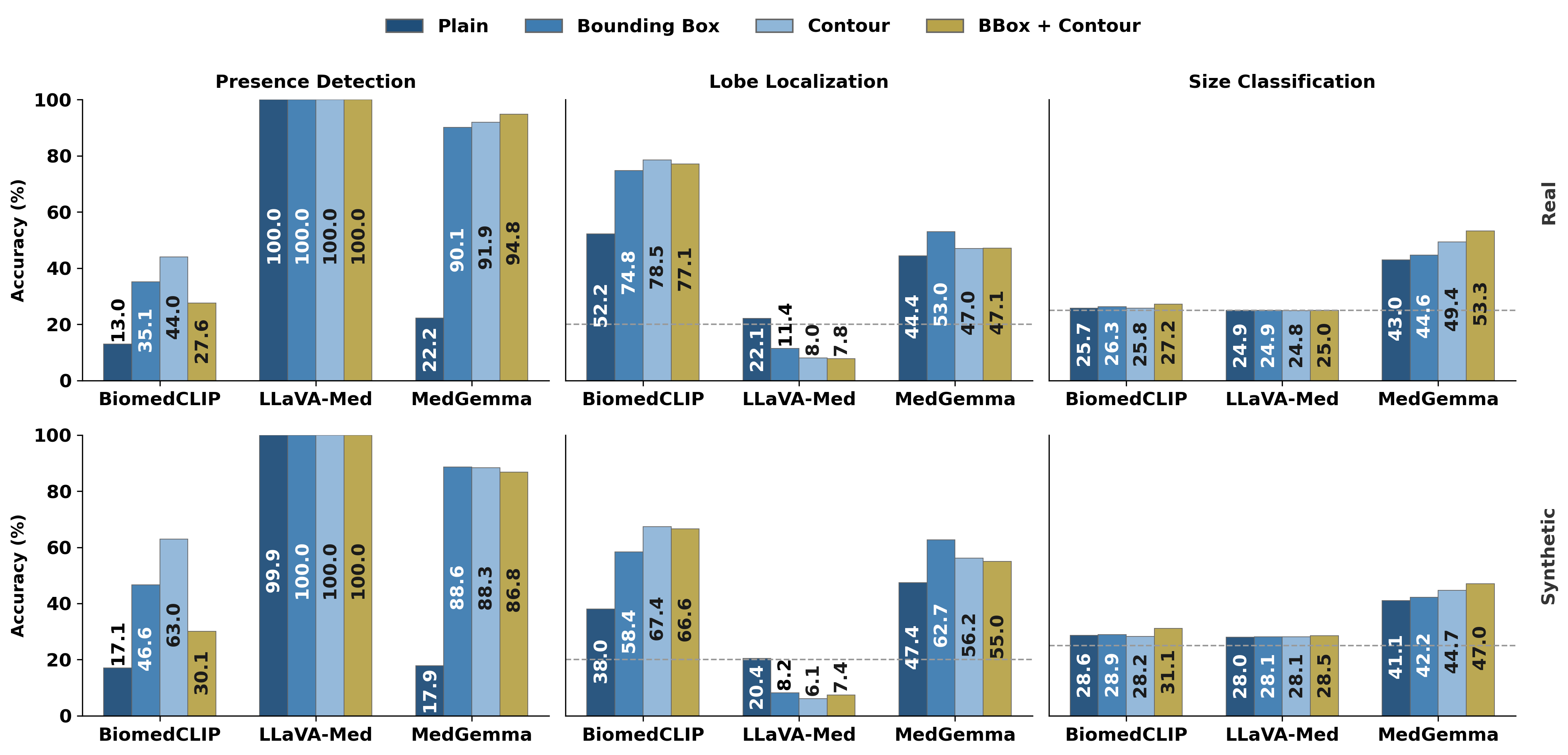}
  \vspace{-5mm}
  \caption{\textbf{Spatial-guidance effects across models, tasks, and domains.} Real and synthetic accuracies for BiomedCLIP, LLaVA-Med, and MedGemma under plain input, bounding-box guidance, contour guidance, and combined guidance. Guidance substantially improves presence detection and lobe localisation for BiomedCLIP and MedGemma, whereas size classification remains the most resistant task across domains.}
  \label{fig:vlm_overview}
\end{figure}

We next ask whether controlled cohorts expose architecture-specific dependencies on spatial guidance, and whether those dependencies are preserved across real and synthetic data. Table~\ref{tab:vlm_real_syn} and Fig.~\ref{fig:vlm_overview} show three findings.

\textbf{Spatial guidance is transformative but architecture-dependent.} MedGemma’s presence accuracy rises from 22.2\% to 94.8\% on real data and from 17.9\% to 88.6\% on synthetic data, whereas BiomedCLIP improves most strongly under contour guidance and LLaVA-Med degrades on lobe localisation under all guided conditions.

\textbf{Guidance preference is architecture-specific.} BiomedCLIP responds best to contour for presence and lobe across both domains; MedGemma prefers BBox for lobe in both domains, but presence splits by domain, with BBox+Contour best on real data and BBox best on synthetic data; LLaVA-Med’s size predictions remain effectively unchanged under all conditions, consistent with instruction-tuning priors overriding visual evidence.

\textbf{Size classification remains the hardest task.} No model exceeds 53.3\% on 4-class size prediction in any setting, and guidance lifts are modest relative to the large gains observed for presence and lobe. This suggests that boundary overlays help localize lesions, but do not by themselves resolve the harder representation problem required for robust size estimation.

\subsection{Synthetic--Real Ranking Preservation}
\label{sec:synth_real}
For \sys to be useful as an evaluation substrate, synthetic cohorts do not need to reproduce real accuracies exactly; instead, they must preserve the relative conclusions that matter for evaluation and model selection. On this criterion, transfer is strong. Across all 36 model$\times$task$\times$condition cells, synthetic and real accuracies are strongly correlated (Spearman $\rho = 0.93$, $p < 10^{-15}$). Appendix~\ref{app:domain_gap} reports the full comparison. In addition, among the eight non-degenerate model--task cases where a best guidance condition is meaningfully defined, synthetic and real data agree on the best condition in seven (Table~\ref{tab:vlm_real_syn}). Absolute offsets remain model- and task-specific rather than uniformly optimistic. BiomedCLIP lobe accuracy is consistently higher on real data across all four conditions, whereas MedGemma lobe accuracy is consistently lower. LLaVA-Med’s degenerate present-always behaviour is nearly identical across domains (100.0\% on real; 99.9--100.0\% on synthetic), confirming it as a transferable failure mode rather than a synthetic artefact.

\subsection{Controlled Trial Modes Expose Prior-Driven Behaviour}
\label{sec:mode_ablation}

\begin{figure}[t]
  \centering
  \includegraphics[width=1.0\linewidth]{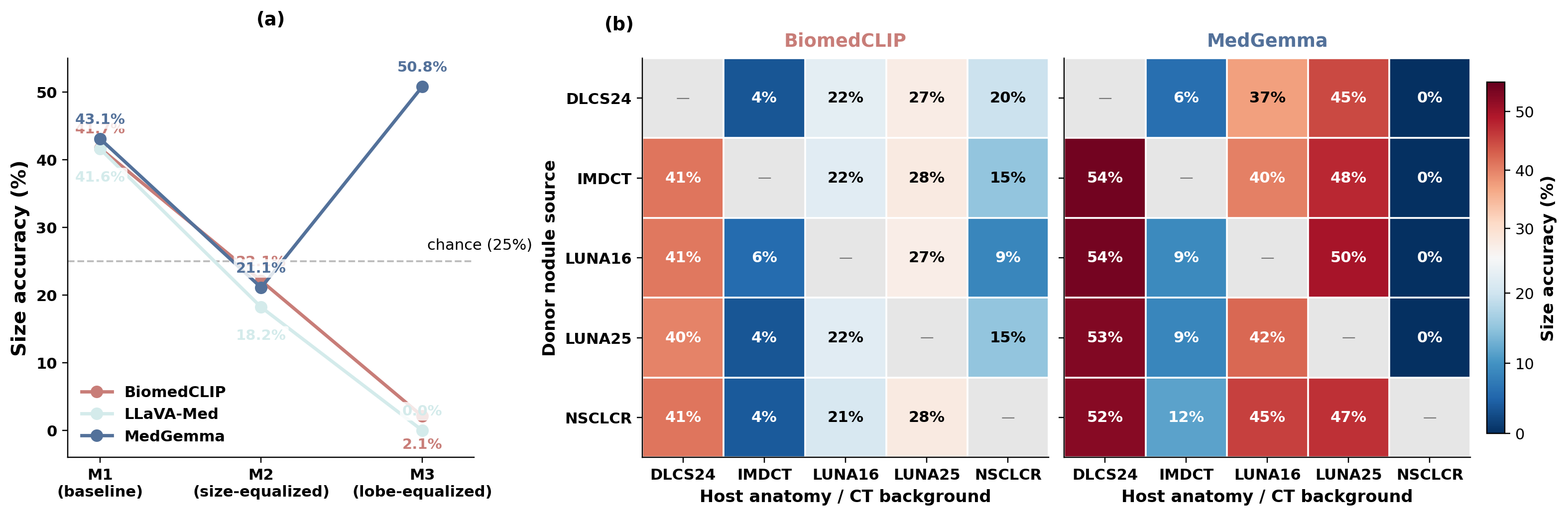}
\caption{\textbf{Controlled interventions reveal shortcut and host effects in size classification.}
\textbf{(a)} Size accuracy across \textbf{M1}--\textbf{M3}; the dashed line marks 25\% chance. \textbf{BiomedCLIP} and \textbf{LLaVA-Med} collapse to near chance by \textbf{M3}, while \textbf{MedGemma} remains above chance.
\textbf{(b)} \textbf{M13} host--donor transfer matrices. Accuracy varies more across host anatomy than donor source.}
  \label{fig:shortcut_ablation}
\end{figure}

Equalising the size distribution across six bins (\textbf{M2}) collapses BiomedCLIP and MedGemma size accuracy from 41.3\% and 51.5\% (M1, plain) to 20.3\% and 20.8\%, at or below the 25\% chance level regardless of guidance condition (means: 22.1\% and 21.1\%; Table~\ref{tab:per_mode_plain}). This indicates that above-chance size performance under naturalistic cohorts is largely size-prior-driven rather than evidence-driven. Equalising lobe prevalence while fixing diameter range (\textbf{M3}) further exposes an architectural confound: BiomedCLIP size accuracy falls to \textbf{0.6\%} on plain images (mean: 2.1\%), below chance and effectively non-functional, while MedGemma retains 30.6\% plain (mean: 50.8\%), as shown in Fig.~\ref{fig:shortcut_ablation}a. The dissociation indicates that BiomedCLIP's size estimate is confounded by lobe location; removing the natural lobe--size correlation renders its predictions systematically wrong rather than merely random. Per-size-bin accuracy and full mode-by-mode statistics are in Appendix~\ref{app:mode_ablation}.

\subsection{Host-Anatomy Disentanglement}
\label{sec:m13}

\sys's twin-cross mode \textbf{(M13)} transplants each donor nodule into multiple host anatomies under controlled pairing, enabling a variance decomposition that fixed-host benchmarks cannot perform. For size classification, the accuracy spread attributable to host anatomy is \textbf{29.3\,pp} for BiomedCLIP and \textbf{13.4\,pp} for MedGemma, compared with donor-attributable spreads of only 3.3 pp and 4.0 pp, yielding host-to-donor ratios of \textbf{8.9$\times$} and \textbf{3.3$\times$} (Fig.~\ref{fig:shortcut_ablation}b). The effect is not uniform across hosts. BiomedCLIP size accuracy collapses to 4--6\% when IMDCT anatomy is the host regardless of donor origin, while the same nodules achieve 27--41\% in DLCS24 anatomy. MedGemma shows an analogous sensitivity: size accuracy falls to 0\% across all donors when NSCLCR anatomy is the host. These results show that host anatomy can matter more than the donor nodule itself, a confound that is structurally invisible to evaluation designs that hold host anatomy fixed.

Taken together, these results support iTRIALSPACE as an evaluation substrate: synthetic cohorts preserve the evaluation conclusions that matter without requiring large labelled real datasets.

\section{Limitations}
\label{sec:limitations}
Several limitations bound the present findings.
\textbf{Generator dependence.} Results are conditional on NodMAISI as the CT synthesis backbone; alternative generators may change absolute fidelity and accuracy.
\textbf{VLM coverage.} Only three medical VLM families are evaluated, so observed architectural patterns may not generalize to other models or fine-tuning regimes.
\textbf{Lesion and modality scope.} The pipeline is optimized for pulmonary nodules in CT; extending it to other lesion types or imaging modalities would require modality-specific placement constraints and generators.
\textbf{Task scope.} We study presence, lobe localisation, and size classification, but not higher-level tasks such as malignancy risk, growth, or texture characterization.
\textbf{Segmentation dependence.} Stage~3 depends on lobe masks, so segmentation errors can propagate to insertion position and bias anatomically atypical cases.
\section{Conclusion}
\label{sec:conclusion}

We presented \sys{}, a programmable evaluation framework for lung CT models that composes real clinical CTs, lesion profiles, and controlled insertion policies into auditable cohorts. Across a large-scale Virtual Lesion Study, \sys{} exposes shortcut-driven behaviour, separates host-anatomy effects from lesion effects, and preserves the relative conclusions needed for model evaluation. More broadly, it shifts benchmark design from fixed retrospective datasets toward controllable evaluation cohorts.

\section*{Data and Code Availability}
The project resources released publicly. The Hugging Face repository is available at \href{https://huggingface.co/TusharLab/iTrialSpace}{https://huggingface.co/TusharLab/iTrialSpace}. The source code released on GitHub at \href{https://github.com/tusharlabratory/itrialspace}{https://github.com/tusharlabratory/itrialspace}. The dataset is currently available at \href{https://huggingface.co/datasets/TusharLab/iTrialSpace_Lung}{https://huggingface.co/datasets/TusharLab/iTrialSpace\_Lung}.

\section*{Acknowledgements}
\sys{} builds on many open research efforts. We thank the creators and maintainers of the public CT datasets used in this work, including DLCS24, LUNA25, LUNA16/LIDC-IDRI, LUNGx, LNDb, NSCLC-Radiomics, and IMDCT, several of which are distributed through TCIA. We also acknowledge the model and tool ecosystems that support the pipeline, including NodMAISI, MAISI/Project-MONAI, Hugging Face diffusers, VISTA3D, TotalSegmentator, PiNS, RadImageNet, BiomedCLIP, LLaVA-Med, and MedGemma. This work was supported by startup funding from the Department of Radiology and Imaging Sciences, University of Arizona, with initial computational support from the Center for Virtual Imaging Trials, Duke University. Any errors are our own, and acknowledgement does not imply endorsement by the cited projects or their authors.


\newpage

\appendix
\setcounter{figure}{0}
\setcounter{table}{0}
\renewcommand{\thefigure}{A\arabic{figure}}
\renewcommand{\thetable}{A\arabic{table}}

\section*{Appendix Contents}
\thispagestyle{plain}

\noindent Appendix~\ref{app:datasets}: Detailed Dataset Summary \dotfill \pageref{app:datasets}

\medskip
\noindent Appendix~\ref{app:schema}: Nodule Profile Dataset: Full 54-Attribute Schema \dotfill \pageref{app:schema}

\medskip
\noindent Appendix~\ref{app:modes_defi}: Mode Definitions \dotfill \pageref{app:modes_defi}

\medskip
\noindent Appendix~\ref{app:modes}: Virtual Lesion Study (VLS): Trial Mode Specification \dotfill \pageref{app:modes}

\medskip
\noindent Appendix~\ref{app:generation}: Generation Pipeline: Scale, Throughput, and Failure Analysis \dotfill \pageref{app:generation}

\medskip
\noindent Appendix~\ref{app:quality}: Synthetic CT Quality Assessment \dotfill \pageref{app:quality}

\medskip
\noindent Appendix~\ref{app:vlm}: VLM Evaluation \dotfill \pageref{app:vlm}

\newpage

\section{Detailed Dataset Summary}
\label{app:datasets}

The seven clinical datasets cover complementary clinical regimes. DLCS24 contributes a screening-oriented U.S. cohort with full demographic metadata and clinical management variables; LUNA25 is the largest source, with 6{,}163 nodules from 2{,}120 patients across 4{,}069 CT scans; LUNA16 contributes 1{,}186 nodules with morphology and location annotations but no malignancy labels; LUNGx is a small challenge cohort with 83 nodules and a near-balanced benign--malignant split (49.4\% malignant); LNDbv4 contributes 768 nodules from 212 patients across 294 CT scans with multi-reader malignancy ratings; NSCLCR contains 421 surgically confirmed malignant nodules; and IMDCT contributes 2{,}032 nodules from a diagnostic/staging context with 78.6\% malignancy.

Demographic information was available for 4 of the 7 datasets (DLCS24, LUNA25, NSCLCR, and IMDCT), and the per-dataset summary is provided in Table~\ref{tab:app_datasets}. Race and ethnicity metadata were available only for DLCS24 and LUNA25: DLCS24 comprised 74.1\% White, 22.7\% Black/African American, and 3.2\% Other/Unknown participants, with 96.4\% non-Hispanic, 0.4\% Hispanic, and 3.2\% unavailable ethnicity; LUNA25 comprised 93.2\% White, 3.6\% Black/African American, and 3.3\% Other/Unknown participants, with 97.7\% non-Hispanic, 2.0\% Hispanic, and 0.7\% unavailable ethnicity.

\begin{table}[htbp]
\caption{Dataset summary for the seven CT sources used in the Nodule Profile Dataset. Malignant percentages are computed over labelled nodules only.}
\label{tab:app_datasets}
\centering\footnotesize
\setlength{\tabcolsep}{3pt}
\resizebox{\linewidth}{!}{%
\begin{tabular}{@{}llrrrrrl@{}}
\toprule
\textbf{Dataset} & \textbf{Origin} & \textbf{Patients} & \textbf{Female (\%)} & \textbf{CT Scans} & \textbf{Nodules} & \textbf{Malignant (\%)} & \textbf{Labels}\\
\midrule
DLCS24~\citep{wang2025duke} & USA & 1{,}613 & 802 (49.7) & 1{,}613 & 2{,}487 & 264 (10.6\%) & Binary\\
LUNA25~\citep{peeters2025luna25} & USA & 2{,}120 & 909 (42.9) & 4{,}069 & 6{,}163 & 555 (9.0\%) & Binary\\
LUNA16~\citep{setio2017luna16} & USA & 601 & --- & 601 & 1{,}186 & --- & None\\
LUNGx~\citep{armato2016lungx} & USA & 70 & --- & 70 & 83 & 41 (49.4\%) & Binary\\
LNDbv4~\citep{pedrosa2021lndb} & Portugal & 212 & --- & 294 & 768 & --- & Rating$^\dag$\\
NSCLCR~\citep{aerts2014nsclcr} & USA & 421 & 131 (31.1) & 421 & 421 & 421 (100.0\%) & All mal.\\
IMDCT~\citep{zhao2025integrated} & China & 2{,}032 & 1{,}216 (59.8) & 2{,}032 & 2{,}032 & 1{,}598 (78.6\%) & Binary\\
\midrule
\textbf{Total} & --- & \textbf{7{,}069} & --- & \textbf{9{,}100} & \textbf{13{,}140} & \textbf{2{,}879 (25.7\%)}$^\ddagger$ & ---\\
\bottomrule
\end{tabular}}\\[2pt]
{\footnotesize $^\dag$LNDbv4: Multi-radiologist ratings (1--5 scale); threshold $\geq$3.0 yields 253 malignant and 515 benign nodules, but these are excluded from the binary counts above because the source labels are soft ratings. $^\ddagger$Malignant percentage computed over 11{,}186 labelled nodules (excludes LUNA16 and LNDbv4).}
\end{table}

Table~\ref{tab:app_nodules_per_patient} shows the distribution of nodule count per patient across datasets. NSCLCR and IMDCT contain exactly one annotated nodule per patient. LUNA25 and LNDbv4 show the highest multi-nodule burden (52.7\% and 53.3\% with $\geq$3 nodules).
\begin{table}[htbp]
\caption{Distribution of nodules per patient across datasets.}
\label{tab:app_nodules_per_patient}
\centering\footnotesize
\setlength{\tabcolsep}{4pt}
\resizebox{\linewidth}{!}{%
\begin{tabular}{@{}lcccccrr@{}}
\toprule
\textbf{Dataset} & \textbf{Patients} & \textbf{1 Nodule} & \textbf{2 Nodules} & \textbf{3+ Nodules} & \textbf{Mean} & \textbf{Median} & \textbf{Max}\\
\midrule
DLCS24 & 1{,}613 & 1{,}129 (70.0\%) & 257 (15.9\%) & 227 (14.1\%) & 1.54 & 1 & 7\\
LUNA25 & 2{,}120 & 614 (29.0\%) & 388 (18.3\%) & 1{,}118 (52.7\%) & 2.91 & 2 & 17\\
LUNA16 & 601 & 263 (43.8\%) & 167 (27.8\%) & 171 (28.5\%) & 1.97 & 1 & 12\\
LUNGx & 70 & 59 (84.3\%) & 11 (15.7\%) & 0 (0.0\%) & 1.19 & 1 & 2\\
LNDbv4 & 212 & 30 (14.2\%) & 69 (32.5\%) & 113 (53.3\%) & 3.62 & 3 & 15\\
NSCLCR & 421 & 421 (100\%) & 0 & 0 & 1.00 & 1 & 1\\
IMDCT & 2{,}032 & 2{,}032 (100\%) & 0 & 0 & 1.00 & 1 & 1\\
\bottomrule
\end{tabular}}
\end{table}

\newpage
\section{Nodule Profile Dataset: Full 54-Attribute Schema}
\label{app:schema}

This appendix details the unified nodule profile schema underlying \sys, which examines lesion morphology, anatomical context, spatial position, distances to major thoracic structures, inter-nodule relationships, and reinsertion descriptors for synthetic generation. The analysis spans 13,140 nodules from seven datasets. Each profile CSV contains 54 shared core attributes plus dataset-specific auxiliary fields, so the total column count varies by source dataset. For clarity, the core attributes are grouped into nine functional categories, summarized in Table~\ref{tab:app_schema}.

\begin{table}[t]
\caption{Full 54-attribute Nodule Profile Dataset schema (9 functional groups).}
\label{tab:app_schema}
\centering
\footnotesize
\setlength{\tabcolsep}{3pt}
\renewcommand{\arraystretch}{1.18}
\begin{tabular}{@{}L{1.65cm}L{5.20cm}L{1.70cm}L{2.85cm}@{}}
\toprule
\textbf{Group} & \textbf{Attributes (\#)} & \textbf{Coverage} & \textbf{Purpose}\\
\midrule
Identity & \texttt{ct\_path} (1) & 100\% & Links each nodule to its source CT volume on disk\\
Bounding box & \shortstack[l]{\texttt{coordX}, \texttt{coordY},\\\texttt{coordZ}, \texttt{w}, \texttt{h}, \texttt{d} (6)} & 100\% & World-space centroid and bounding-box dimensions\\
Morphology & \shortstack[l]{\texttt{nodule\_mean\_diam\_mm},\\\texttt{nodule\_vol\_mm3} (2)} & 100 / 99.5\% & Size characterization\\
Anatomy & \shortstack[l]{\texttt{organ\_label\_id/name},\\\texttt{nearby\_organs\_10mm},\\\texttt{lung\_side}, \texttt{lobe\_name},\\\texttt{lung\_zone}, \texttt{central\_peripheral} (7)} & 96--100\%$^\ast$ & Anatomical location and organ context\\
Lung position & \shortstack[l]{\texttt{cranio\_caudal\_pct},\\\texttt{mediolateral\_pct},\\\texttt{anteroposterior\_pct} (3)} & 100\%$^\ast$ & Relative position within the whole lung\\
Lobe position & \shortstack[l]{\texttt{lobe\_cc\_pct}, \texttt{lobe\_ml\_pct},\\\texttt{lobe\_ap\_pct} (3)} & 100\%$^\ast$ & Relative position within the lobe\\
Distances & \shortstack[l]{6 structure-specific \texttt{dist\_to\_*\_mm} fields,\\\texttt{pleural\_distance\_mm},\\\texttt{airway\_distance\_mm} (8)} & 99.2--100\% & Distance in mm to major thoracic structures\\
Inter-nodule & \shortstack[l]{\texttt{n\_nodules\_in\_patient},\\\texttt{nearest\_nodule\_*}, \texttt{all\_nodule\_*},\\\texttt{ipsilateral}, \texttt{nearest\_same\_lobe},\\\texttt{bilateral\_distribution} (11)} & 0--95\%$^\dagger$ & Multi-nodule spatial relationships within a patient\\
Reinsertion & \shortstack[l]{\texttt{reinsertion\_\{lobe,}\\\texttt{lung\_side, lung\_zone,}\\\texttt{*\_pct, *\_dist, diam\}} (13)} & 97--100\% & Precomputed reinsertion parameters for synthetic trial generation\\
\midrule
\textbf{Total} & \textbf{54 attributes} & --- & Structured representation for profiling, querying, and trial generation\\
\bottomrule
\end{tabular}\\[2pt]
{\footnotesize $^\ast$ Coverage reflects the near-complete core anatomy and position fields used for localization. Some attributes, such as \texttt{nearby\_organs\_10mm}, are more dataset-dependent, and some position percentile fields are affected by known normalization issues.\\
$^\dagger$ Inter-nodule attributes are populated only for multi-nodule patients; single-nodule datasets therefore contain structurally empty nearest-neighbor fields.}
\end{table}
\paragraph{Bounding-box attributes (6).}
The bounding-box attributes describe each lesion through its world-space centroid (\texttt{coordX}, \texttt{coordY}, \texttt{coordZ}) and axis-aligned dimensions (\texttt{w}, \texttt{h}, \texttt{d}), with 100\% coverage across all 13{,}140 nodules. These fields provide the core geometric representation of the lesion, enabling consistent localization in the source CT, standardized patch extraction, and alignment with downstream profiling and synthetic reinsertion steps.

\begin{figure}[htbp]
  \centering
  \begin{subfigure}{0.82\linewidth}
    \centering
    \includegraphics[width=\linewidth]{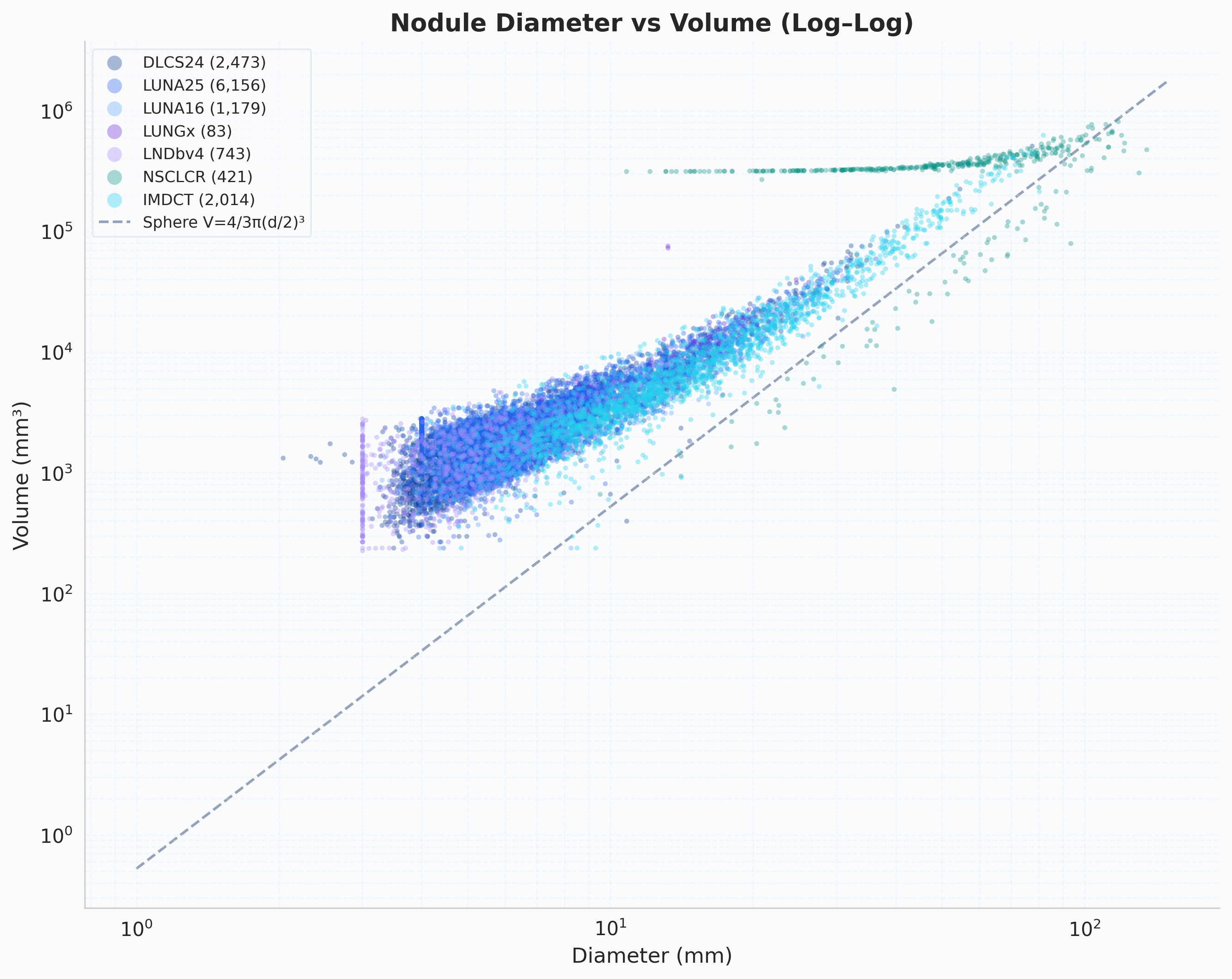}
    \caption{\textbf{Diameter--volume scatter across datasets.} Diameter--volume relationship, follows the expected spherical $V = \frac{4}{3}\pi(d/2)^3$ relationship on a log-log scale, with scatter above the theoretical line for irregular/non-spherical nodules.}
    \label{fig:app_diameter_volume_scatter}
  \end{subfigure}

  \vspace{3mm}

  \begin{subfigure}{\linewidth}
    \centering
    \includegraphics[width=\linewidth]{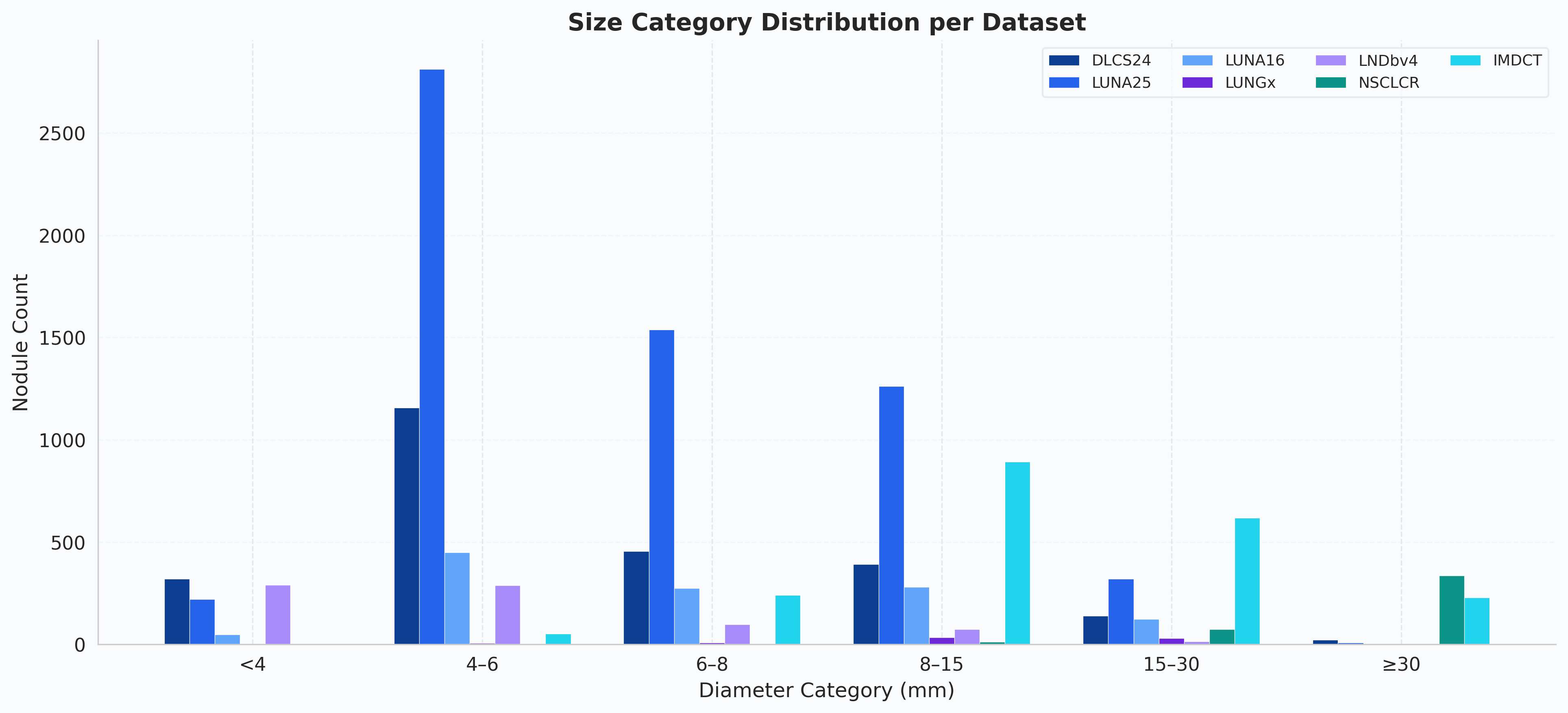}
    \caption{\textbf{Nodule size distribution across datasets.} Diameter-based size
    groupings (Lung--RADS management recommendations) used to summarize lesion morphology across the nodule profiles.
    }
    \label{fig:app_size_categories}
  \end{subfigure}

  \caption{\textbf{Morphology attributes.} Per--dataset summary statistics for nodule diameter and volume across the seven source datasets.}
  \label{fig:app_morphology_panels}
\end{figure}

\paragraph{Morphology attributes (2).}
The morphology group characterizes lesion size using two attributes: mean nodule diameter (\texttt{nodule\_mean\_diam\_mm}) and nodule volume (\texttt{nodule\_vol\_mm3}). Diameter was available for all 13{,}140 nodules (100\% coverage), while volume was available for 13{,}069 nodules (99.5\% coverage), with the small number of missing volume values arising primarily in LUNA25 and LUNA16 for the smallest nodules. Substantial cross-dataset variation was observed. Median diameter ranged from 4.55\,mm in LNDbv4 and 5.37\,mm in DLCS24 to 13.51\,mm in IMDCT and 51.85\,mm in NSCLCR, reflecting the expected contrast between screening-oriented cohorts and advanced cancer datasets. Volume followed the same pattern, with median values ranging from 1{,}328\,mm$^3$ in LNDbv4 and 1{,}596\,mm$^3$ in DLCS24 to 5{,}481\,mm$^3$ in IMDCT and 337{,}423\,mm$^3$ in NSCLCR. Together, these features provide a compact but highly informative representation of lesion size across the unified profile space. Table~\ref{tab:app_morphology_stats} and Figure~\ref{fig:app_morphology_panels} provide two complementary views of cross-dataset lesion-size variation in the Nodule Profile Dataset.

\begin{table}[htbp]
\caption{Per-dataset morphology statistics for lesion diameter and volume.}
\label{tab:app_morphology_stats}
\centering\footnotesize
\setlength{\tabcolsep}{3pt}
\resizebox{\linewidth}{!}{%
\begin{tabular}{@{}lccccc@{}}
\toprule
\textbf{Dataset} & \shortstack{\textbf{Diameter Mean $\pm$ SD}\\\textbf{(mm)}} & \shortstack{\textbf{Diameter Median}\\\textbf{(IQR)}} & \shortstack{\textbf{Volume Mean $\pm$ SD}\\\textbf{(mm$^3$)}} & \shortstack{\textbf{Volume}\\\textbf{Median}} & \textbf{$N$}\\
\midrule
DLCS24 & $7.08 \pm 6.87$ & 5.37 (3.86--7.51) & $3{,}817 \pm 13{,}851$ & 1{,}596 & 2{,}487 / 2{,}473\\
LUNA25 & $7.18 \pm 6.21$ & 5.68 (4.26--7.98) & $3{,}207 \pm 4{,}233$ & 2{,}077 & 6{,}163 / 6{,}156\\
LUNA16 & $8.31 \pm 5.65$ & 6.75 (4.63--9.97) & $3{,}784 \pm 4{,}961$ & 2{,}029 & 1{,}186 / 1{,}179\\
LUNGx & $13.22 \pm 8.16$ & 11.25 (7.19--17.36) & $11{,}516 \pm 13{,}453$ & 8{,}599 & 83 / 83\\
LNDbv4 & $5.43 \pm 4.18$ & 4.55 (3.55--5.99) & $2{,}124 \pm 3{,}709$ & 1{,}328 & 768 / 743\\
NSCLCR & $54.39 \pm 25.55$ & 51.85 (34.82--72.49) & $330{,}229 \pm 146{,}809$ & 337{,}423 & 421 / 421\\
IMDCT & $17.04 \pm 13.41$ & 13.51 (8.51--21.49) & $20{,}257 \pm 47{,}902$ & 5{,}481 & 2{,}032 / 2{,}014\\
\bottomrule
\end{tabular}}\\[2pt]
{\footnotesize In the last column, the first number denotes diameter $N$ and the second denotes volume $N$.}
\end{table}

\begin{figure}[htbp]
  \centering
  \includegraphics[width=\linewidth]{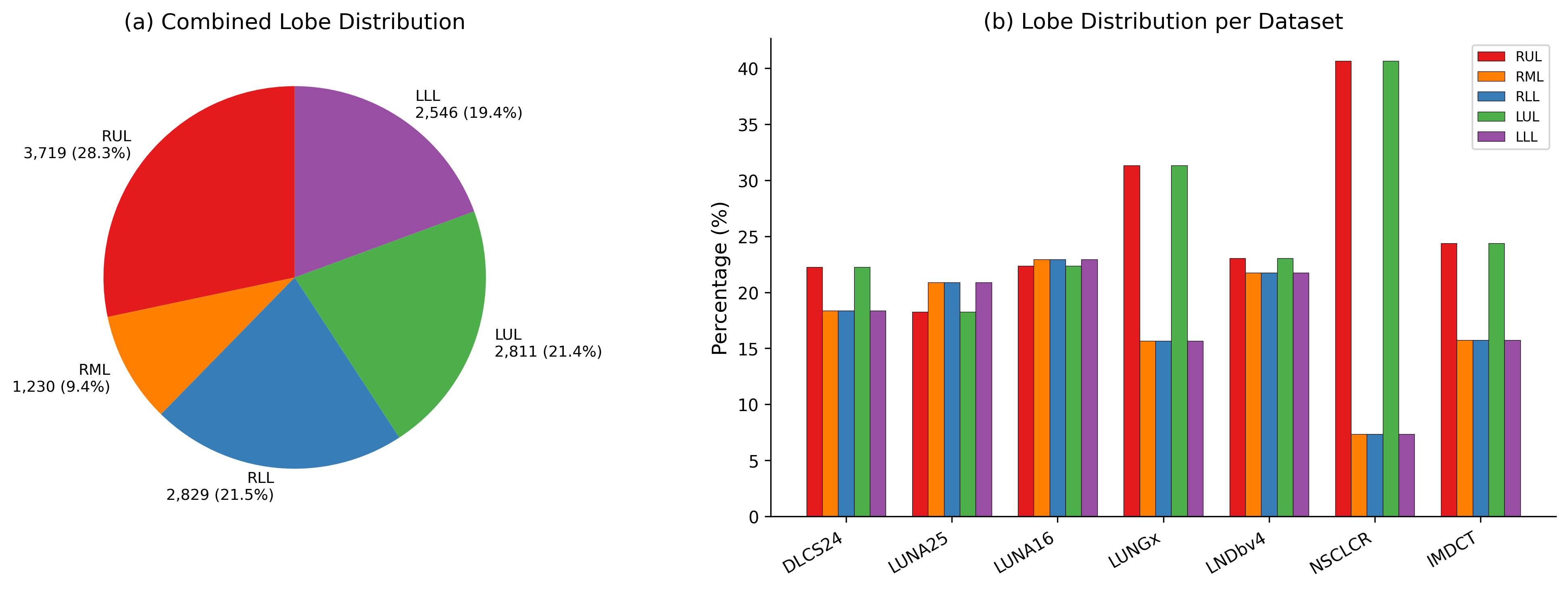}
  \caption{\textbf{Anatomical lobe distribution.} (a) Combined distribution showing upper-lobe predominance (RUL 28.3\%, LUL 21.4\%, total 49.7\%) and right-lung predominance (59.2\%). (b) Per-dataset lobe percentages. Distribution is broadly consistent across datasets with minor variations reflecting different clinical populations. RUL = right upper lobe, RML = right middle lobe, RLL = right lower lobe, LUL = left upper lobe, LLL = left lower lobe.}
  \label{fig:app_lobe_distribution}
\end{figure}

\paragraph{Anatomy attributes (7).}
The anatomy group captures each lesion's anatomical context using organ labels, nearby-organ context, lung side, lobe, lung zone, and central-versus-peripheral location. Core localization fields were near-complete across the cohort, with lobe labels available for 13{,}105 of 13{,}140 nodules (99.6\%). Anatomically, nodules showed right-lung predominance (57.7\% right versus 42.3\% left) and upper-lobe predominance, with 49.9\% of nodules located in the RUL or LUL combined. Across datasets, lesions were overwhelmingly peripheral, with peripheral fractions ranging from 91.2\% in NSCLCR to 98.9\% in LUNA25 and LUNA16; NSCLCR correspondingly exhibited the highest central fraction (8.8\%), consistent with its more advanced tumour cohort. In contrast, nearby-organ annotations were more dataset-dependent and were not uniformly suitable for cross-dataset comparison. Figure~\ref{fig:app_lobe_distribution} shows a pie chart of the combined lobe distribution and a grouped bar chart of dataset-specific lobe percentages.

\begin{figure}[ht]
  \centering
  \includegraphics[width=\linewidth]{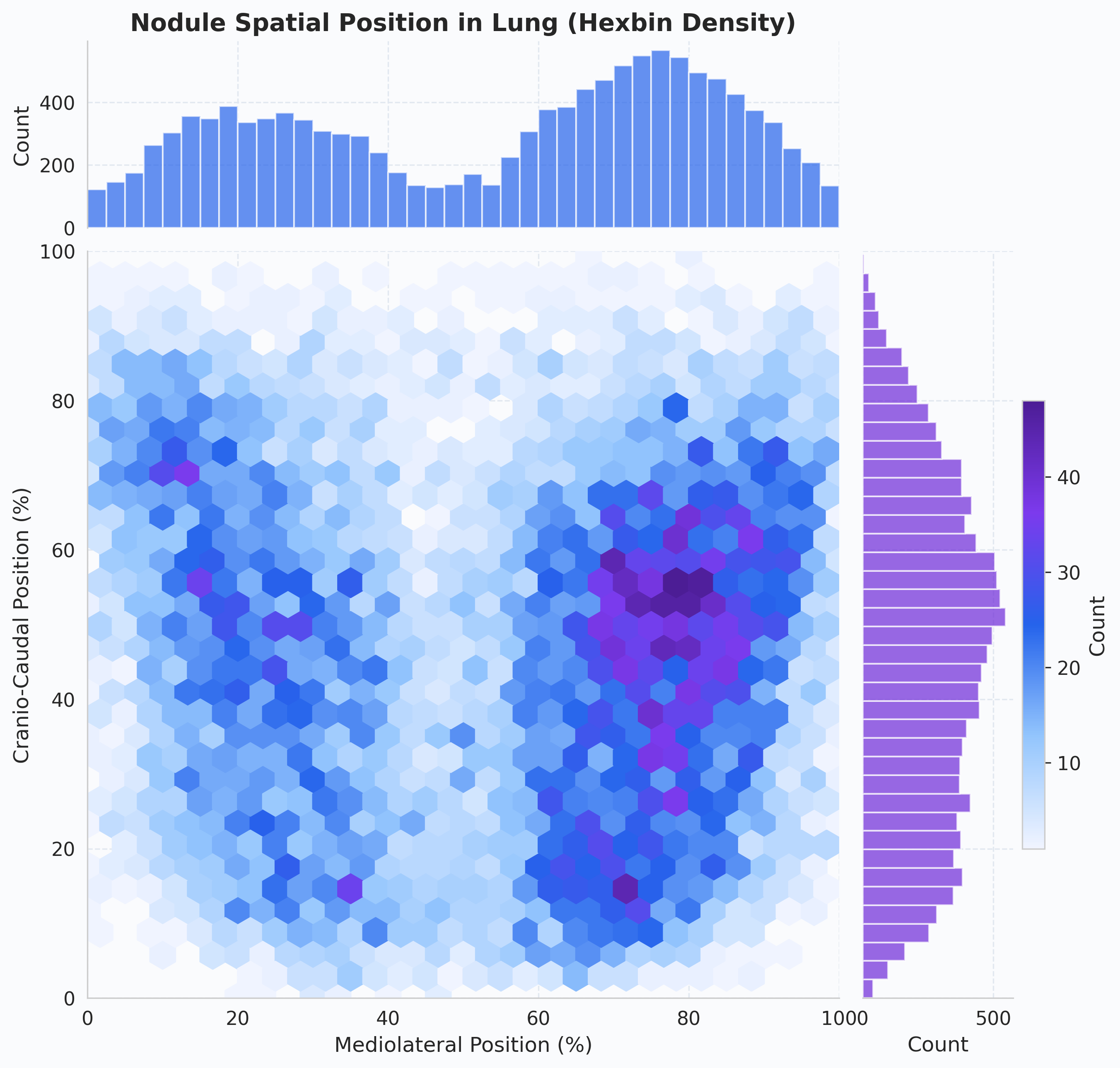}
  \caption{\textbf{Lung-level spatial distribution of nodules.} Cranio-caudal position shows a slight apical bias across datasets (mean $\sim$45--49\%), with IMDCT exhibiting the lowest mean (40.6\%). The density map indicates concentration in the central and upper lung regions.}
  \label{fig:app_spatial_density}
\end{figure}

\paragraph{Lung position attributes (3).}
The lung-position group encodes each lesion's relative location within the whole lung using cranio-caudal, mediolateral, and anteroposterior percentiles. Among these, \texttt{cranio\_caudal\_pct} showed the most reliable behavior, with 97.6--100\% of values falling within the expected range across datasets, while \texttt{mediolateral\_pct} was moderately reliable, with 93--97\% of values in range. In contrast, \texttt{anteroposterior\_pct} was affected by a known normalization issue and was not reliable for quantitative interpretation, with only 0--62\% of values falling within the expected range and complete failure in some datasets. Accordingly, lung-level positional analyses focused on the reliable cranio-caudal and mediolateral descriptors, which indicated a slight apical bias overall, with mean cranio-caudal position typically around 45--49\% and a lower mean of 40.6\% in IMDCT, suggesting relatively more upper-lung nodules in that cohort. Figure~\ref{fig:app_spatial_density} visualizes the lung-level spatial distribution of nodules using positional descriptors.

\paragraph{Lobe position attributes (3).}
The lobe-position group encodes each lesion's relative location within its assigned lobe using cranio-caudal, mediolateral, and anteroposterior percentiles. Among these, \texttt{lobe\_cc\_pct} was the only reliable descriptor, with 97.6--100\% of values falling within the expected range across datasets. In contrast, \texttt{lobe\_ml\_pct} and \texttt{lobe\_ap\_pct} were affected by the same normalization issue observed in the lung-level AP field and were therefore excluded from quantitative interpretation. Accordingly, lobe-level positional analysis was restricted to the cranio-caudal axis.

\begin{figure}[t]
  \centering
  \includegraphics[width=\linewidth]{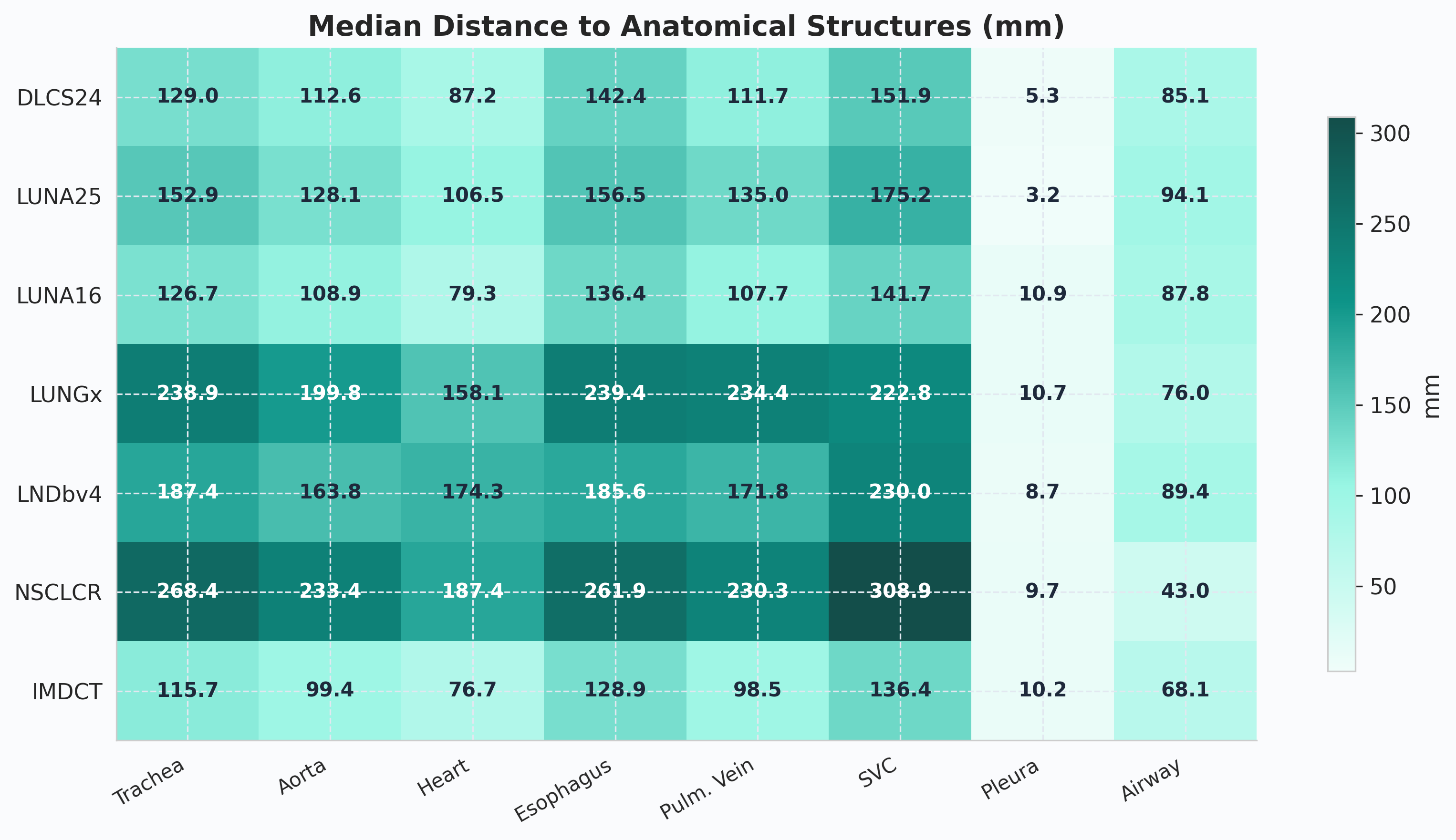}
  \caption{\textbf{Distance visualizations.} Heatmap showing cross-dataset patterns in lesion distance to major thoracic structures, highlighting the anatomical organization captured by the distance-based descriptors.}
  \label{fig:app_distance_heatmap}
\end{figure}

\paragraph{Distances attributes (8).}
The distance group captures lesion proximity to surrounding thoracic anatomy using eight Euclidean distance features measured from the nodule centroid to the nearest surface of the pleura, airway, heart, aorta, pulmonary vein, trachea, esophagus, and superior vena cava. These descriptors were highly complete across datasets, with coverage ranging from 99.2\% to 100\%. As shown in Table~\ref{tab:distance_summary}, pleural distance had the smallest median value at 5.8\,mm (IQR 1.8--13.2\,mm), reflecting the strong peripheral predominance of nodules, while superior vena cava distance had the largest median value at 172.8\,mm (IQR 108.2--233.2\,mm). The remaining structures showed intermediate median distances of 85.1\,mm to the airway, 101.5\,mm to the heart, 121.7\,mm to the aorta, 129.8\,mm to the pulmonary vein, 143.9\,mm to the trachea, and 149.6\,mm to the esophagus. Figure~\ref{fig:app_distance_heatmap} further illustrates the cross-dataset organization of these features, highlighting the anatomical structure captured by the distance-based descriptors.

\begin{table}[t]
\caption{Distance attributes. Combined summary statistics for lesion distance to major thoracic structures across the Nodule Profile Dataset.}
\label{tab:distance_summary}
\centering
\footnotesize
\setlength{\tabcolsep}{6pt}
\begin{tabular}{@{}lcc@{}}
\toprule
\textbf{Structure} & \textbf{Median distance (mm)} & \textbf{Q1--Q3 (mm)} \\
\midrule
Pleura & 5.8 & 1.8--13.2 \\
Airway & 85.1 & 61.2--110.3 \\
Heart & 101.5 & 51.2--175.1 \\
Aorta & 121.7 & 75.5--167.3 \\
Pulmonary vein & 129.8 & 77.2--186.0 \\
Trachea & 143.9 & 89.7--198.5 \\
Esophagus & 149.6 & 98.5--197.0 \\
Superior vena cava & 172.8 & 108.2--233.2 \\
\bottomrule
\end{tabular}
\end{table}

\begin{figure}[t]
  \centering
  \includegraphics[width=0.90\linewidth]{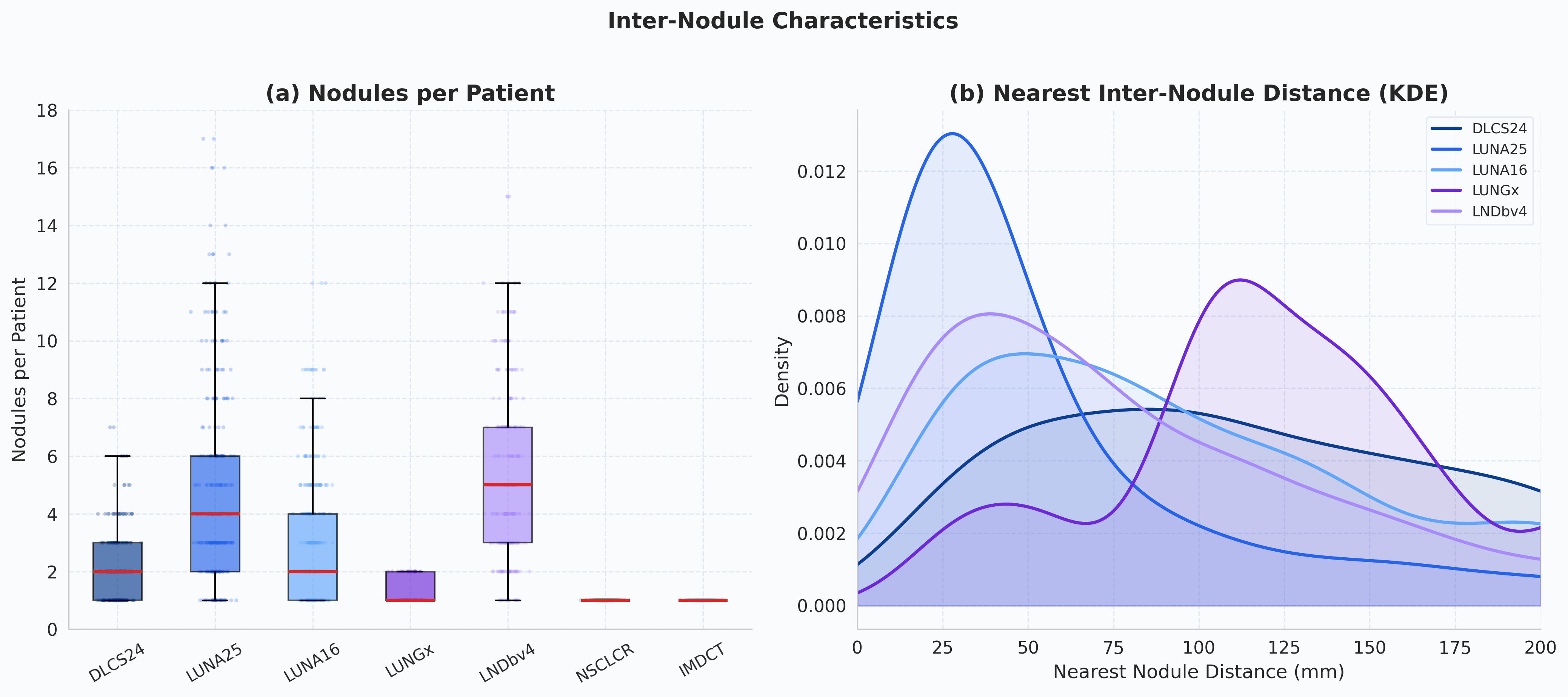}
  \caption{\textbf{Inter-nodule attribute visualization.} Cross-dataset summary of patient-level lesion multiplicity, highlighting variation in nodule burden and multi-nodule prevalence across the seven source datasets.}
  \label{fig:app_internodule}
\end{figure}

\paragraph{Inter-nodule attributes (11).}
The inter-nodule group captures patient-level multiplicity and spatial relationships between nodules using attributes such as the number of nodules per patient, nearest-nodule identity and distance, directional offsets, laterality, same-lobe status, and bilateral distribution. These features revealed substantial cross-dataset variation in lesion burden. As summarized in Table~\ref{tab:app_inter_nodule_summary} and visualized in Figure~\ref{fig:app_internodule}, LUNA25 and LNDbv4 showed the highest multi-nodule burden, with mean counts of 4.86 and 5.21 nodules and $\geq 3$ nodules in 52.7\% and 53.3\% of cases, respectively, whereas NSCLCR and IMDCT were single-nodule-per-patient by design. Nearest-nodule distance was defined only for multi-nodule cases and was lowest in LUNA25 and LNDbv4, with mean values of 83.7\,mm and 80.9\,mm, respectively, indicating denser within-patient lesion patterns in these cohorts. Bilateral distribution further highlighted these differences, reaching 76.2\% in LNDbv4 compared with 36.4\% in DLCS24 and 43.5\% in LUNA25. Together, these attributes provide a structured description of multi-lesion context that is not captured by single-nodule morphology or anatomy alone.

\begin{table}[t]
\caption{Inter-nodule burden summary across datasets.}
\label{tab:app_inter_nodule_summary}
\centering\footnotesize
\setlength{\tabcolsep}{6pt}
\begin{tabular}{@{}lccccc@{}}
\toprule
\textbf{Dataset} & \textbf{Mean} & \textbf{Median} & \textbf{Max} & \textbf{Multi-nodule ($>1$)} & \textbf{Multi-nodule ($\geq 3$)}\\
\midrule
DLCS24 & 2.06 & 2 & 7 & 61.2\% & 16.8\%\\
LUNA25 & 4.86 & 4 & 17 & 85.5\% & 52.7\%\\
LUNA16 & 3.08 & 2 & 12 & 71.1\% & 36.2\%\\
LUNGx & 1.31 & 1 & 2 & 31.3\% & 0.0\%\\
LNDbv4 & 5.21 & 5 & 15 & 89.4\% & 53.3\%\\
NSCLCR & 1.00 & 1 & 1 & 0.0\% & 0.0\%\\
IMDCT & 1.00 & 1 & 1 & 0.0\% & 0.0\%\\
\bottomrule
\end{tabular}
\end{table}

\begin{figure}[t]
  \centering
  \includegraphics[width=0.92\linewidth]{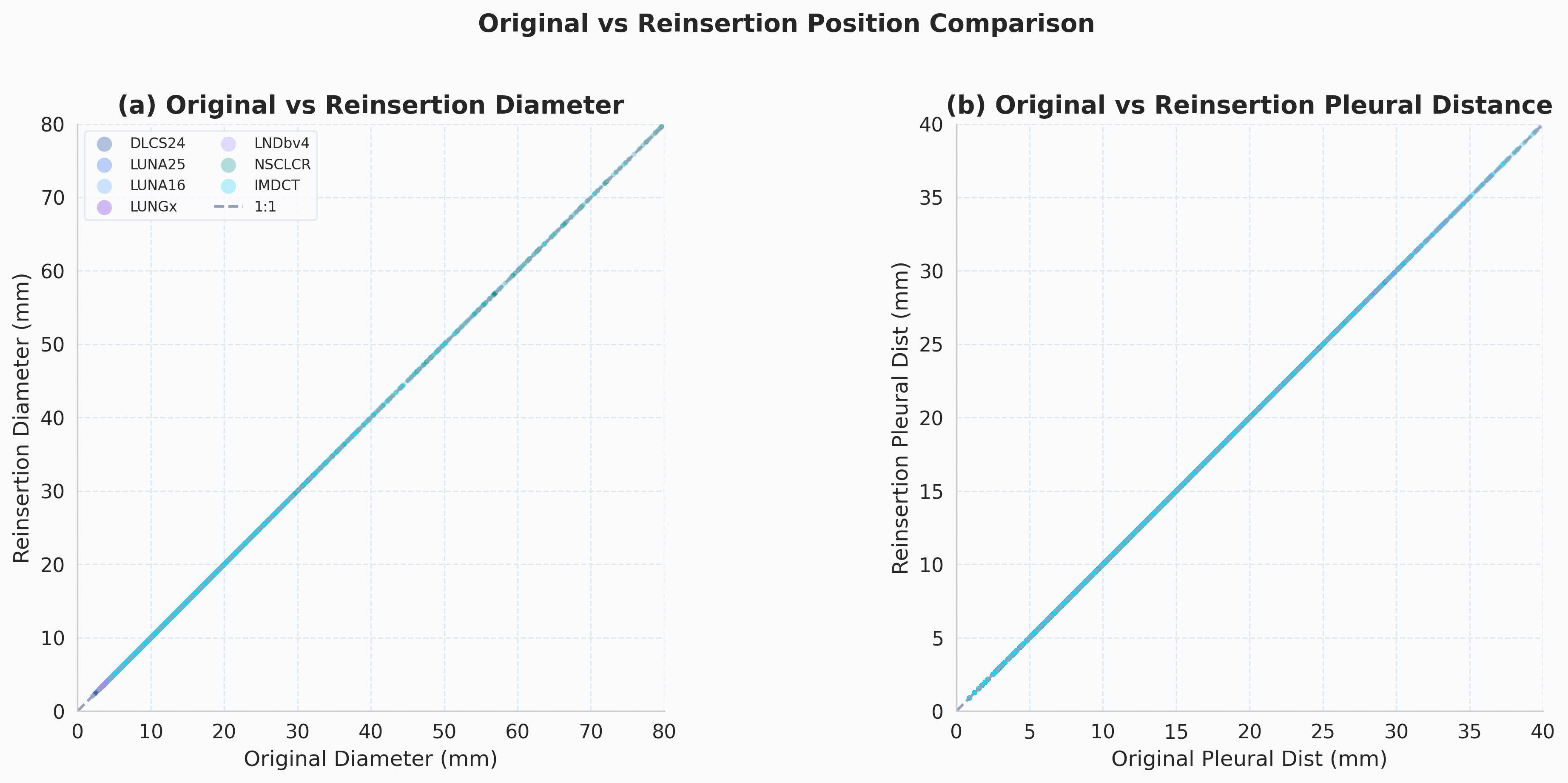}
  \caption{\textbf{Original vs. reinsertion attribute comparison.} Reinsertion parameters preserve donor nodule diameter exactly by design while maintaining close agreement in pleural-distance context under host-specific anatomical constraints.}
  \label{fig:app_reinsertion_comparison}
\end{figure}

\textbf{Reinsertion attributes (13).} The reinsertion group defines the precomputed placement blueprint used for anatomy-aware synthetic generation. These attributes specify the target lobe, lung side, and lung zone, together with lung- and lobe-level positional percentiles, pleural and airway distance constraints, and target nodule diameter. As summarized in Table~\ref{tab:app_schema}, coverage was near-complete across datasets: categorical and positional reinsertion fields were available for all nodules, distance-based reinsertion fields achieved 97--100\% coverage, and reinsertion diameter was available in all cases. Figure~\ref{fig:app_reinsertion_comparison} further illustrates the relationship between original and reinsertion parameters, showing exact preservation of diameter by design and approximate preservation of pleural-distance context under host-specific anatomical constraints. Together, these attributes provide a deterministic and anatomically informed representation for relocating donor nodules into compatible host CT volumes during synthetic trial construction.

\newpage
\section{Mode Definitions.}
\label{app:modes_defi}
All 13 modes are parameterisations of $\mathcal{S}$; they share the same Build operator and differ only in how they constrain or vary its components.
Two families address fundamentally different evaluation questions.

\medskip
\noindent\textbf{Synthetic cohort modes (M1--M10).}
In all synthetic cohort modes the donor nodule $(p,j)$ and host anatomy $q$ come from \emph{different} patients ($q \neq p$): the synthetic CT is a controlled composition of real anatomy plus a placed real nodule, not a reconstruction of any original scan.
Each mode holds the full TrialSpec $\mathcal{S}$ fixed except for a single experimental dimension.

\begin{enumerate}

\item \textbf{Prevalence Control (M1).}
Establishes the baseline reference cohort calibrated to a published screening prior, against which all other modes are compared. \emph{Definition:} $\mathcal{S}_{\mathrm{M1}} = (n,\;\pi{=}0.04,\;T_{\mathrm{NLST}},\;\phi_{\mathrm{nod}}{=}\top,\;\phi_{\mathrm{ins}}{=}\phi_{\mathrm{ins}}^{\mathrm{default}},\;\phi_{\mathrm{demo}}{=}\top,\;\sigma{=}0,\;B{=}0,\;\mathcal{D}_{\mathrm{excl}}{=}\emptyset)$, under $T_{\mathrm{NLST}}$, 4\% of sampled cases are malignant ($\pi{=}0.04$).

\item \textbf{Size Control (M2).} Detectability expected to have non-monotonic sensitivity to nodule diameter, this mode stratifies by size bins while holding all other \sys components, enabling a direct measurement of size effect. 
  \emph{Definition:} Six sub-cohorts $\{\mathcal{S}_{\mathrm{M2}}^{(i)}\}_{i=1}^{6}$ with
  $\phi_{\mathrm{nod}}^{(i)} = [d \in s_i]$,
  $s_i \in \{[0,4),\,[4,6),\,[6,10),\,[10,20),\,[20,30),\,[30,\infty)\}\,\mathrm{mm}$,
  $n$ cases per sub-cohort.

\item \textbf{Location isolation (M3).} Enables controlled evaluation of lobe-specific model
behaviour by holding nodule size fixed across all five lobes, a cohort
design that cannot be constructed from real registries, where lobe and
size are naturally correlated.
\textit{Definition}: Five lobe sub-cohorts $\{\mathcal{S}_{\mathrm{M3}}^{(\ell)}\}_{\ell \in \mathcal{L}}$
with $\phi_{\mathrm{nod}}^{(\ell)} = [\mathrm{lobe}{=}\ell \wedge d \in [6,15]\,\mathrm{mm}]$,
$n$ cases per lobe,
$\ell \in \{\mathrm{RUL,\,RML,\,RLL,\,LUL,\,LLL}\}$.

\item \textbf{Demographic stratification (M4).} Lung CT texture, airway anatomy, and parenchymal density differ across age and sex, which can drive spurious model accuracy differences.
  M4 constructs matched sub-cohorts per demographic stratum using the source dataset with complete demographic metadata.
\textit{Definition}:
$\phi_{\mathrm{demo}}^{(s)}(q) = [\mathrm{dataset}(q){=}\mathrm{DLCS24} \wedge s(q)]$
where $s \in \{\mathrm{M}_{<\tau},\;\mathrm{M}_{\ge\tau},\;\mathrm{F}_{<\tau},\;\mathrm{F}_{\ge\tau}\}$,
and $\tau$ is an age threshold in years,
$n_s$ cases per stratum ($N_{\mathrm{total}}{=}4n_s$).

\item \textbf{Counterfactual prevalence (M5).} Apparent model accuracy shifts with class prevalence even when the model is unchanged, because threshold-based metrics depend on the positive:negative ratio.
  M5 separates this statistical artifact from true model behaviour by holding host anatomy constant while varying prevalence.
  \emph{Definition:}
  \begin{equation*}
    \mathcal{C} = \bigl\{\mathcal{S}_v \;:\; v \in \{0.01, 0.02, 0.05, 0.10, 0.20\},\;\sigma\;\text{fixed}\bigr\}
  \end{equation*}
  Fixing $\sigma$ reuses the same host anatomies at every $v$; the causal prevalence effect is
  $\Delta_{\mathrm{acc}}(v_1,v_2) = \mathrm{acc}(\mathcal{M}_{v_1}) - \mathrm{acc}(\mathcal{M}_{v_2})$.

\item \textbf{Cross-dataset transfer (M6).} Nodule appearance (texture, boundary sharpness, attenuation) varies with acquisition scanner and reconstruction kernel across datasets; M6 tests whether a model's performance is confounded by nodule source independent of anatomy.
  \emph{Definition:} Five sub-cohorts, one per source dataset $D_k$:
  $\mathcal{D}_{\mathrm{excl}}^{(k)} = \mathcal{D} \setminus \{D_k\}$ with $\phi_{\mathrm{demo}}^{(k)}(q) = [\mathrm{dataset}(q) \neq D_k]$ (donors from $D_k$ only, hosts from the complement).

\item \textbf{Bootstrap CI (M7).} A single cohort of size $n$ yields a point estimate of accuracy; M7 quantifies the sampling uncertainty of that estimate by resampling the specification with different seeds.
  \emph{Definition:}
  $B$ replicates $\mathcal{M}^{(b)} = \mathrm{Build}(\mathcal{S}_{\mathrm{M1}},\,\sigma{+}b)$,
  $b = 1,\ldots,B$.
  Metric CI from percentile interval $[\hat{\theta}_{(\alpha/2)},\,\hat{\theta}_{(1-\alpha/2)}]$
  over the $B$ replicates.

\item \textbf{Algorithm comparison (M8).} Head-to-head model comparisons are confounded if each model is tested on a different cohort realisation; M8 eliminates this by fixing a single held-out manifest shared across all models. We therefore define a shared union exclusion set $\mathcal{D}_{\mathrm{train}}^{\cup} = \bigcup_m \mathcal{D}_{\mathrm{train}}^{(m)}$, where $\mathcal{D}_{\mathrm{train}}^{(m)}$ denotes the public datasets documented in model $m$'s pre-training corpus. With $\sigma{=}42$ and $n$ cases, M8 applies $\mathcal{D}_{\mathrm{excl}} = \mathcal{D}_{\mathrm{train}}^{\cup}$ at donor-pool construction time, so the same held-out manifest is submitted to every evaluated model.

\item \textbf{Screening simulation (M9).} In longitudinal screening, most prevalent nodules are
detected in round 1, so subsequent cohorts are enriched for harder, incident
lesions at progressively lower prevalence.
M9 replicates this dynamic in a controlled setting.
\textit{Definition}: Three rounds with geometrically decaying prevalence
$\pi_r = \pi_0 \cdot \gamma^{\,r}$, $r \in \{0,1,2\}$,
where $\pi_0$ is the baseline prevalence and $\gamma \in (0,1)$ is the
round-to-round incidence decay factor;
$n_r$ cases per round, hosts and donors drawn independently per round.

\item \textbf{Multi-nodule context (M10).} Real lung CTs often contain multiple nodules, and visual
models may attend to the most salient lesion rather than the designated
target; M10 measures how concurrent nodule presence affects per-target
accuracy.
\textit{Definition}: Mixed manifest with fraction $\alpha$ of single-nodule
cases and $(1{-}\alpha)$ multi-nodule cases, each with
$N_{\mathrm{nod}} \sim \mathcal{U}[N_{\min}, N_{\max}]$ nodules inserted
simultaneously into a single host CT.

\end{enumerate}

\medskip
\noindent\textbf{Digital twin modes (M11--M13).}
Digital twin modes preserve the identity relationship between donor and host.
Rather than pairing nodules with foreign anatomies, they use specialised Build operators to construct patient-specific ground-truth scenarios that cannot be obtained from real scans (where you cannot separate a single nodule from its native anatomy without intervention).

\begin{enumerate}

\item \textbf{Twin isolation (M11).} To obtain per-lesion performance under a patient's own anatomy, removing cross-patient anatomical confounds. Each nodule is re-inserted into its original scan as if it were placed anew.
  \emph{Definition:}
  $\mathrm{Build}_{\mathrm{Iso}}$: host $q{=}p$; donor $(p,j)$ placed independently.
  One specification per annotated nodule.

\item \textbf{Twin complete (M12).} A patient's clinical CT contains \emph{all} of their nodules simultaneously; M12 reconstructs this complete scenario to evaluate multi-nodule CT quality and per-patient accuracy, producing one synthetic CT per patient rather than one per nodule.
  \emph{Definition:}
  $\mathrm{Build}_{\mathrm{Comp}}$: host $q{=}p$; all $N_p$ native nodules $(p,1),\ldots,(p,N_p)$ inserted simultaneously. One specification per patient.

\item \textbf{Twin cross (M13).} To isolate the contribution of host anatomy to per-nodule accuracy independent of nodule properties, M13 transplants a nodule into a \emph{different} patient's anatomy, enabling direct anatomy ablation.
  \emph{Definition:}
  $\mathrm{Build}_{\mathrm{Cross}}(D_{\mathrm{host}}, D_{\mathrm{donor}})$: host $q \neq p$.
  Assignment matrix $\mathbf{A} \in \{0,1\}^{|\mathcal{P}| \times |\mathcal{P}|}$ governs donor-to-host pairing via three policies:
  (i)~\texttt{one\_to\_one}: each patient hosts exactly one foreign nodule;
  (ii)~\texttt{one\_to\_many\_hosts} ($N{=}3$): each donor placed into 3 hosts;
  (iii)~\texttt{donor\_patient\_complete}: all nodules of patient $p$ transplanted into patient $q$.

\end{enumerate}

\newpage
\section{Virtual Lesion Study (VLS): Trial Mode Specification}
\label{app:modes}

We instantiated 13 trial modes to construct the VLS corpus, comprising ten synthetic cohort modes (M1--M10) and three digital twin modes (M11--M13). For each mode, we specify the configuration used in our experiments, the corresponding trial formulation, and the scientific question it was designed to evaluate. Clinical-trial template parameters shared across modes are reported in Table~\ref{tab:app_templates}, and per-mode cohort counts are summarized in Table~\ref{tab:mode_summary}.

\textbf{Clinical Trial Templates.} We calibrated synthetic cohort modes to two published screening trials using
the template $T = (\pi_T, \mathbf{w}_T, \boldsymbol{\lambda}_T,
\boldsymbol{\mu}_T)$, where $\pi_T$ is the default malignancy prevalence,
$\mathbf{w}_T \in \Delta^6$ is the six-bucket size prior over
$\{[0,4),[4,6),[6,10),[10,20),[20,30),[30,\infty)\}$\,mm,
$\boldsymbol{\lambda}_T \in \Delta^5$ gives lobe prior weights
(RUL, RML, RLL, LUL, LLL), and $\boldsymbol{\mu}_T$ encodes age
distribution, sex ratio, and smoking stratum weights.
Any mode-specific $\phi_{\mathrm{nod}}$, $\phi_{\mathrm{demo}}$, or $\pi$
fields override these template defaults at Build time.

\begin{table}[htbp]
\caption{Clinical trial template parameters used in this paper.
NLST values from \citet{aberle2011nlst};
NELSON values from \citet{de_koning2020nelson}.}
\label{tab:app_templates}
\centering\small
\setlength{\tabcolsep}{5pt}
\begin{tabular}{@{}lcc@{}}
\toprule
\textbf{Parameter}
  & \textbf{NLST} ($T_{\mathrm{NLST}}$)
  & \textbf{NELSON} ($T_{\mathrm{NELSON}}$) \\
\midrule
Prevalence $\pi_T$             & 0.040 & 0.022 \\
$w_1$: $[0,4)$\,mm             & 0.42  & 0.38  \\
$w_2$: $[4,6)$\,mm             & 0.38  & 0.40  \\
$w_3$: $[6,10)$\,mm            & 0.11  & 0.14  \\
$w_4$: $[10,20)$\,mm           & 0.05  & 0.05  \\
$w_5$: $[20,30)$\,mm           & 0.03  & 0.02  \\
$w_6$: $[30,\infty)$\,mm       & 0.01  & 0.01  \\
Age distribution               & $\mathcal{N}(61.4,5.0)$, clip $[55,74]$
                               & $\mathcal{U}[50,75]$ \\
Male sex ratio                 & 0.59  & 0.84  \\
$\boldsymbol{\lambda}$: RUL/RML/RLL/LUL/LLL
                               & 0.30/0.08/0.18/0.28/0.16
                               & 0.28/0.08/0.19/0.27/0.18 \\
\bottomrule
\end{tabular}
\end{table}

\paragraph{M1, Prevalence control.}
We drew 1{,}000 cases from the full pool under the NLST prior, placing no
filter constraints on size, lobe, or demographics, and fixed the seed at
$\sigma{=}0$ to ensure exact reproducibility:
\begin{equation*}
  \mathcal{S}_{\mathrm{M1}} = \left(
  \begin{aligned}
    &n{=}1{,}000,\;\pi{=}0.04,\;T_{\mathrm{NLST}},\\
    &\phi_{\mathrm{nod}}{=}\top,\;\phi_{\mathrm{ins}}{=}\phi_{\mathrm{ins}}^{\mathrm{default}},\\
    &\phi_{\mathrm{demo}}{=}\top,\;\sigma{=}0,\;B{=}0,\;\mathcal{D}_{\mathrm{excl}}{=}\emptyset
  \end{aligned}
  \right).
\end{equation*}

\paragraph{M2, Size control.}
We ran six sub-cohorts of 100 cases each, one per diameter bin, holding all
other fields at M1 defaults so that diameter was the only varying factor:
\begin{equation*}
  \begin{aligned}
    \mathcal{S}_{\mathrm{M2}}^{(i)} &=
    \mathcal{S}_{\mathrm{M1}}\bigl[
      \phi_{\mathrm{nod}} \leftarrow [d \in s_i],\;n \leftarrow 100\bigr],\\
    s_i &\in \{[0,4),[4,6),[6,10),[10,20),[20,30),[30,\infty)\}\,\mathrm{mm}.
  \end{aligned}
\end{equation*}

\paragraph{M3, Location isolation.}
We ran five sub-cohorts of 100 cases each, one per lobe, restricting the
diameter to $[6,15]$\,mm across all lobes so that lobe identity was the only
remaining variable:
\begin{equation*}
  \begin{aligned}
    \mathcal{S}_{\mathrm{M3}}^{(\ell)} &=
    \mathcal{S}_{\mathrm{M1}}\bigl[
      \phi_{\mathrm{nod}} \leftarrow [\mathrm{lobe}{=}\ell \wedge d\in[6,15]\,\mathrm{mm}],\;
      n \leftarrow 100\bigr],\\
    \ell &\in \{\mathrm{RUL,RML,RLL,LUL,LLL}\}.
  \end{aligned}
\end{equation*}

\paragraph{M4, Demographic stratification.}
We ran four matched sub-cohorts of 200 cases each across sex-and-age strata,
restricting hosts to DLCS24:
\begin{equation*}
  \mathcal{S}_{\mathrm{M4}}^{(g)} =
  \mathcal{S}_{\mathrm{M1}}\bigl[
    \phi_{\mathrm{demo}} \leftarrow [\mathrm{dataset}{=}\mathrm{DLCS24} \wedge g],\;
    n \leftarrow 200\bigr],\quad
  g \in \{\mathrm{M}{<}65,\mathrm{M}{\ge}65,\mathrm{F}{<}65,\mathrm{F}{\ge}65\}.
\end{equation*}

\paragraph{M5, Counterfactual prevalence.}
We ran five specifications of 500 cases each across a prevalence grid,
holding the seed fixed at $\sigma{=}42$ so that the same host anatomies
appeared at every level, making any metric shift causally attributable to
prevalence alone:
\begin{equation*}
  \begin{aligned}
    \mathcal{C} &= \bigl\{
      \mathcal{S}_{\mathrm{M1}}[\pi\!\leftarrow\!v,\;\sigma\!\leftarrow\!42]
      : v\in\{0.01,0.02,0.05,0.10,0.20\}\bigr\},\\
    \Delta_\theta(v_1,v_2) &= \theta(\mathcal{M}_{v_1})-\theta(\mathcal{M}_{v_2}).
  \end{aligned}
\end{equation*}

\paragraph{M6, Cross-dataset transfer.}
We ran five sub-cohorts of 300 cases each, drawing donors exclusively from
one source dataset $D_k$ while hosts came from all remaining datasets:
\begin{equation*}
  \mathcal{S}_{\mathrm{M6}}^{(k)} =
  \mathcal{S}_{\mathrm{M1}}\bigl[
    \mathcal{D}_{\mathrm{excl}} \leftarrow \mathcal{D}\setminus\{D_k\},\;
    \phi_{\mathrm{demo}} \leftarrow [\mathrm{dataset} \neq D_k],\;
    n \leftarrow 300\bigr],\quad k\in\{1,\ldots,5\}.
\end{equation*}

\paragraph{M7, Bootstrap confidence intervals.}
We ran 20 independent replicates of M1 under incremented seeds and computed
the 95\% empirical percentile CI across replicates:
\begin{equation*}
  \mathcal{M}^{(b)} = \mathrm{Build}(\mathcal{S}_{\mathrm{M1}},\;\sigma{=}b),
  \quad b=1,\ldots,20;\qquad
  \mathrm{CI}_{0.95}(\hat\theta) =
  \bigl[\hat\theta_{(0.025)},\;\hat\theta_{(0.975)}\bigr].
\end{equation*}

\paragraph{M8, Algorithm comparison.}
We ran a single fixed-seed held-out manifest of 500 cases submitted
identically to all evaluated models. To preserve a shared manifest while
enforcing contamination control, we defined a model-union exclusion set
\begin{equation*}
  \mathcal{D}_{\mathrm{train}}^{\cup} = \bigcup_m \mathcal{D}_{\mathrm{train}}^{(m)},
\end{equation*}
where $\mathcal{D}_{\mathrm{train}}^{(m)}$ denotes the public datasets documented in
model $m$'s pre-training corpus. M8 applies this shared set through the
donor-pool exclusion field:
\begin{equation*}
  \mathcal{S}_{\mathrm{M8}} =
  \mathcal{S}_{\mathrm{M1}}\bigl[
    \sigma\leftarrow 42,\;n\leftarrow 500,\;
    \mathcal{D}_{\mathrm{excl}}\leftarrow\mathcal{D}_{\mathrm{train}}^{\cup}\bigr].
\end{equation*}

\paragraph{M9, Screening simulation.}
We ran three independent cohorts of 500 cases each under geometrically
decaying prevalence calibrated to NLST incidence data, drawing hosts and
donors independently per round:
\begin{equation*}
  \pi_r = 0.04\cdot 0.7^{\,r},\quad r\in\{0,1,2\},\quad
  \text{giving }\pi_0{=}0.040,\;\pi_1{=}0.028,\;\pi_2{=}0.020.
\end{equation*}

\paragraph{M10, Multi-nodule context.}
We ran a mixed manifest with a 3:1 single-to-multi-nodule ratio, inserting
$N_{\mathrm{nod}}$ donors simultaneously into a single host for multi-nodule
cases:
\begin{equation*}
  \mathcal{S}_{\mathrm{M10}}:\quad
  \lfloor 0.75n\rfloor\;\text{single-nodule},\quad
  \lceil 0.25n\rceil\;\text{multi-nodule},\quad
  N_{\mathrm{nod}}\sim\mathcal{U}[2,5].
\end{equation*}

\paragraph{M11, Twin isolation ($\mathrm{Build}_{\mathrm{Iso}}$).}
We ran one specification per annotated nodule, 13{,}092 in total, re-inserting
each nodule into its own patient's anatomy for direct comparison against the
original real scan:
\begin{equation*}
  \mathrm{Build}_{\mathrm{Iso}}:\quad
  \forall\,(p,j)\in\Omega,\quad q\leftarrow p,\quad
  \text{donor}\leftarrow(p,j).
\end{equation*}

\paragraph{M12, Twin complete ($\mathrm{Build}_{\mathrm{Comp}}$).}
We ran one specification per host-eligible patient CT that also contained at
least one donor-eligible nodule, yielding 9{,}003 unique patient CTs embedding
13{,}092 nodules simultaneously, with all of each patient's annotated nodules
inserted at once into their own anatomy:
\begin{equation*}
  \mathrm{Build}_{\mathrm{Comp}}:\quad
  \forall\,p,\quad q\leftarrow p,\quad
  \text{donors}\leftarrow\{(p,1),\ldots,(p,N_p)\}\;\text{simultaneously}.
\end{equation*}
This is the only mode where overlapping nodule masks are expected by design.

\paragraph{M13, Twin cross ($\mathrm{Build}_{\mathrm{Cross}}$).}
We ran 9{,}498 specifications across three pairing policies and a
cross-dataset variant, transplanting donor nodules into foreign patient
anatomies under a controlled assignment matrix
$\mathbf{A}\in\{0,1\}^{|\mathcal{P}|\times|\mathcal{P}|}$:
\begin{equation*}
  \mathrm{Build}_{\mathrm{Cross}}:\quad q\neq p.
\end{equation*}
The three policies were: \texttt{one\_to\_one} (bijective, one host per donor
patient); \texttt{one\_to\_many\_hosts} ($N{=}3$ hosts per donor nodule,
enabling within-nodule host-anatomy variance estimation); and
\texttt{donor\_patient\_complete} (all nodules of patient $p$ transplanted
simultaneously into a single host $q\neq p$, preserving multi-nodule
topology while replacing host anatomy).
The cross-dataset variant ($D_{\mathrm{host}}\neq D_{\mathrm{donor}}$)
additionally transplants across acquisition conditions.

\begin{table}[htbp]
\caption{Design rationale for all 13 trial modes. \emph{Variable Isolated}:
the single experimental factor each mode varies while holding all others
fixed via $\mathcal{S}$.}
\label{tab:mode_rationale}
\centering\small
\setlength{\tabcolsep}{2pt}
\begin{tabular}{@{}clL{3.1cm}L{6.65cm}@{}}
\toprule
\textbf{\#} & \textbf{Mode}
  & \textbf{Variable Isolated}
  & \textbf{Design Rationale} \\
\midrule
M1  & Prevalence Control
  & None (reference)
  & Calibration baseline at NLST prior; all other modes are deviations
    from M1. Confirms reproducibility under fixed seed. \\[3pt]
M2  & Size Control
  & Nodule diameter $d$
  & Measures the size--sensitivity curve directly; removes size--covariate
    confounds present in pooled datasets. \\[3pt]
M3  & Location Isolation
  & Lobe $\ell$ (size fixed)
  & Fixes $d$ across all lobes so that accuracy differences are attributable
    to anatomical location alone, not size-frequency effects. \\[3pt]
M4  & Demographic Strat.
  & Sex $\times$ age stratum
  & Tests whether host-anatomy differences across demographic groups drive
    accuracy variation independent of nodule properties. \\[3pt]
M5  & Counterfactual Prev.
  & Prevalence $\pi$ (anatomy fixed via $\sigma$)
  & Holds host anatomy constant while sweeping $\pi$, enabling causal
    attribution of metric shifts to prevalence alone. \\[3pt]
M6  & Cross-Dataset
  & Nodule source $D_k$
  & Isolates nodule-source domain shift from host-anatomy variation by
    fixing one and varying the other across sub-cohorts. \\[3pt]
M7  & Bootstrap CI
  & Random seed $\sigma$
  & Quantifies metric sampling uncertainty via $B$ replicates; separates
    cohort-construction noise from genuine model differences. \\[3pt]
M8  & Algorithm Comparison
  & Model identity
  & Eliminates cohort-confounding in head-to-head comparisons by
    submitting one fixed-seed manifest to all models. \\[3pt]
M9  & Screening Simulation
  & Screening round $r$
  & Models longitudinal prevalence decay $\pi_r = \pi_0\cdot\gamma^r$
    to test whether ranking is preserved as incidence falls. \\[3pt]
M10 & Multi-Nodule Context
  & Nodule count $N_{\mathrm{nod}}$
  & Measures interference from concurrent lesions on per-target accuracy
    under a controlled single-to-multi ratio. \\
\cmidrule{1-4}
M11 & Twin Isolation
  & Host ($q{=}p$, native)
  & Re-inserts each nodule into its own anatomy; removes cross-patient
    confound and enables direct synthetic-vs-real comparison. \\[3pt]
M12 & Twin Complete
  & All $N_p$ nodules simultaneously
  & Reconstructs the full clinical acquisition; the only mode where
    overlapping masks are expected by design. \\[3pt]
M13 & Twin Cross
  & Host anatomy ($q{\neq}p$)
  & Transplants the same nodule into multiple foreign hosts to isolate
    the independent contribution of host anatomy to accuracy. \\
\bottomrule
\end{tabular}
\end{table}


\section{Generation Pipeline: Scale, Throughput, and Failure Analysis}
\label{app:generation}

\subsection{Mask Insertion Pipeline (Stage~3)}

\paragraph{Per-mode success rates.}
Insertion succeeded at 100.0\% for Modes 1--9, reflecting the effectiveness
of the snap-correction and re-placement logic for standard cohort designs.
Three modes exhibited sub-100\% rates: M10 (99.8\%, 1 failure from an
irresolvable multi-nodule overlap conflict), M11 and M12 (99.96\% each,
5 patients with undefined lobe assignments in the segmentation output),
and M13 (99.2\%, 78 failures from cross-patient placement conflicts
where no valid host position satisfied the pleural and overlap constraints). \textbf{Insertion compute.}
240.3 CPU-hours across all 13 modes,
with a mean of 17.6\,s per case (per-mode range 15.1--20.0\,s).

\paragraph{Snap correction.}
Overall 5.9\% of insertions (2{,}907/49{,}024) triggered snap correction,
with per-mode rates ranging from 4.5\% to 8.2\%.
The mean snap distance when triggered was ${\sim}$3.0 voxels
(per-mode range 2.5--3.45 voxels), indicating that boundary violations
were shallow and correctable without material disruption to the
intended insertion location.

\paragraph{Pleural distance.}
The enforced minimum pleural distance $\rho_{\min}{=}2.0$\,mm was
respected in $>$99\% of insertions.
Per-mode mean pleural distances ranged from 19.0 to 21.4\,mm
(medians 20.2--23.1\,mm), consistent with the peripheral nodule
distribution expected under the NLST template.
Rare violations down to 0.7\,mm occurred only in extreme-anatomy edge
cases in M10--M13.
\paragraph{Scale factor distribution.}
NLST-template modes (M1, M4, M5, M7, M8, M9) required isotropic
resampling ($\alpha \neq 1.0$) in 76--78\% of cases, with $\alpha$
ranging from 0.037 to 1.500.
Modes M2, M3, M6, and M11--M13 preserved native resolution
($\alpha{=}1.0$, 100\% of cases); no isotropic resampling was triggered in
those cohorts.

\paragraph{Overlap.}
Non-zero nodule-mask overlap occurred only in M10 (5.7\%, expected for
multi-nodule simultaneous insertion) and M12 (31.2\%, expected by
design: all native nodules of a patient are inserted simultaneously and
naturally co-located).
All other modes reported 0\% overlap.

\paragraph{Lobe distribution.}
The insertion lobe distribution closely matched the NLST template prior:
RUL 29.5\% (target 30\%), LUL 22.1\% (28\%), RLL 18.5\% (18\%),
LLL 14.6\% (16\%), RML 9.1\% (8\%).
The LUL under-representation relative to the template ($-$5.9 pp)
reflects the pool composition rather than a pipeline bias.


\begin{figure}[t]
  \centering
  \begin{subfigure}[t]{0.46\linewidth}
    \includegraphics[width=\linewidth]{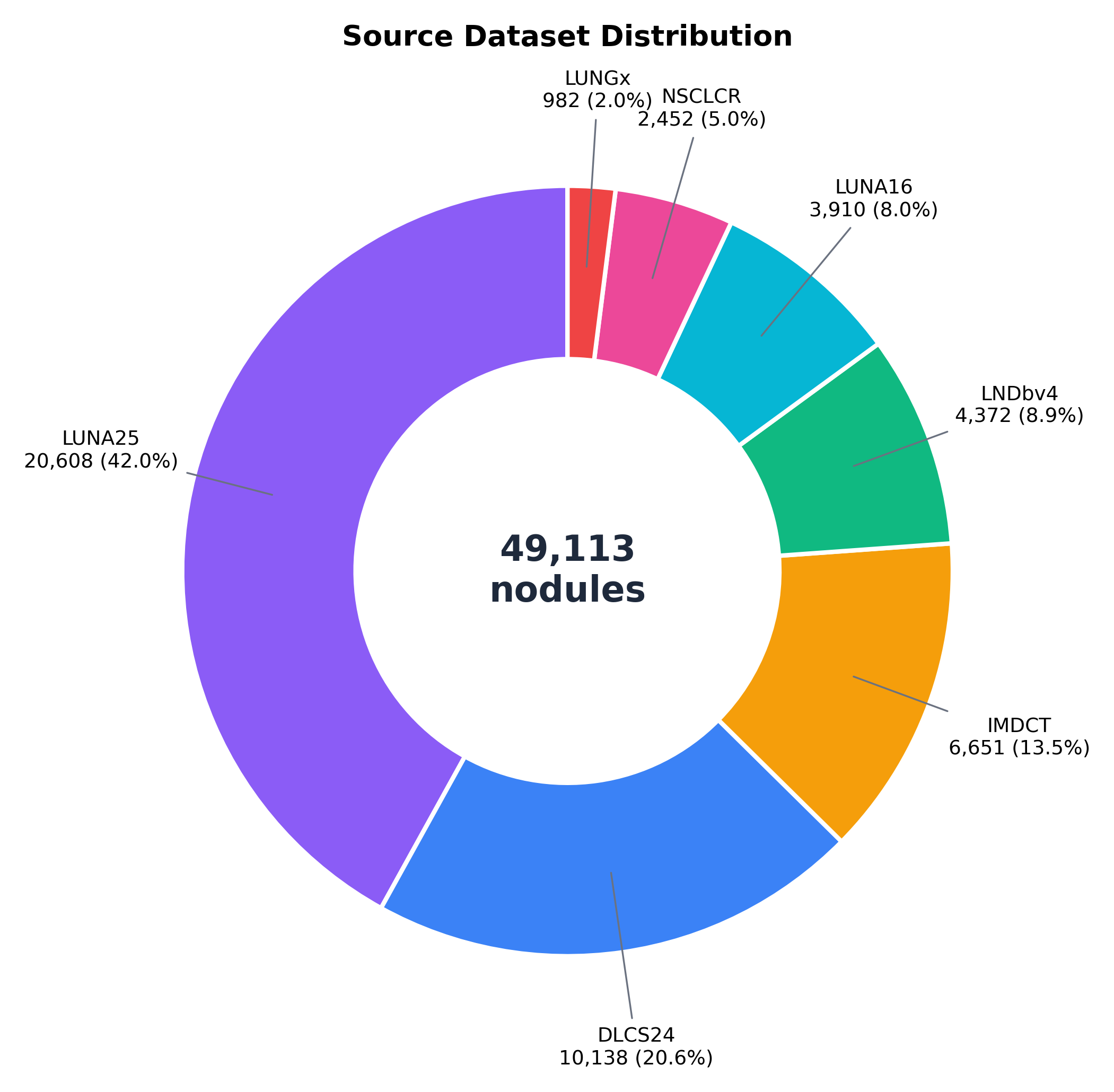}
    \caption{Source dataset distribution across the
             nodule pool (49{,}113 nodules total).
             LUNA25 dominates (42.0\%); six additional
             datasets span 2.0\%--20.6\%.}
    \label{fig:app_dataset_dist}
  \end{subfigure}
  \hfill
  \begin{subfigure}[t]{0.50\linewidth}
    \includegraphics[width=\linewidth]{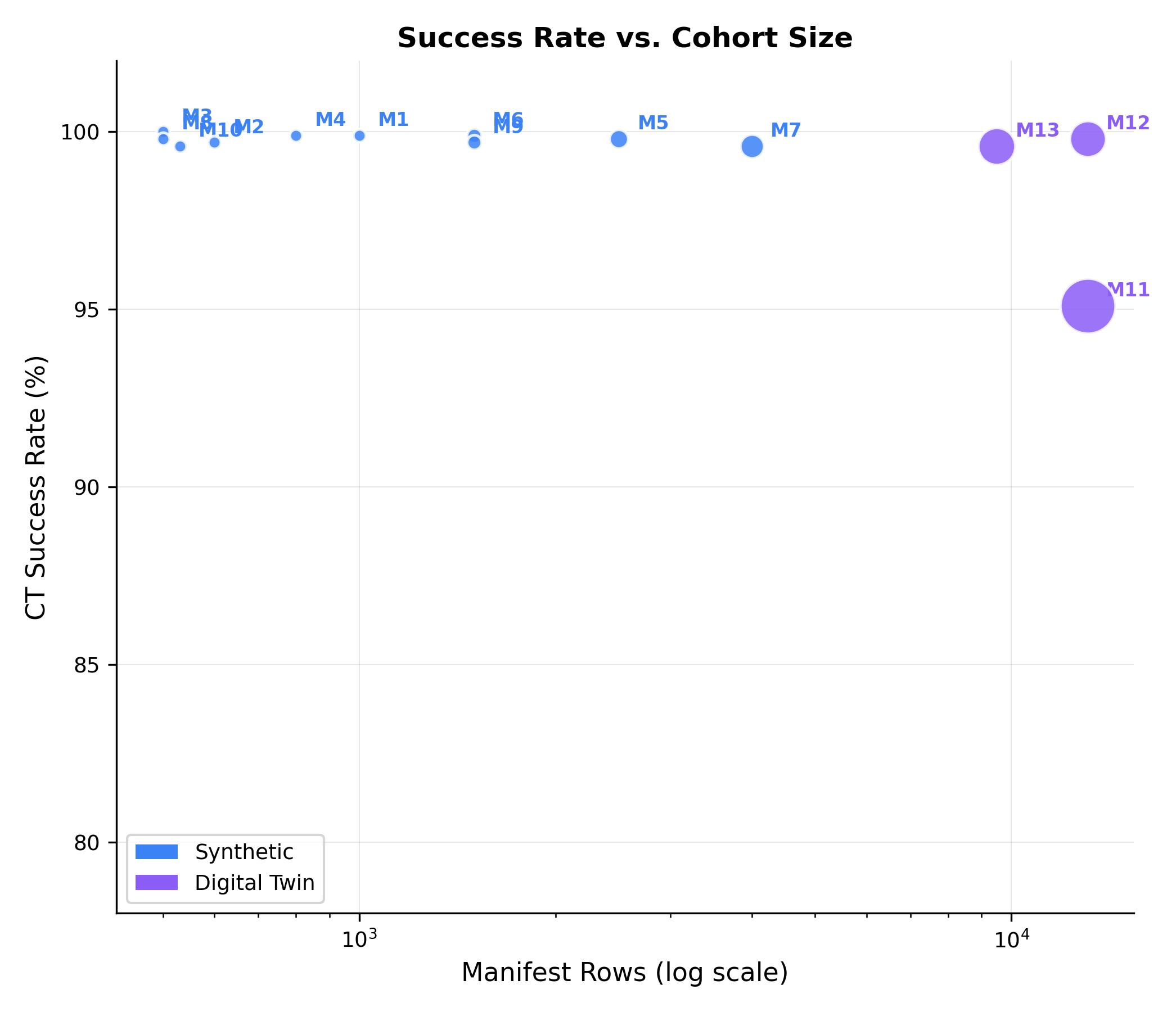}
    \caption{CT synthesis success rate vs.\ manifest
             size (log scale). Bubble area encodes
             $N_\text{CT}$. Synthetic modes (M1--M10,
             blue) achieve $\geq$99.7\%; M11 (digital
             twin, purple) reaches 95.1\% post-fix.}
    \label{fig:app_success_scatter}
  \end{subfigure}
  \caption{\textbf{Nodule pool composition and per-mode generation success.}}
  \label{fig:app_pool_and_success}
\end{figure}

\subsection{CT Synthesis Success by Mode (Stage~4)}

Synthetic cohort modes M1--M10 achieved a collective yield of \textbf{99.7\%}
(31 failures across all ten modes combined; Table~\ref{tab:app_generation}),
demonstrating that the pipeline operates at near-perfect reliability for standard cohort designs.
Digital twin modes M11--M13 collectively yielded \textbf{97.7\%}
(30{,}807/31{,}510). The dominant contributor to the overall 1.6\% gap from perfect yield is M11,
which is analysed in detail in \S\ref{app:mode11_failure}. 

Digital twin modes consumed 72.5\% of total GPU compute despite representing
69.5\% of successfully generated CTs (Table~\ref{tab:app_generation}), reflecting the larger average volume
sizes in the digital twin pool (native patient CTs, vs.\ the NLST-matched
host volumes in synthetic cohort modes).
Across mode families, mean synthesis throughput remained tightly clustered at
${\sim}$1.05 min/case for synthetic cohort modes, ${\sim}$0.88 for M11,
1.07 for M12, and 1.08 for M13, confirming that synthesis time is
volume-size-driven rather than mode-specific.

\begin{table}[htbp]
\caption{Per-mode NodMAISI synthesis success and GPU compute.
Modes 1--10 are synthetic cohort; Modes 11--13 are digital twin.
GPU hours for Modes 2--3 are $<$0.5\,h and reported as ${\approx}$0.}
\label{tab:app_generation}
\centering\small
\setlength{\tabcolsep}{4pt}
\begin{tabular}{@{}clrrrrc@{}}
\toprule
\textbf{\#} & \textbf{Mode}
  & \textbf{Attempted} & \textbf{Generated} & \textbf{Failed}
  & \textbf{Yield (\%)} & \textbf{GPU-h} \\
\midrule
M1  & Prevalence Control    & 1{,}000  &   999    &    1 & 99.9 &   0.3 \\
M2  & Size Curve            &   600    &   600    &    0 & 100.0 & ${\approx}$0 \\
M3  & Location Isolation    &   500    &   500    &    0 & 100.0 & ${\approx}$0 \\
M4  & Demographic Strat.    &   800    &   799    &    1 & 99.9 &  13.9 \\
M5  & Counterfactual Prev.  & 2{,}500  & 2{,}496  &    4 & 99.8 &  43.4 \\
M6  & Cross-Dataset         & 1{,}500  & 1{,}499  &    1 & 99.9 &  26.7 \\
M7  & Bootstrap CI          & 4{,}000  & 3{,}984  &   16 & 99.6 &  69.9 \\
M8  & Algorithm Comparison  &   500    &   499    &    1 & 99.8 &   8.9 \\
M9  & Screening Simulation  & 1{,}500  & 1{,}495  &    5 & 99.7 &  26.3 \\
M10 & Multi-Nodule Context  &   500    &   498    &    2 & 99.6 &   9.1 \\
\cmidrule{1-7}
M11 & Twin Isolation        & 13{,}087 & 12{,}448 &  639 & 95.1 & ${\sim}$373 \\
M12 & Twin Complete         &  9{,}003 &  8{,}981 &   22 & 99.8 & 161.2 \\
M13 & Twin Cross            &  9{,}420 &  9{,}378 &   42 & 99.6 & 169.4 \\
\cmidrule{1-7}
    & \textbf{Total}
  & \textbf{44{,}910} & \textbf{44{,}176} & \textbf{734}
  & \textbf{98.4} & ${\sim}$\textbf{900} \\
\bottomrule
\end{tabular}
\end{table}

\subsection{Failure Investigation}
\label{app:mode11_failure}

M11 is the only mode with a materially elevated failure rate. Of the 639 failures (4.9\% of M11), 585 (91.5\%) are
NodMAISI inference-level rejections with no output, and 54 (8.5\%) are
pre-synthesis QC rejections (empty label-23 anatomy mask).
Both failure types are concentrated in volumes with extreme aspect ratios
that push NodMAISI beyond its training distribution.

Dataset-specific M11 success rates varied substantially across the seven source datasets (Table~\ref{tab:app_m11_dataset}),
reflecting differences in CT acquisition protocols and volume dimensions:

\begin{table}[htbp]
\caption{Mode~11 (Twin Isolation) CT generation success rate
by source dataset. LUNA25 and NSCLCR exhibit the lowest yields, driven
by their prevalence of high-slice-count volumes.}
\label{tab:app_m11_dataset}
\centering\small
\setlength{\tabcolsep}{5pt}
\begin{tabular}{@{}lrrr@{}}
\toprule
\textbf{Dataset} & \textbf{Specs} & \textbf{Synthetic CTs} & \textbf{Yield (\%)} \\
\midrule
DLCS24  & 2{,}473 & ${\sim}$2{,}470 & ${\sim}$99.9 \\
LUNA16  & 1{,}184 & ${\sim}$1{,}180 & ${\sim}$99.7 \\
LUNGx   &    83   &    ${\sim}$83   & ${\sim}$100.0 \\
LNDbv4  &   743   &  ${\sim}$700    & ${\sim}$94.2 \\
IMDCT   & 2{,}032 & ${\sim}$2{,}020 & ${\sim}$99.4 \\
LUNA25  & 6{,}156 & ${\sim}$5{,}640 & ${\sim}$91.6 \\
NSCLCR  &   421   &  ${\sim}$355    & ${\sim}$84.3 \\
\midrule
\textbf{Total} & \textbf{13{,}092} & \textbf{12{,}448} & \textbf{95.1} \\
\bottomrule
\end{tabular}
\end{table}

LUNA25 and NSCLCR exhibit the lowest yields (${\sim}$91.6\% and
${\sim}$84.3\% respectively), attributable to their prevalence of
high-slice-count volumes (many $>$400 slices) and staging-context
acquisitions with larger field-of-view.
These datasets contribute 6{,}577 of the 13{,}092 M11 specifications
(50.2\%) and account for the majority of residual failures.
DLCS24, LUNA16, LUNGx, and IMDCT achieve near-perfect yields.

\section{Synthetic CT Quality Assessment}
\label{app:quality}

\subsection{Evaluation Protocol and Metric Selection}

\paragraph{Cohort-level distributional metrics (FID, KID).}
Fréchet Inception Distance (FID)~\citep{Heusel2017GANs} and Kernel Inception Distance (KID) measure
the distance between real and synthetic feature distributions without requiring
case-level correspondence.
Features were extracted using a RadImageNet-pretrained ResNet50 (2048-dim),
following the MAISI evaluation protocol~\citep{chen2024maisi,tushar2025nodmaisi}: full-volume NIfTI
inputs are resampled to 1.0\,mm$^3$ isotropic, padded/cropped to
512$\times$512$\times$256, and windowed to [$-$1000, 1000]\,HU; the centre
40\% of slices from each of three orthogonal planes (XY/axial,
YZ/sagittal, ZX/coronal) are then extracted and features averaged per volume.
For all comparisons, we used 100 volumes per cohort.
KID is computed from the same feature tensors using a cubic polynomial kernel
(unbiased estimator; reported alongside FID in Appendix~\ref{app:r2r}; FID is the
primary metric throughout).

\paragraph{Per-case intensity fidelity metrics.}

Because NodMAISI is a rectified flow model and uses \texttt{noise\_factor}\,=\,1.0, the entire CT is
re-generated from Gaussian noise conditioned on the anatomy-segmentation mask
($\mathbf{m}^{\mathrm{ins}}$). We instead computed the following per-case intensity-distribution metrics
between each synthetic CT and its host:

\textbf{Histogram intersection (HI):} overlap between normalised HU
    histograms (256-bin, body voxels $>$\,$-$950\,HU); range $[0,1]$, higher is
    better.

\textbf{KL divergence} ($D_\text{KL}(\text{host}\,\|\,\text{synth})$):
    lower is better; 0 = identical distributions.

\textbf{Wasserstein-1 distance} (earth mover's distance in HU):
    lower is better; measures HU distribution shift.

\textbf{Nodule ROI HU statistics:} mean, std, min, max HU extracted from the insertion mask (label 23) to verify clinically plausible tissue densities. Per-case metrics were computed for $n\,=\,36$--50 host--synthetic pairs per mode
(607 total), drawn as a random subset matched on subject ID.

\paragraph{Cohort-level histogram similarity.}
Separately from per-case metrics, we compared the global HU distributions of
50 synthetic vs.\ 50 real volumes (LUNA25 and DLCS24 as reference datasets)
using 256-bin histograms over [$-$1024, 3071]\,HU.
This measures whether the \emph{cohort} preserves realistic HU distributions,
complementing FID (which operates in a learned feature space).

\subsection{Real-to-Real Baseline}
\label{app:r2r}

\begin{figure}[htbp]
  \centering
  \begin{subfigure}[t]{0.48\linewidth}
    \includegraphics[width=\linewidth]{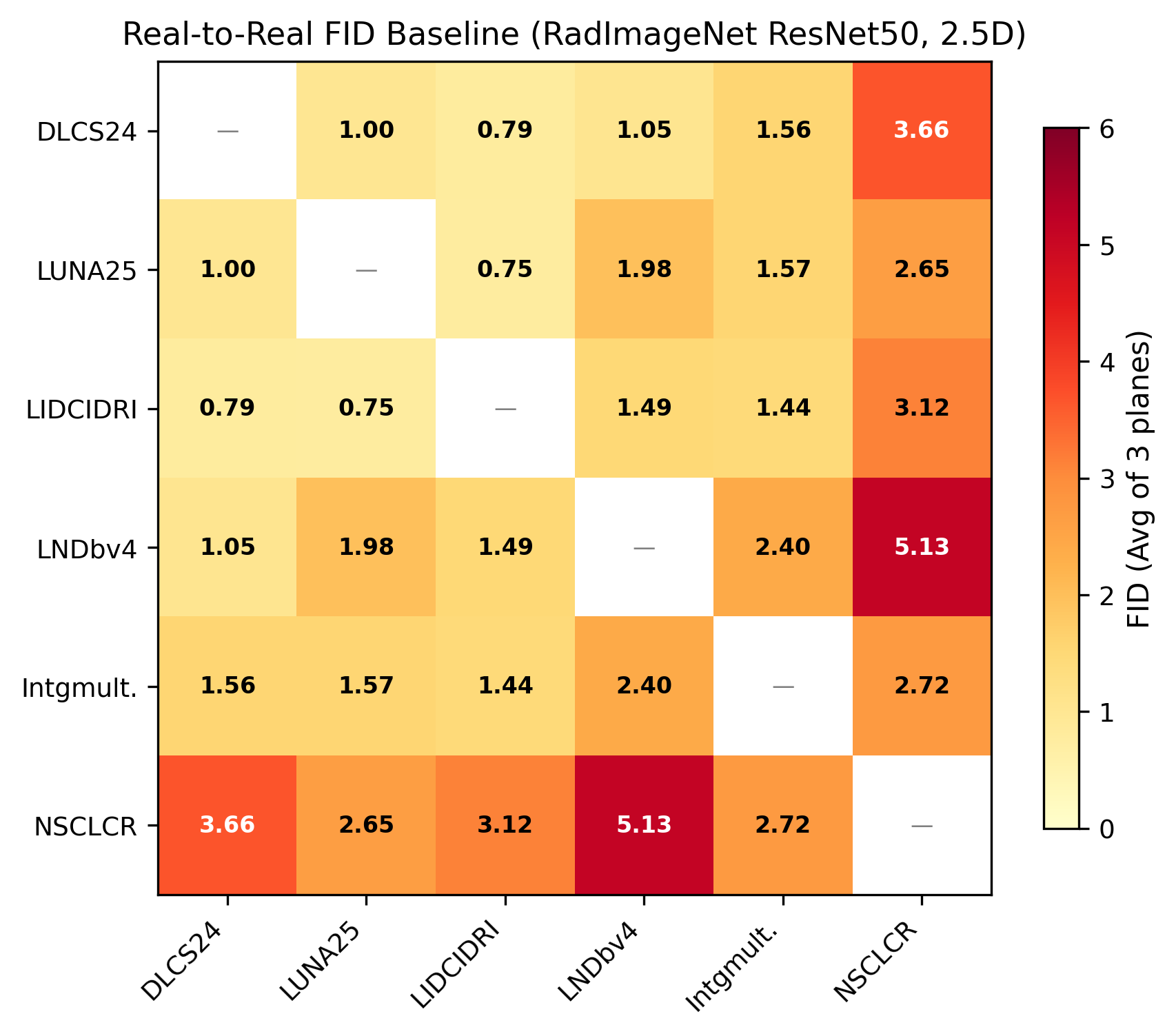}
    \caption{Real-to-real FID heatmap (15 pairwise comparisons, six datasets).
    The dashed line indicates the median (1.57); shaded band = IQR [1.05, 2.72].}
    \label{fig:app_r2r_heatmap}
  \end{subfigure}
  \hfill
  \begin{subfigure}[t]{0.48\linewidth}
    \includegraphics[width=\linewidth]{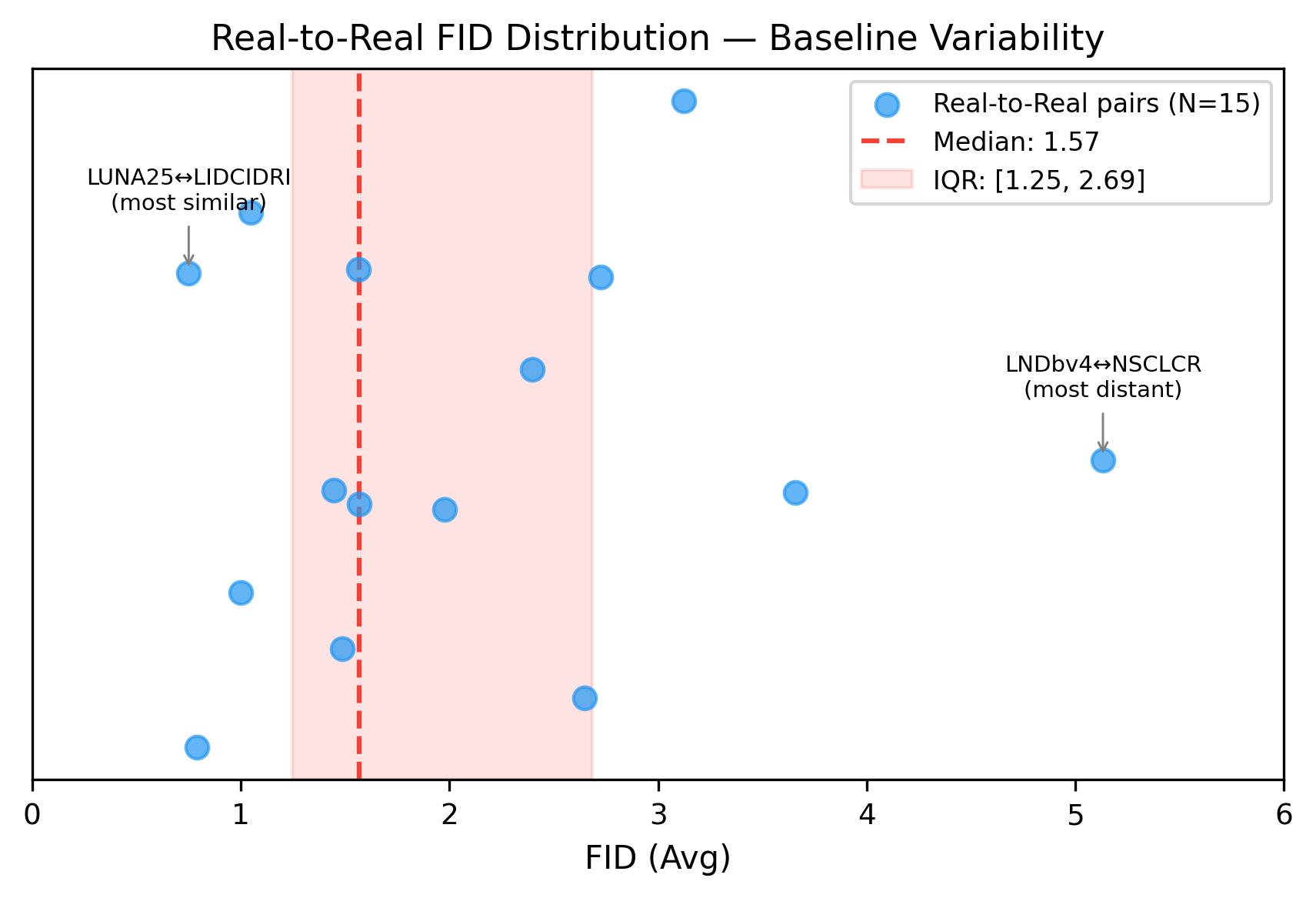}
    \caption{R2R FID distribution (left) vs.\ real-to-synthetic FID by mode
    (right), directly comparing synthetic quality against natural inter-dataset
    variability.}
    \label{fig:app_r2r_vs_synth}
  \end{subfigure}
  \caption{\textbf{Real-to-real FID baseline.}}
  \label{fig:app_r2r}
\end{figure}

Table~\ref{tab:app_r2r} and Figure~\ref{fig:app_r2r_heatmap} report FID across
all 15 pairwise comparisons of six real clinical CT datasets.

\begin{table}[htbp]
\caption[Real-to-real FID baseline. Pairwise FID across six real clinical CT datasets.]{\textbf{Real-to-real FID baseline.}
Pairwise FID$_\text{avg}$ (mean of XY, YZ, ZX planes) across six clinical CT datasets
(100 volumes per dataset, RadImageNet ResNet50, 2.5D). Lower values indicate
greater similarity and define the natural inter-dataset variability baseline
for synthetic CT quality.}
\label{tab:app_r2r}
\centering\small
\setlength{\tabcolsep}{5pt}
\begin{tabular}{@{}llr@{}}
\toprule
\textbf{Dataset A} & \textbf{Dataset B} & \textbf{FID$_\text{avg}$}$\downarrow$ \\
\midrule
LUNA25          & LUNA16          & 0.750 \\
DLCS24          & LUNA16          & 0.790 \\
DLCS24          & LUNA25            & 1.000 \\
DLCS24          & LNDbv4            & 1.046 \\
IMDCT  & LUNA16          & 1.444 \\
LNDbv4          & LUNA16          & 1.487 \\
DLCS24          & IMDCT    & 1.564 \quad \emph{(median)} \\
LUNA25          & IMDCT    & 1.568 \\
LUNA25          & LNDbv4            & 1.977 \\
LNDbv4          & IMDCT    & 2.396 \\
LUNA25          & NSCLCR    & 2.647 \\
NSCLCR  & IMDCT    & 2.724 \\
NSCLCR  & LUNA16          & 3.122 \\
DLCS24          & NSCLCR    & 3.658 \\
LNDbv4          & NSCLCR    & 5.132 \\
\midrule
\multicolumn{2}{@{}l}{\emph{Summary: mean 2.09\,$\pm$\,1.17, median 1.57,
  IQR [1.05, 2.72]}} & \\
\bottomrule
\end{tabular}
\end{table}

DLCS24, LUNA25, and LUNA16 form the tightest cluster in Table~\ref{tab:app_r2r}, with pairwise FID values of 0.75--1.00.
NSCLCR, a dataset of surgically confirmed malignant nodules, is consistently the most distant cohort in this comparison
(FID 2.6--5.1).
Any synthetic mode with FID within the range [0.75, 5.13] achieves
distributional quality within the span of natural inter-dataset variability.

\subsection{Per-Mode Quality Results}
\label{app:quality_full}

\begin{table}[t]
\caption{\textbf{Synthetic data quality across all 13 trial modes.}
FID$_\text{avg}$ = FID averaged over XY, YZ, ZX planes (2.5D, 1\,mm$^3$, 100 volumes vs.\ LUNA25); lower is
better. Per-case HI = histogram intersection between each synthetic CT and its
host (mean\,$\pm$\,std, $n$\,=\,36--50 pairs per mode); higher is better.
R2R: real-to-real baseline across 15 cross-dataset pairs
(FID median\,=\,1.57, IQR [1.05, 2.72]; HI\,=\,0.927 for LUNA25\,$\leftrightarrow$\,DLCS24).}

\label{tab:quality_full}
\centering\small
\setlength{\tabcolsep}{3pt}
\begin{tabular}{@{}lrcccccc@{}}
\toprule
\textbf{Mode} & \textbf{$n$} &
\textbf{FID$_\text{avg}$}$\downarrow$ &
\textbf{Per-case HI}$\uparrow$ &
\textbf{KL div.}$\downarrow$ &
\textbf{W1 (HU)}$\downarrow$ &
\textbf{Cohort HI}$\uparrow$ &
\textbf{Nod.\ HU}\\
\midrule
M1  & 50 & 1.77 & .843\,$\pm$\,.054 & .087\,$\pm$\,.058 & 80.7\,$\pm$\,43.2 & .827 & $-$647 \\
M2  & 36 & 2.03 & .829\,$\pm$\,.070 & .107\,$\pm$\,.094 & 81.8\,$\pm$\,42.0 & .788 & $-$560 \\
M3  & 43 & 2.07 & .816\,$\pm$\,.066 & .120\,$\pm$\,.081 & 105.3\,$\pm$\,54.5 & .772 & $-$534 \\
M4  & 50 & 1.68 & .845\,$\pm$\,.056 & .087\,$\pm$\,.066 & 75.9\,$\pm$\,37.9 & .834 & $-$604 \\
M5  & 50 & 1.60 & .845\,$\pm$\,.058 & .086\,$\pm$\,.063 & 75.7\,$\pm$\,41.2 & .838 & $-$564 \\
M6  & 37 & 2.13 & .827\,$\pm$\,.072 & .113\,$\pm$\,.097 & 90.1\,$\pm$\,54.3 & .772 & $-$583 \\
M7  & 50 & 1.31 & .840\,$\pm$\,.058 & .091\,$\pm$\,.068 & 76.0\,$\pm$\,40.7 & .837 & $-$582 \\
M8  & 50 & \textbf{1.29} & \textbf{.847\,$\pm$\,.057} & \textbf{.084\,$\pm$\,.063} & 77.5\,$\pm$\,41.4 & .838 & $-$601 \\
M9  & 50 & 1.66 & .841\,$\pm$\,.059 & .089\,$\pm$\,.062 & 76.9\,$\pm$\,41.1 & .840 & $-$573 \\
M10 & 41 & 1.96 & .844\,$\pm$\,.050 & .091\,$\pm$\,.063 & 78.2\,$\pm$\,40.9 & .797 & $-$585 \\
M11 & 50 & 1.74 & .843\,$\pm$\,.041 & .083\,$\pm$\,.041 & 89.3\,$\pm$\,37.2 & .794 & $-$598 \\
M12 & 50 & 1.82 & .843\,$\pm$\,.046 & .083\,$\pm$\,.044 & 87.9\,$\pm$\,42.4 & .794 & $-$582 \\
M13 & 50 & 2.22 & .841\,$\pm$\,.051 & .086\,$\pm$\,.049 & 89.0\,$\pm$\,41.2 & .792 & $-$510 \\
\midrule
\emph{R2R}$^*$ & --- &
  \emph{1.57\,[1.05,\,2.72]} & \emph{---} &
  \emph{0.017} & \emph{14.1} & \emph{.927} & --- \\
\bottomrule
\end{tabular}
\\[2pt]
{\footnotesize $^*$: \emph{LUNA25\,$\leftrightarrow$\,DLCS24}.}
\end{table}

Table~\ref{tab:quality_full} reports all six quality metrics for all 13
trial modes. \textbf{Key observations:}

(1)~\textbf{All modes within R2R IQR.}
FID$_\text{avg}$ ranges 1.29--2.22, with all 13 modes within the R2R IQR [1.05, 2.72].
Modes M7 (1.31) and M8 (1.29) achieve FID below the R2R median (1.57),
matching the similarity of the closest real-to-real dataset pairs.

(2)~\textbf{Attribute-manipulation modes show higher FID and lower cohort HI.}
Modes M2 (size variation), M3 (location shifting), and M6 (cross-dataset
host sampling) exhibit the highest FID (2.03--2.13) and lowest cohort HI
(0.77--0.79), reflecting their deliberate distributional shifts rather than
synthesis artefacts.
Cross-metric consistency is confirmed: FID rank order is highly correlated
with cohort HI rank order ($r > 0.90$).

(3)~\textbf{Per-case HI $>$ 0.80 for all modes.}
The grand mean per-case HI of 0.838 across 607 pairs (range 0.816--0.847)
indicates that NodMAISI reliably preserves the host CT's intensity
characteristics.
Modes M2, M3, and M6 show the lowest per-case HI (0.816--0.829) and highest
Wasserstein-1 distances (81.8--105.3\,HU), consistent with the larger
distributional shifts introduced by those modes.

(4)~\textbf{Digital twin modes: consistent quality, elevated W1.}
Modes M11--M13 achieve per-case HI of 0.841--0.843 (comparable to the best
insertion modes) with notably lower variance (std 0.041--0.051 vs.\
0.054--0.072 for insertion modes), reflecting the more homogeneous anatomy
of digital twin hosts.
However, W1 distances (87.9--89.3\,HU) are higher than the best insertion
modes (75.7--80.7\,HU), attributable to systematic HU offsets from processing
diverse host datasets in whole-body re-generation.

(5)~\textbf{Nodule HU is clinically plausible across all modes.}
Generated nodule mean HU ranges from $-$510 (M13) to $-$647 (M1),
consistent with ground-glass and part-solid pulmonary nodule morphologies.
M13 (digital twin cross) exhibits the highest (least negative) nodule HU,
reflecting the diversity of anatomy into which nodules are cross-inserted.

\begin{figure}[t]
  \centering
  \begin{subfigure}[t]{0.48\linewidth}
    \includegraphics[width=\linewidth]{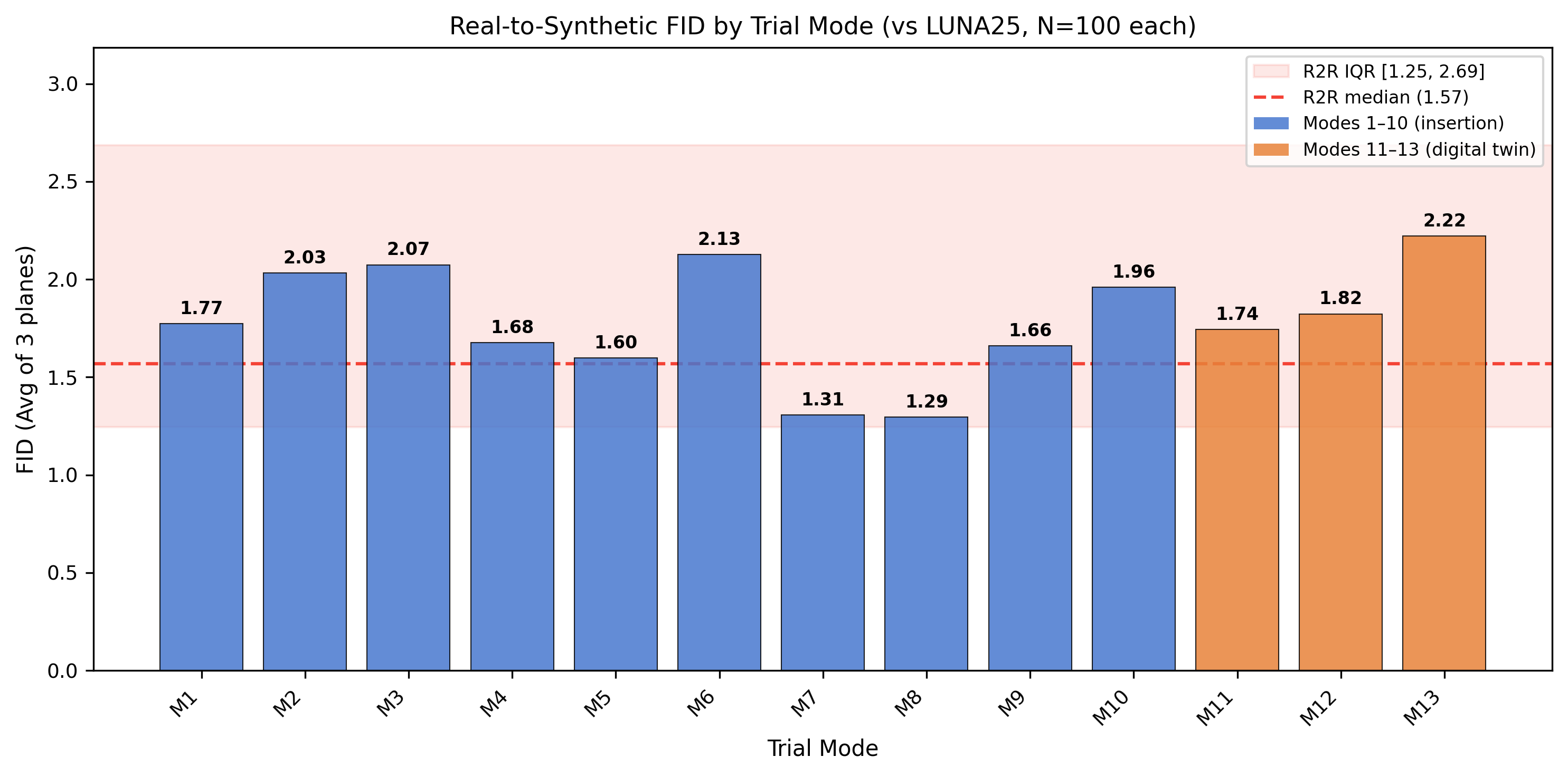}
    \caption{FID$_\text{avg}$ per mode vs.\ LUNA25.
    Dashed line = R2R median (1.57); shaded band = R2R IQR [1.05, 2.72].
    Blue = insertion modes (M1--M10), orange = digital twin modes (M11--M13).}
    \label{fig:app_fid_by_mode}
  \end{subfigure}
  \hfill
  \begin{subfigure}[t]{0.48\linewidth}
    \includegraphics[width=\linewidth]{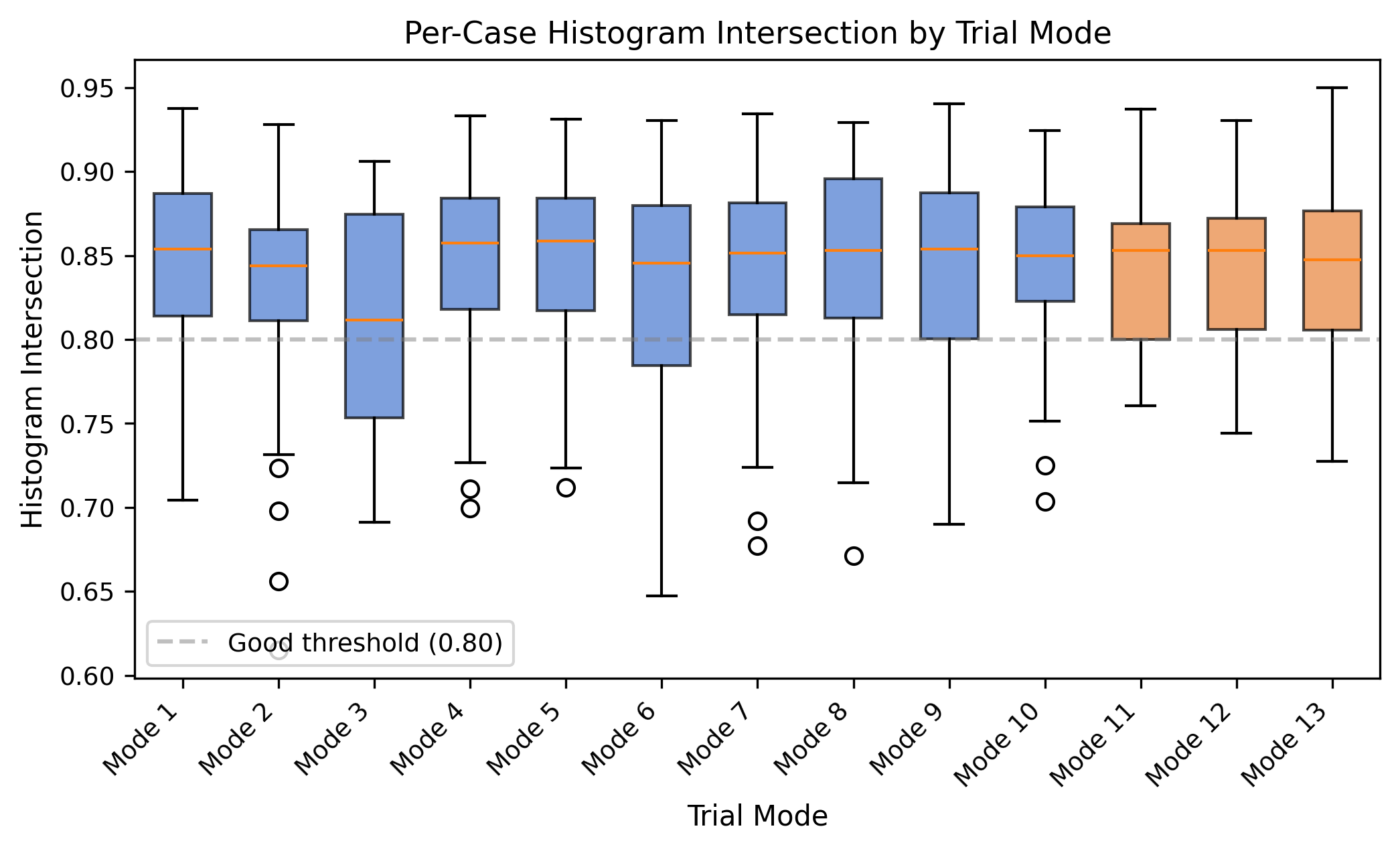}
    \caption{Per-case histogram intersection (host vs.\ synthetic) across all
    13 modes ($n = 36$--50 per mode). Dashed line = 0.80 threshold.
    Violin width encodes density.}
    \label{fig:app_hi_by_mode}
  \end{subfigure}

  \vspace{4pt}

  \begin{subfigure}[t]{0.48\linewidth}
    \includegraphics[width=\linewidth]{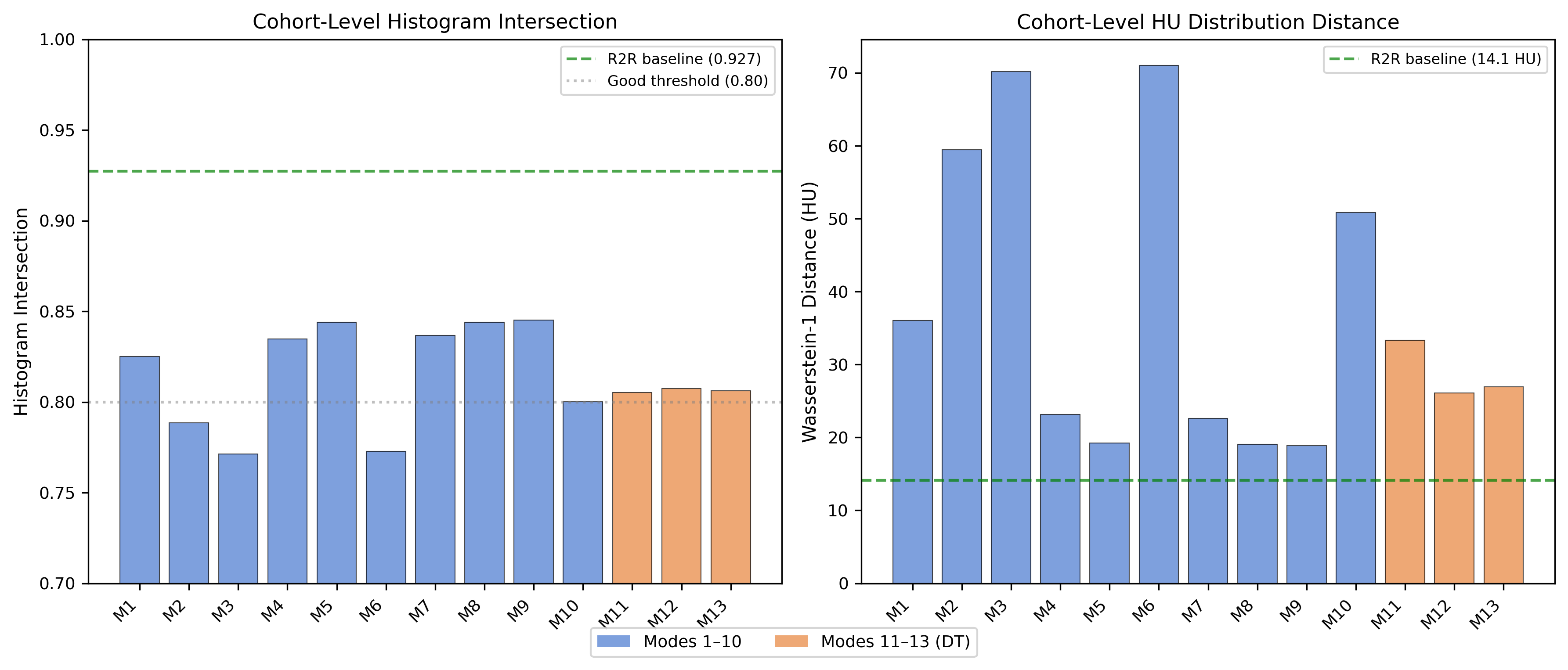}
    \caption{Cohort-level histogram intersection (left) and Wasserstein-1
    distance in HU (right) for all 13 modes vs.\ LUNA25 (50 volumes each).
    R2R baseline (LUNA25\,$\leftrightarrow$\,DLCS24): HI\,=\,0.927,
    W1\,=\,14.1\,HU.}
    \label{fig:app_cohort_hi}
  \end{subfigure}
  \hfill
  \begin{subfigure}[t]{0.48\linewidth}
    \includegraphics[width=\linewidth]{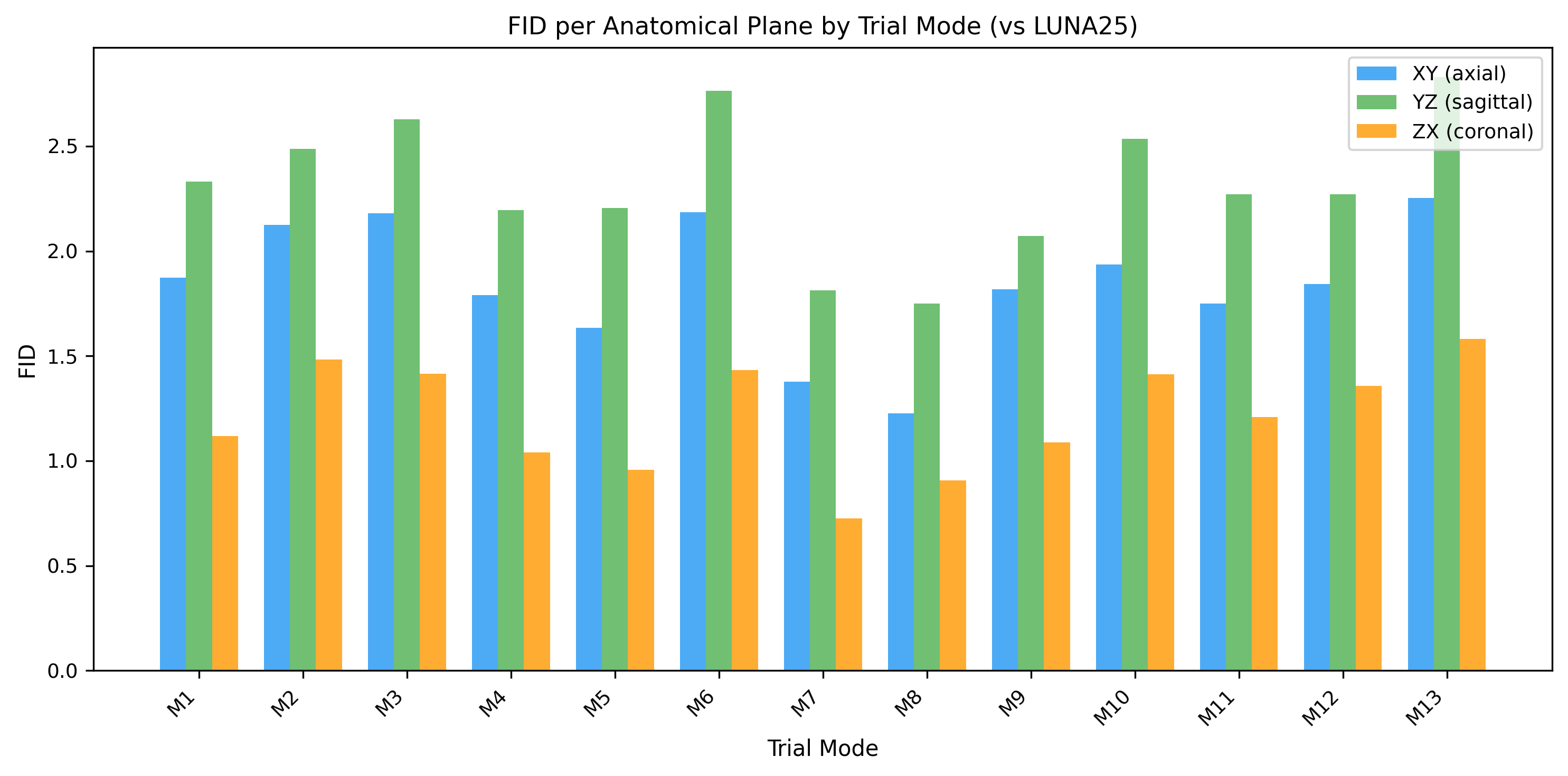}
    \caption{FID decomposed by orthogonal plane (XY/axial, YZ/sagittal,
    ZX/coronal) for all 13 modes. Anisotropy (ZX\,$>$\,XY) is observed
    across all modes, consistent with the 2.5D-to-3D resolution difference
    in the feature extraction protocol.}
    \label{fig:app_fid_planes}
  \end{subfigure}

  \caption{\textbf{Synthetic CT quality metrics across all 13 trial modes.}
  (a)~FID by mode relative to the real-to-real baseline; all modes fall within
  the R2R IQR.
  (b)~Per-case intensity distribution fidelity; all modes exceed HI\,$>$\,0.80.
  (c)~Cohort-level HU distribution similarity; modes with deliberate
  attribute manipulation (M2, M3, M6) and digital twin cross-anatomy (M13)
  show the largest distributional shifts.
  (d)~Plane-wise FID decomposition; ZX plane consistently exhibits higher
  FID, reflecting through-plane resolution anisotropy.}
  \label{fig:app_quality}
\end{figure}

\paragraph{Limitations.}
FID operates on 2D slice representations (2.5D, centre 40\% of slices) and assumes Gaussian
feature distributions, potentially missing 3D volumetric artefacts and multi-modal structure;
it is also a global metric, insensitive to localised nodule-level defects that occupy
$<$1\% of slices.
Per-case histogram metrics capture global intensity distributions but discard spatial information,
and the elevated Wasserstein-1 distances in digital twin modes (88--89\,HU) reflect FOV
differences rather than synthesis artefacts. Finally, modes whose host CTs originate predominantly from LUNA25 naturally achieve lower
FID against that reference; cohort HI computed against both LUNA25 and DLCS24
($r > 0.99$) confirms that reference choice does not affect mode rank ordering.

\section{VLM Evaluation}
\label{app:vlm}

All inference is \textbf{zero-shot, single-prompt}. No fine-tuning, no few-shot examples.

\paragraph{Models.}
Three models spanning contrastive and generative paradigms: \textbf{BiomedCLIP}~\citep{biomedclip2023}: ViT-B/16 + PubMedBERT $\sim$150\,M parameters, contrastive image--text matching.
Inputs: single axial lung-window slice; inference via cosine similarity to class-label text prompts. \textbf{LLaVA-Med}~\citep{llavamed2024}: LLaMA-2 7\,B + CLIP ViT-L, generative instruction-tuned, single axial slice. 
\textbf{MedGemma}~\citep{medgemma2024}: Gemma 4\,B + SigLIP, generative, 3-channel axial stack (multi-slice context window).

\paragraph{Slice extraction.}
For each nodule sample, three consecutive axial slices are extracted at the nodule centroid ($z$) and its immediate neighbours ($z{\pm}1$), rendered at lung window
(W\,=\,1500, L\,=\,$-$600\,HU) and saved as PNG.
Two preprocessing profiles are generated per sample: three separate 224$\times$224\,px PNGs for BiomedCLIP and
LLaVA-Med (\texttt{lung\_axial})~\citep{biomedclip2023}~\citep{llavamed2024}; and 
a single 448$\times$448\,px 3-channel composite stacking $z{-}1$, $z$, $z{+}1$ as RGB channels.
Slice localisation succeeds via the input mask centroid for 97.9\%
of samples~\citep{medgemma2024}.

\paragraph{Spatial guidance overlays.}
BBox: axis-aligned green rectangle from the projected 3D segmentation mask.
Contour: red boundary contour tracing the nodule margin on the centroid slice.
BBox+Contour: both overlays combined.
Overlay generation succeeds for 97.5\% of all samples
(1,336 failures concentrated in M12 digital-twin complete, where dense multi-organ
masks produce degenerate projection geometries; these samples are evaluated
under the plain condition only).

\paragraph{Task prompts and decoding.}
BiomedCLIP classifies via cosine similarity between the image embedding and a set
of class-specific text descriptors (e.g., ``A CT scan image showing a pulmonary nodule
in the right upper lobe''); the class with the highest similarity is the prediction.
LLaVA-Med and MedGemma generate free text from a structured task prompt followed by
``Respond with exactly one of: [class labels]''; outputs are parsed to the nearest
valid class by fuzzy string match.
All inference is greedy (temperature\,=\,0), zero-shot, single-prompt.

Three tasks applied uniformly to all cases:
\begin{itemize}
  \item \textbf{Presence detection} (binary): ``Is a lung nodule visible in
    this CT slice?'' — Yes / No.
  \item \textbf{Lobe localisation} (5-class): ``In which pulmonary lobe is
    the lung nodule located?'' — RUL / RML / RLL / LUL / LLL.
  \item \textbf{Size classification} (4-class):
    ``Estimate the size of the lung nodule.'' —
    $<$6\,mm / 6--10\,mm / 10--20\,mm / $>$20\,mm.
\end{itemize}

Table~\ref{tab:app_vlm_prompts} shows representative prompts for each task
and model family. Contrastive models (BiomedCLIP) use class-label strings
matched against the image embedding; generative models (LLaVA-Med,
MedGemma) receive the image and a structured instruction prompt.

\begin{table}[htbp]
\caption{\textbf{Representative prompts by task and model type.}
Spatial guidance conditions append a bracketed cue to the instruction.}
\label{tab:app_vlm_prompts}
\centering\small
\setlength{\tabcolsep}{4pt}
\begin{tabular}{@{}llp{8cm}@{}}
\toprule
\textbf{Task} & \textbf{Model type} & \textbf{Prompt / label set} \\
\midrule
Presence & Contrastive &
  Positive: ``a CT scan showing a lung nodule'';
  Negative: ``a CT scan with no visible lung nodule'' \\
Presence & Generative &
  ``Does this CT slice show a lung nodule? Answer Yes or No.
  \textit{[BBox: A bounding box marks the region of interest.]}'' \\
\midrule
Lobe & Contrastive &
  ``a lung nodule in the right upper lobe'' (×5, one per lobe) \\
Lobe & Generative &
  ``In which pulmonary lobe is the lung nodule located?
  Answer one of: RUL, RML, RLL, LUL, LLL.
  \textit{[Contour: A red contour marks the nodule boundary.]}'' \\
\midrule
Size & Contrastive &
  ``a lung nodule smaller than 6mm''; ``6 to 10mm'';
  ``10 to 20mm''; ``larger than 20mm'' \\
Size & Generative &
  ``Estimate the size of the lung nodule. Answer one of:
  $<$6mm, 6--10mm, 10--20mm, $>$20mm.
  \textit{[BBox+Contour: Both a bounding box and red contour mark the nodule.]}'' \\
\bottomrule
\end{tabular}
\end{table}



\subsection{Synthetic--Real Domain Gap}
\label{app:domain_gap}

\begin{figure}[htbp]
  \centering
  \includegraphics[width=0.72\linewidth]{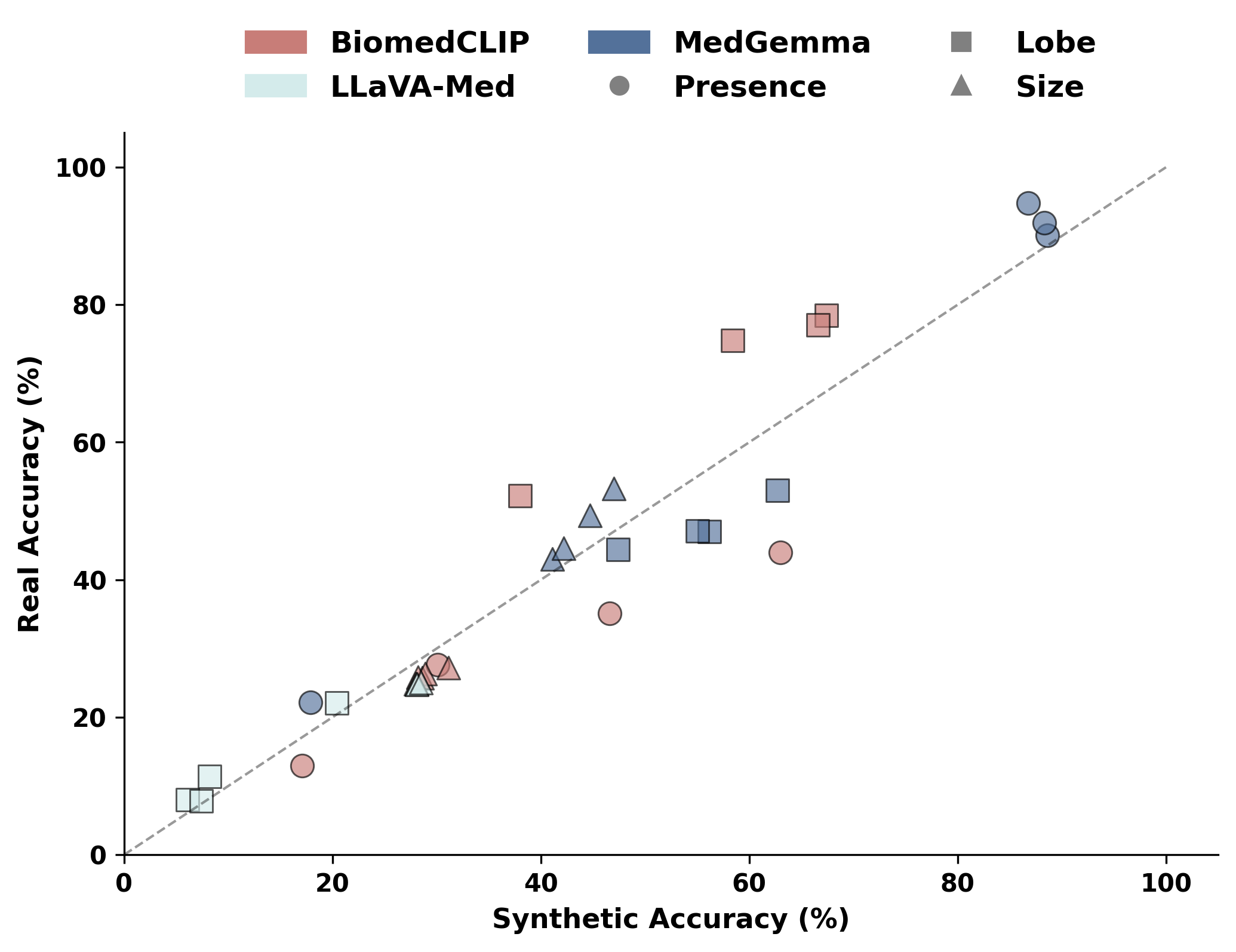}
  \caption{\textbf{Synthetic--real accuracy scatter across all 36 model$\times$task$\times$condition cells.}
  Each point is one cell; colour denotes task (presence / lobe / size).
  Spearman $\rho = 0.93$ ($p < 10^{-15}$). The dashed line is the identity.
  BiomedCLIP lobe points (circles) cluster above the identity; MedGemma lobe points cluster below it,
  illustrating the bidirectional, model-specific nature of the absolute offset.}
  \label{fig:synth_real_scatter}
\end{figure}

Table~\ref{tab:domain_gap_full} reports the complete real--synthetic accuracy comparison across all model$\times$task$\times$condition cells. Absolute offsets vary by model and task; because the real and synthetic cohorts are independent and differ in size, the table reports descriptive gaps rather than paired significance tests. Figure~\ref{fig:synth_real_scatter} visualises the 36-cell Spearman correlation ($\rho = 0.93$, $p < 10^{-15}$).

\begin{table}[htbp]
\caption{\textbf{Full synthetic--real accuracy gap across all model$\times$task$\times$condition cells.}
Real: $n{=}13{,}087$. Synthetic: $n{=}42{,}382$. $\Delta$: real minus synthetic (positive = real higher).
Differences are reported descriptively because the two cohorts are independent and unequal in size.}
\label{tab:domain_gap_full}
\centering\small
\setlength{\tabcolsep}{4pt}
\begin{tabular}{@{}llcccc@{}}
\toprule
\textbf{Model} & \textbf{Task} & \textbf{Condition}
  & \textbf{Real (\%)} & \textbf{Synth (\%)} & \textbf{$\Delta$ (pp)} \\
\midrule
\multirow{12}{*}{BiomedCLIP}
  & \multirow{4}{*}{Presence}
    & Plain        & 13.0 & 17.1 & -4.1  \\
  & & BBox         & 35.1 & 46.6 & -11.5 \\
  & & Contour      & 44.0 & 63.0 & -19.0 \\
  & & BBox+Contour & 27.6 & 30.1 & -2.5  \\
\cmidrule{2-6}
  & \multirow{4}{*}{Lobe}
    & Plain        & 52.2 & 38.0 & +14.2 \\
  & & BBox         & 74.8 & 58.4 & +16.4 \\
  & & Contour      & 78.5 & 67.4 & +11.1 \\
  & & BBox+Contour & 77.1 & 66.6 & +10.5 \\
\cmidrule{2-6}
  & \multirow{4}{*}{Size}
    & Plain        & 25.7 & 28.6 & -2.9  \\
  & & BBox         & 26.3 & 29.0 & -2.7  \\
  & & Contour      & 25.8 & 28.2 & -2.4  \\
  & & BBox+Contour & 27.2 & 31.1 & -3.9  \\
\midrule
\multirow{12}{*}{LLaVA-Med}
  & \multirow{4}{*}{Presence$^\dagger$}
    & Plain        & 100.0 & 99.9 & +0.1 \\
  & & BBox         & 100.0 & 100.0& 0.0  \\
  & & Contour      & 100.0 & 100.0& 0.0  \\
  & & BBox+Contour & 100.0 & 100.0& 0.0  \\
\cmidrule{2-6}
  & \multirow{4}{*}{Lobe}
    & Plain        & 22.1 & 20.4 & +1.7 \\
  & & BBox         & 11.4 &  8.2 & +3.2 \\
  & & Contour      &  8.0 &  6.1 & +1.9 \\
  & & BBox+Contour &  7.8 &  7.4 & +0.4 \\
\cmidrule{2-6}
  & \multirow{4}{*}{Size}
    & Plain        & 24.9 & 28.0 & -3.1 \\
  & & BBox         & 24.9 & 28.1 & -3.2 \\
  & & Contour      & 24.8 & 28.1 & -3.3 \\
  & & BBox+Contour & 25.0 & 28.5 & -3.5 \\
\midrule
\multirow{12}{*}{MedGemma}
  & \multirow{4}{*}{Presence}
    & Plain        & 22.2 & 17.9 & +4.3 \\
  & & BBox         & 92.1 & 88.6 & +3.5 \\
  & & Contour      & 91.9 & 88.3 & +3.6 \\
  & & BBox+Contour & 94.8 & 86.8 & +8.0 \\
\cmidrule{2-6}
  & \multirow{4}{*}{Lobe}
    & Plain        & 44.4 & 47.4 & -3.0 \\
  & & BBox         & 53.0 & 62.7 & -9.7 \\
  & & Contour      & 47.0 & 56.2 & -9.2 \\
  & & BBox+Contour & 47.1 & 55.0 & -7.9 \\
\cmidrule{2-6}
  & \multirow{4}{*}{Size}
    & Plain        & 43.0 & 41.1 & +1.9 \\
  & & BBox         & 49.6 & 42.2 & +7.4 \\
  & & Contour      & 47.5 & 44.7 & +2.8 \\
  & & BBox+Contour & 53.3 & 47.0 & +6.3 \\
\bottomrule
\end{tabular}\\[4pt]
{\footnotesize $^\dagger$LLaVA-Med predicts \emph{present} for ${>}99.9\%$ of cases;
differences of 0.0--0.1\,pp reflect near-identical degenerate behaviour across domains.}
\end{table}


\subsection{Per-Dataset Real Accuracy Breakdown}
\label{app:per_dataset}

Table~\ref{tab:per_dataset} reports accuracy for BiomedCLIP and MedGemma separately for each of the seven real clinical datasets, averaged across all four guidance conditions. LLaVA-Med is omitted (degenerate presence; size and lobe near-chance in all datasets). The key dataset-level confound is NSCLCR, which consists exclusively of large nodules ($93.6\%$ diameter ${>}20$\,mm and $0\%$ ${<}5$\,mm), producing near-zero BiomedCLIP size accuracy ($6.7\%$) and near-zero MedGemma size accuracy ($1.6\%$) relative to the $25\%$ chance level, compared with $43$--$48\%$ on screening datasets with natural size distributions. This confirms that the real--synthetic gap in size accuracy is partly attributable to dataset-level size-prior mismatch rather than model capability, a confound that \sys{} removes by design through controlled size stratification (M2) and lobe stratification (M3).

\begin{table}[htbp]
\caption{\textbf{Per-dataset real accuracy (\%), averaged across four guidance conditions.}
Chance: 50\% presence, 20\% lobe, 25\% size.
BC = BiomedCLIP, MG = MedGemma.}
\label{tab:per_dataset}
\centering\small
\setlength{\tabcolsep}{4pt}
\begin{tabular}{@{}lrrrrrrrr@{}}
\toprule
\multirow{2}{*}{\textbf{Dataset}} & \multirow{2}{*}{\textbf{$n$}}
  & \multicolumn{3}{c}{\textbf{BiomedCLIP}}
  & \multicolumn{3}{c}{\textbf{MedGemma}} \\
\cmidrule(lr){3-5}\cmidrule(lr){6-8}
  & & Pres & Lobe & Size & Pres & Lobe & Size \\
\midrule
LUNA25   & 6{,}156 & 23.2 & 70.7 & 26.1 & 73.2 & 49.3 & 46.7 \\
DLCS24   & 2{,}473 & 24.1 & 71.1 & 23.4 & 72.4 & 49.2 & 38.8 \\
IMDCT    & 2{,}032 & 49.3 & 73.5 & 26.8 & 78.9 & 46.1 & 47.2 \\
LUNA16   & 1{,}179 & 43.2 & 73.0 & 28.6 & 79.9 & 47.0 & 46.1 \\
LNDbv4   &   743   & 26.9 & 68.9 & 26.1 & 74.7 & 44.0 & 43.2 \\
NSCLCR$^\dagger$ & 421 & 35.2 & 49.1 &  6.7 & 75.8 & 37.0 &  1.6 \\
LUNGx    &    83   & 48.5 & 79.2 & 35.6 & 84.6 & 49.4 & 37.6 \\
\midrule
\textbf{All real} & \textbf{13{,}087} & 30.2 & 70.7 & 26.2 & 74.8 & 47.9 & 47.6 \\
\bottomrule
\end{tabular}\\[4pt]
{\footnotesize $^\dagger$NSCLCR: $93.6\%$ nodules ${>}20$\,mm diameter, $0\%$ ${<}5$\,mm;
size-accuracy collapse reflects dataset prior rather than model failure.}
\end{table}

\subsection{Per-Size-Bin Accuracy and Mode Ablation Detail}
\label{app:mode_ablation}

\begin{figure}[htbp]
  \centering
  \includegraphics[width=0.82\linewidth]{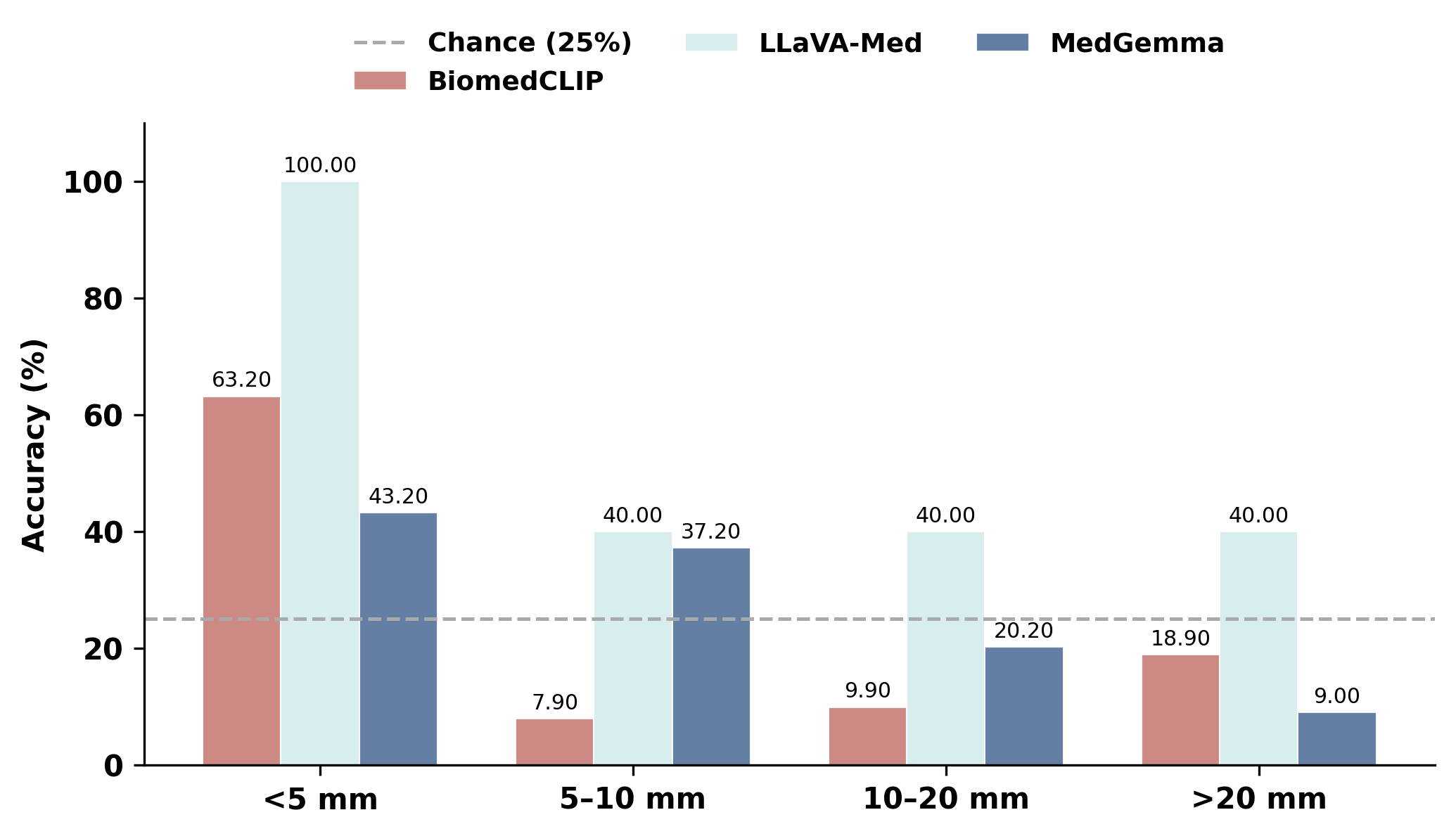}
  \caption{\textbf{Per-size-bin accuracy under the M2 (size-equalized) cohort.}
  BiomedCLIP's accuracy is concentrated in the ${<}5$\,mm bin (63.2\%) and falls well below chance for all larger nodule sizes, explaining its collapse to near-chance overall accuracy when bins are equalized. Chance level: 25\% (dashed).}
  \label{fig:size_bucket}
\end{figure}

Figure~\ref{fig:size_bucket} shows per-size-bin accuracy under plain condition on the M2 cohort, explaining the BiomedCLIP collapse observed in \S\ref{sec:mode_ablation}. BiomedCLIP achieves 63.2\% on nodules ${<}5$\,mm but falls to 7.9\%, 9.9\%, and 18.9\% on the three larger bins, indicating a systematic small-nodule bias. Under M2's equalized distribution, this bias yields near-chance overall accuracy. MedGemma shows a monotonically decreasing but shallower profile (43.2\% to 9.0\%), consistent with partial but not complete reliance on size priors.

\subsection{Per-Mode Performance Analysis}
\label{app:per_mode}

\begin{figure}[htbp]
  \centering
  \includegraphics[width=\linewidth]{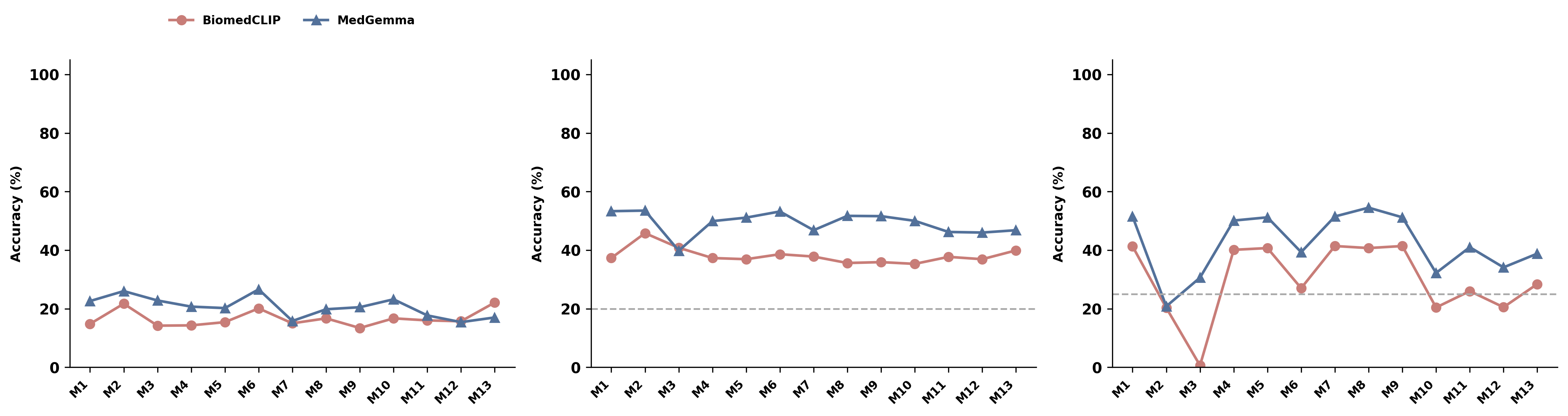}
  \caption{\textbf{Accuracy trajectories across M1--M13 for BiomedCLIP and MedGemma.}
  Left: presence. Centre: lobe. Right: size. Dashed line indicates chance level.
  Size accuracy (right) shows the sharpest mode-induced variation: BiomedCLIP collapses at M3 (lobe-equalized) and M12 (digital twin complete); MedGemma is more stable but drops across the digital twin family (M11--M13).}
  \label{fig:mode_difficulty}
\end{figure}

Table~\ref{tab:per_mode_plain} reports plain-condition accuracy for BiomedCLIP and MedGemma across all 13 trial modes. Figure~\ref{fig:mode_difficulty} visualises accuracy trajectories across modes for both models. LLaVA-Med is omitted from per-mode analysis (degenerate presence; size and lobe near-chance across all modes). Four mode families reveal distinct iTrialSpace-specific findings.

\begin{table}[htbp]
\caption{\textbf{Plain-condition accuracy (\%) across all 13 trial modes.}
BC = BiomedCLIP, MG = MedGemma. Chance: 50\% presence, 20\% lobe, 25\% size.}
\label{tab:per_mode_plain}
\centering\small
\setlength{\tabcolsep}{3.5pt}
\begin{tabular}{@{}clcccccc@{}}
\toprule
\multirow{2}{*}{\textbf{Mode}} & \multirow{2}{*}{\textbf{Description}}
  & \multicolumn{3}{c}{\textbf{BiomedCLIP}}
  & \multicolumn{3}{c}{\textbf{MedGemma}} \\
\cmidrule(lr){3-5}\cmidrule(lr){6-8}
  & & Pres & Lobe & Size & Pres & Lobe & Size \\
\midrule
M1  & NLST prior (baseline)        & 14.8 & 37.3 & 41.3 & 22.6 & 53.3 & 51.5 \\
M2  & Size-equalized               & 21.7 & 45.8 & 20.3 & 26.0 & 53.5 & 20.8 \\
M3  & Lobe-equalized               & 14.2 & 40.8 & 0.6  & 22.8 & 39.8 & 30.6 \\
M4  & Demographic-equalized        & 14.3 & 37.3 & 40.1 & 20.7 & 49.9 & 50.1 \\
M5  & Counterfactual prevalence    & 15.4 & 36.9 & 40.7 & 20.2 & 51.1 & 51.2 \\
M6  & Cross-dataset insertion      & 20.1 & 38.6 & 27.0 & 26.6 & 53.2 & 39.3 \\
M7  & Bootstrap resample (×20)     & 15.0 & 37.8 & 41.4 & 15.8 & 46.8 & 51.5 \\
M8  & Fixed-manifest (shared exclusion) & 16.7 & 35.6 & 40.7 & 19.8 & 51.7 & 54.5 \\
M9  & Prevalence decay             & 13.4 & 35.9 & 41.4 & 20.5 & 51.6 & 51.2 \\
M10 & Multi-nodule context         & 16.7 & 35.3 & 20.4 & 23.2 & 50.0 & 32.2 \\
\cmidrule{1-8}
M11 & Digital twin -- isolate      & 16.0 & 37.7 & 26.0 & 17.7 & 46.2 & 41.0 \\
M12 & Digital twin -- complete     & 15.7 & 36.9 & 20.5 & 15.4 & 46.0 & 34.1 \\
M13 & Digital twin -- cross        & 22.1 & 39.9 & 28.4 & 17.0 & 46.8 & 38.8 \\
\bottomrule
\end{tabular}
\end{table}

\paragraph{Baseline stability (M1, M4, M5, M7, M8, M9).}
Six modes spanning demographic stratification (M4), counterfactual prevalence sweeps (M5), bootstrap resampling (M7), fixed-manifest replication (M8), and prevalence decay (M9) produce virtually identical size accuracy: BiomedCLIP ranges 40.1--41.4\% and MedGemma 50.1--54.5\% across all six (plain condition). Lobe accuracy is equally stable (BiomedCLIP 35.6--37.8\%; MedGemma 46.8--53.3\%). This consistency confirms that \sys{} produces reproducible evaluation cohorts: accuracy differences across modes reflect genuine distributional interventions, not pipeline stochasticity or sampling noise.

\paragraph{Distribution interventions (M2, M3).}
Removing the size prior (M2) collapses both models to near chance (BiomedCLIP 20.3\%; MedGemma 20.8\%). Removing the lobe--size correlation (M3) additionally collapses BiomedCLIP size to 0.6\% while MedGemma retains 30.6\%, as analysed in \S\ref{sec:mode_ablation}. These two modes isolate which accuracy components are prior-driven versus architecturally grounded, a decomposition unachievable on fixed-distribution real data.

\begin{figure}[htbp]
  \centering
  \includegraphics[width=0.82\linewidth]{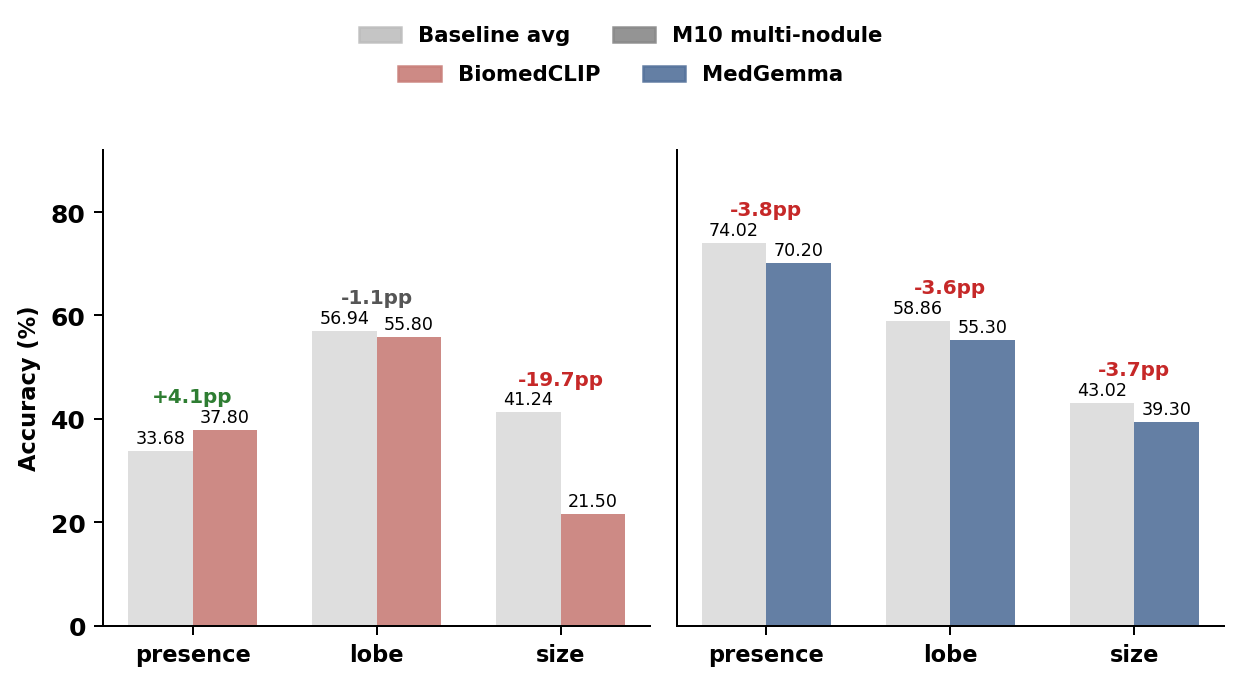}
  \caption{\textbf{Multi-nodule interference (M10 vs baseline average).}
  BiomedCLIP size accuracy drops \dn{19.7}\pp{} under multi-nodule context while lobe remains flat;
  MedGemma shows modest drops across all tasks (\dn{3.5} -- \dn{3.9}\pp{}).
  BiomedCLIP presence rises \up{4.1}\pp{}, suggesting additional nodule mass inflates positive-prediction confidence independently of correctness.}
  \label{fig:multinodule}
\end{figure}

\paragraph{Context generalisation (M6, M10).}
Cross-dataset insertion (M6) restricts each donor nodule to a specific source dataset while hosts are drawn from the complementary datasets. BiomedCLIP size accuracy drops to 27.0\% (\dn{$-$14.3}\pp{} vs M1) and MedGemma to 39.3\% (\dn{$-$12.2}\pp{}), while lobe accuracy remains stable for both (BiomedCLIP 38.6\%; MedGemma 53.2\%) — indicating the cross-dataset gap is size-specific, not global. Multi-nodule context (M10) creates a different pattern: BiomedCLIP size accuracy drops to 20.4\% (\dn{$-$19.7}\pp{} condition-averaged vs baseline; Fig.~\ref{fig:multinodule}), while MedGemma drops only \dn{3.9}\pp{}. Presence accuracy rises slightly for BiomedCLIP under M10 (\up{4.1}\pp{}), suggesting that additional nodule mass in the image inflates positive-prediction confidence independently of correctness.

\paragraph{Digital twin family (M11--M13).}
All three digital twin modes produce lower size accuracy than insertion modes: BiomedCLIP 20.5--28.4\% and MedGemma 34.1--41.0\%, with M12 (all nodules re-inserted simultaneously into their native scan) being the hardest for both models (BiomedCLIP 20.5\%; MedGemma 34.1\%). Lobe accuracy is more stable across the digital twin family (BiomedCLIP 36.9--39.9\%; MedGemma 46.0--46.8\%), indicating that lobe-localisation is less sensitive to scene complexity than size classification. The M13 host--donor decomposition is reported in \S\ref{sec:m13}.

\subsection{Per-Lobe Accuracy Analysis}
\label{app:per_lobe}

\begin{figure}[htbp]
  \centering
  \includegraphics[width=0.52\linewidth]{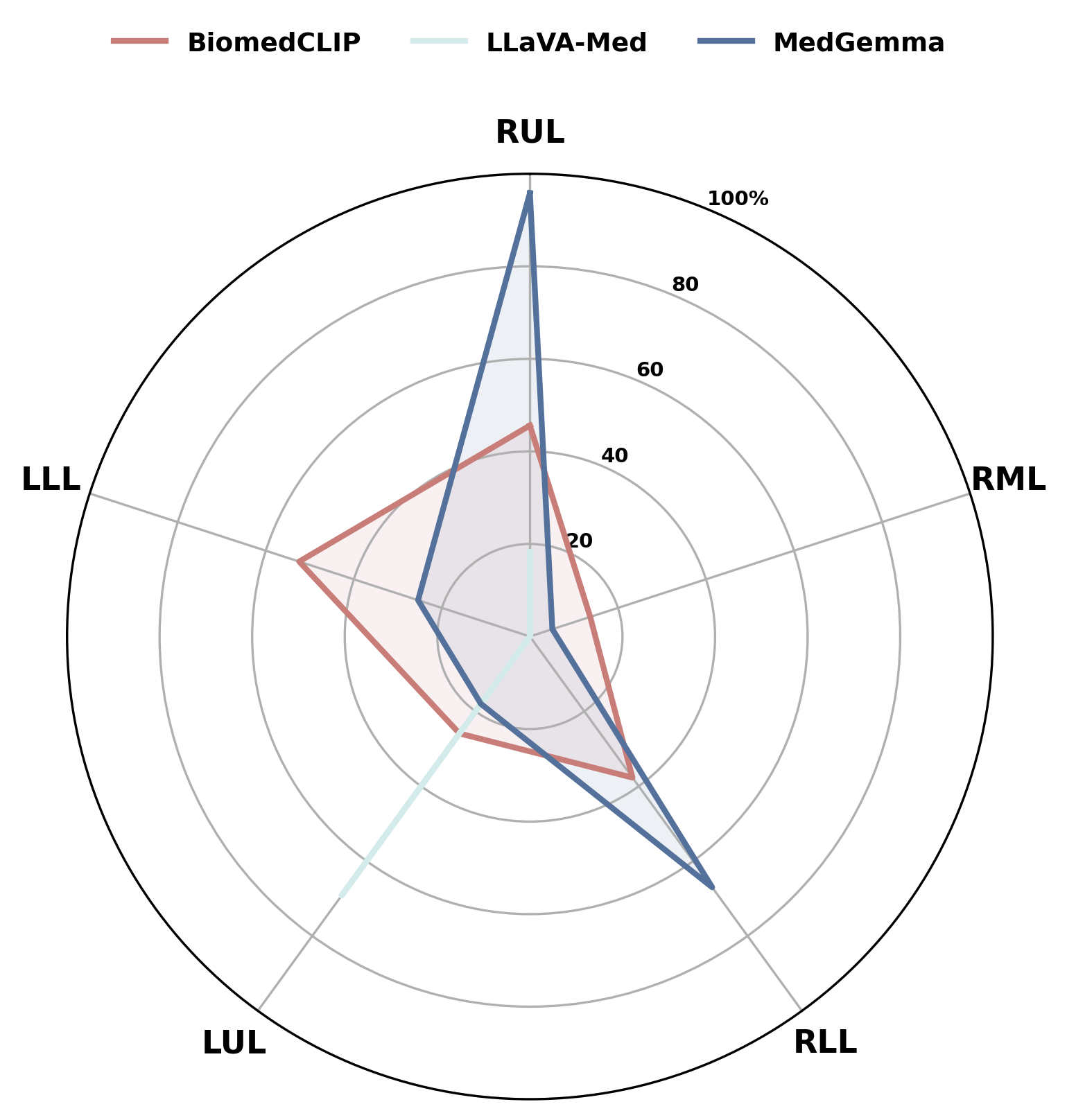}
  \caption{\textbf{Per-lobe accuracy under contour guidance (synthetic cohort).}
  MedGemma achieves $\sim$90\% on RUL but 25--45\% on remaining lobes.
  BiomedCLIP shows a qualitatively similar but shallower asymmetry.
  LLaVA-Med is near-uniform chance across all lobes.
  RML (right middle lobe) is the hardest class for all models.}
  \label{fig:lobe_radar}
\end{figure}

Figure~\ref{fig:lobe_radar} shows per-lobe accuracy under contour guidance (best condition for lobe localisation) averaged across synthetic modes.
MedGemma achieves $\sim$90\% on the right upper lobe (RUL) but only 25--45\% on remaining lobes, producing a strongly asymmetric profile.
BiomedCLIP shows a shallower but qualitatively similar asymmetry (RUL highest; RML lowest).
The RUL spike is consistent with anatomical distinctiveness: the right upper lobe occupies a visually unambiguous apical position in axial slices, whereas the right middle lobe (RML, 4--9\% of the cohort) is the smallest and most variable in extent.
LLaVA-Med remains near uniform chance ($\sim$20\%) across all five lobes, consistent with its instruction-prior-dominated behaviour.
This per-lobe decomposition is only tractable within \sys{} because M3's lobe-equalized cohort ensures each lobe is represented with equal prevalence; standard benchmarks with naturalistic lobe distributions would confound per-lobe accuracy with prevalence effects.

\subsection{Performance Analysis Under Spatial Guidance}
\label{app:guidance}

\begin{figure}[htbp]
  \centering
  \includegraphics[width=\linewidth]{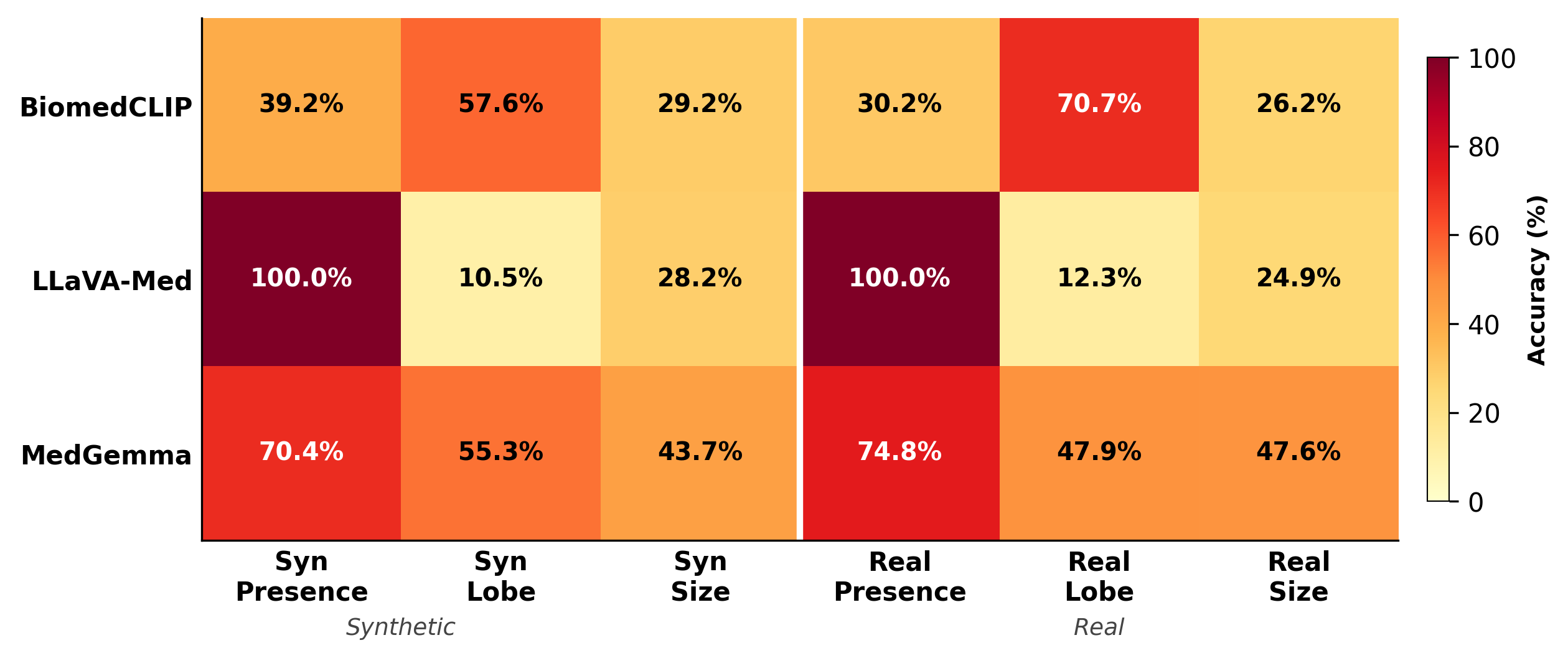}
  \caption{\textbf{Condition-averaged accuracy heatmap across models, tasks, and domains.}
  Values are mean accuracy across all four guidance conditions.
  LLaVA-Med's 100\% presence reflects degenerate present-always behaviour.
  MedGemma leads on presence and size; BiomedCLIP leads on lobe (real data).}
  \label{fig:heatmap_overview}
\end{figure}

Table~\ref{tab:guidance_full} reports accuracy under all four guidance conditions across all model$\times$task$\times$split combinations, with key pairwise guidance-lift statistics from McNemar tests (Bonferroni-corrected). Figure~\ref{fig:heatmap_overview} provides a condition-averaged overview. Three findings extend the main-paper summary in \S\ref{sec:vlm_eval}.

\begin{table}[htbp]
\caption{\textbf{Accuracy (\%) under all four guidance conditions, real and synthetic.}
$\Delta_{\mathrm{best}}$: best guided minus plain. Statistical test: McNemar (Bonferroni-corrected) on the plain$\to$best transition.
B = BBox, C = Contour, B+C = BBox+Contour. $^\dagger$Degenerate: ${>}99.9\%$ present-always.}
\label{tab:guidance_full}
\centering\small
\setlength{\tabcolsep}{3pt}
\begin{tabular}{@{}llccccccc@{}}
\toprule
\textbf{Split} & \textbf{Model} & \textbf{Task}
  & \textbf{Plain} & \textbf{BBox} & \textbf{Contour} & \textbf{B+C}
  & \textbf{Best} & \textbf{$\Delta_{\mathrm{best}}$} \\
\midrule
\multirow{9}{*}{Real}
& \multirow{3}{*}{BiomedCLIP}
  & Presence & 13.0 & 35.1 & 44.0 & 27.6 & C   & \up{$+$31.0}$^{***}$ \\
&& Lobe     & 52.2 & 74.8 & 78.5 & 77.1 & C   & \up{$+$26.3}$^{***}$ \\
&& Size     & 25.7 & 26.3 & 25.8 & 27.2 & B+C & \up{$+$1.5}$^{***}$  \\
\cmidrule{2-9}
& \multirow{3}{*}{LLaVA-Med}
  & Presence & 100.0$^\dagger$ & 100.0$^\dagger$ & 100.0$^\dagger$ & 100.0$^\dagger$ & -- & 0.0\,\emph{ns} \\
&& Lobe     & 22.1 & 11.4 & 8.0  & 7.8  & B   & \dn{$-$10.7}$^{***}$ \\
&& Size     & 24.9 & 24.9 & 24.8 & 25.0 & B+C   & \up{$+$0.1}\,\emph{ns} \\
\cmidrule{2-9}
& \multirow{3}{*}{MedGemma}
  & Presence & 22.2 & 92.1 & 91.9 & 94.8 & B+C & \up{$+$72.6}$^{***}$ \\
&& Lobe     & 44.4 & 53.0 & 47.0 & 47.1 & B   & \up{$+$8.6}$^{***}$  \\
&& Size     & 43.0 & 49.6 & 47.5 & 53.3 & B+C & \up{$+$10.3}$^{***}$ \\
\midrule
\multirow{9}{*}{Synthetic}
& \multirow{3}{*}{BiomedCLIP}
  & Presence & 17.1 & 46.6 & 63.0 & 30.1 & C   & \up{$+$45.9}$^{***}$ \\
&& Lobe     & 38.0 & 58.4 & 67.4 & 66.6 & C   & \up{$+$29.4}$^{***}$ \\
&& Size     & 28.6 & 29.0 & 28.2 & 31.1 & B+C & \up{$+$2.5}$^{***}$  \\
\cmidrule{2-9}
& \multirow{3}{*}{LLaVA-Med}
  & Presence & 99.9$^\dagger$ & 100.0$^\dagger$ & 100.0$^\dagger$ & 100.0$^\dagger$ & -- & 0.0\,\emph{ns} \\
&& Lobe     & 20.4 & 8.2  & 6.1  & 7.4  & B   & \dn{$-$12.2}$^{***}$ \\
&& Size     & 28.0 & 28.1 & 28.1 & 28.5 & B+C   & \up{$+$0.5}\,\emph{ns} \\
\cmidrule{2-9}
& \multirow{3}{*}{MedGemma}
  & Presence & 17.9 & 88.6 & 88.3 & 86.8 & B   & \up{$+$70.7}$^{***}$ \\
&& Lobe     & 47.4 & 62.7 & 56.2 & 55.0 & B   & \up{$+$15.3}$^{***}$ \\
&& Size     & 41.1 & 42.2 & 44.7 & 47.0 & B+C & \up{$+$5.9}$^{***}$  \\
\bottomrule
\end{tabular}\\[4pt]
{\footnotesize $^{***}$ McNemar test, Bonferroni-corrected, $p<0.001$.
MDE: $\pm$1.2\,pp (real), $\pm$0.7\,pp (synthetic).}
\end{table}

\paragraph{BBox+Contour paradox in BiomedCLIP.}
For BiomedCLIP presence, adding a bounding box on top of contour guidance \emph{reduces} accuracy: contour alone achieves 44.0\% (real) and 63.0\% (synthetic), whereas BBox+Contour drops to 27.6\% (\dn{$-$16.4}\pp{}) and 30.1\% (\dn{$-$32.9}\pp{}) respectively. The effect is absent for lobe localisation, where BBox+Contour and contour-only are within 1.4\,pp on both splits. We attribute this to BiomedCLIP's contrastive alignment mechanism: the contour provides a spatially clean boundary that the image encoder matches to the ``nodule'' concept, while the bounding box introduces a second competing visual region that disrupts the patch-level alignment. This degradation is reproducible across both domains and is detectable only by testing all four conditions systematically — a design choice that flat comparison benchmarks do not make.

\paragraph{LLaVA-Med: guidance harms lobe localisation.}
LLaVA-Med lobe accuracy degrades sharply under every guided condition. On real data the sequence is plain (22.1\%) $\to$ BBox (11.4\%) $\to$ Contour (8.0\%) $\to$ BBox+Contour (7.8\%); on synthetic data all guided variants remain below plain (BBox 8.2\%, Contour 6.1\%, BBox+Contour 7.4\% vs.\ plain 20.4\%). Additional visual cues therefore fail to improve lobe prediction and usually make it worse. This is consistent with instruction-tuning priors overriding visual evidence: the model produces lobe predictions from language pattern matching, and overlaid annotations that draw attention to a specific region conflict with the prior rather than reinforcing it. Size predictions show complete insensitivity across all four conditions on both splits (maximum shift: \up{0.5}\pp{} synthetic, \emph{ns}), confirming that LLaVA-Med's size output is decoupled from image content entirely.

\paragraph{Size classification is guidance-resistant across architectures.}
Despite large gains for presence and lobe, size classification responds weakly to all guidance types. BiomedCLIP's maximum size lift is \up{2.5}\pp{} (BBox+Contour, synthetic; statistically significant but clinically negligible). MedGemma's \up{10.3}\pp{} gain (BBox+Contour, real) is the largest observed, but peak accuracy remains 53.3\% --- 28.3\,pp above raw 4-class chance and 10.3\,pp above MedGemma's plain-condition baseline. The contour, which outlines the exact nodule boundary and therefore encodes size implicitly, fails to unlock size perception in BiomedCLIP (contour size: 25.8\% real, 28.2\% synthetic --- indistinguishable from plain). This confirms that size classification requires architectural capacity beyond boundary overlay processing, and that current VLM evaluation on this task is largely measuring distributional priors rather than genuine size perception.

\begin{figure}[htbp]
  \centering
  \includegraphics[width=0.9\linewidth]{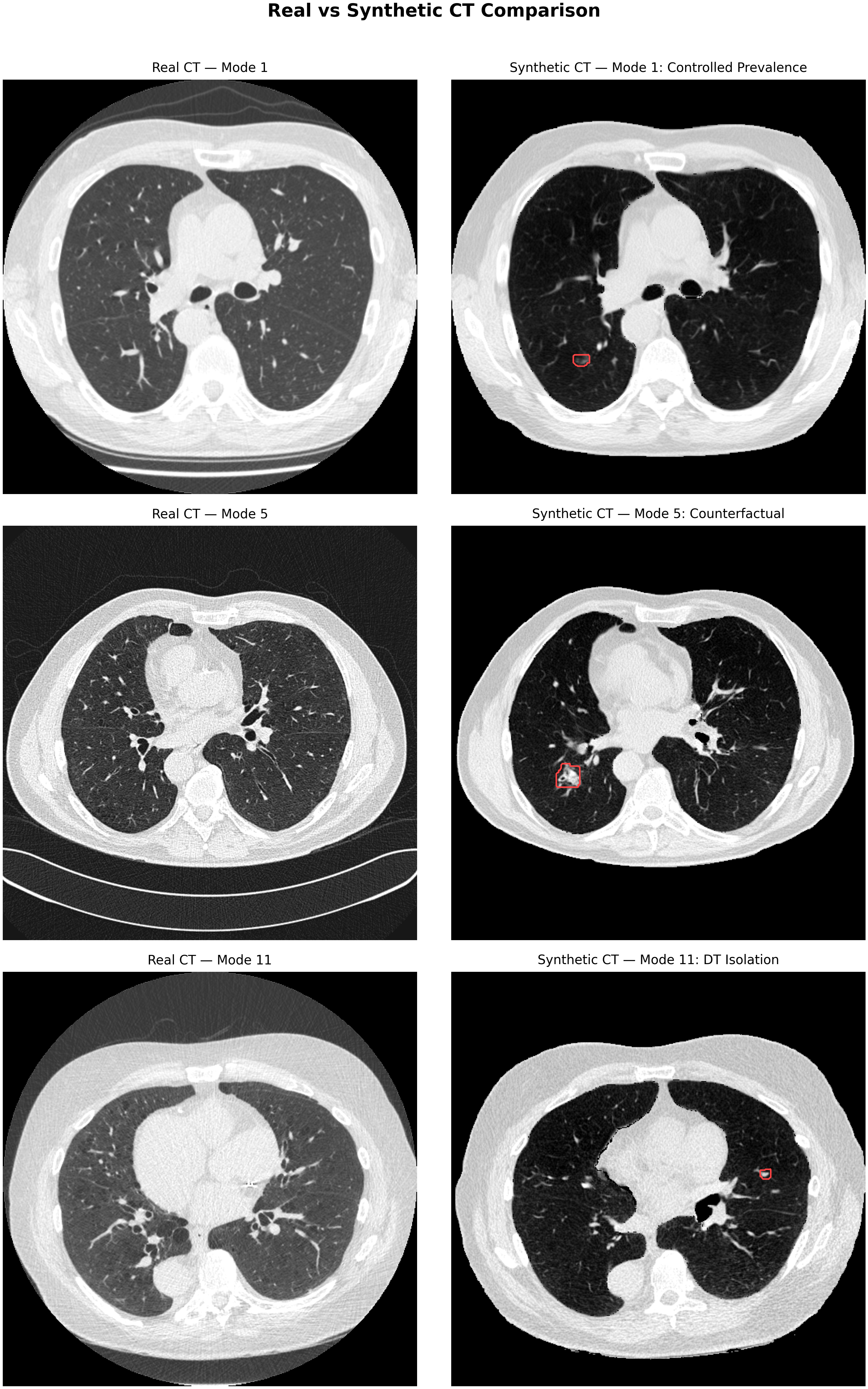}
  \caption{\textbf{Real vs.\ synthetic CTs.} Side-by-side comparisons for three trial modes. Synthetic CTs exhibit realistic parenchymal texture, preserved vascular anatomy, and seamless nodule integration at lung window ($-$600/1500\,HU).}
  \label{fig:app_visual}
\end{figure}

\begin{figure}[htbp]
  \centering
  \includegraphics[width=\linewidth]{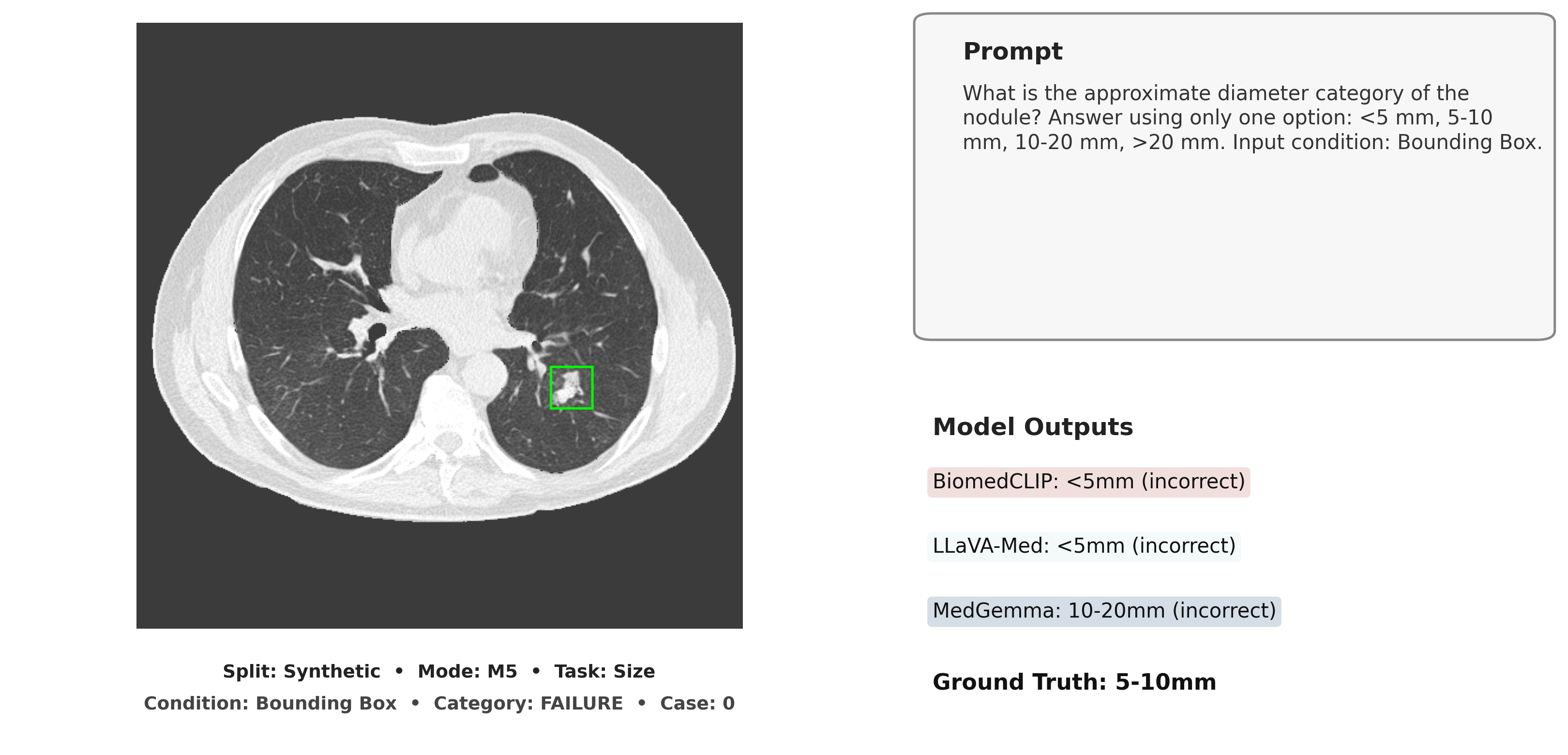}
  \caption{\textbf{Representative failure case for synthetic size classification under bounding-box guidance.} The ground-truth category is \textbf{5--10 mm}, but \textbf{BiomedCLIP} and \textbf{LLaVA-Med} predict \textbf{$<$5 mm}, while \textbf{MedGemma} predicts \textbf{10--20 mm}.}
  \label{fig:val_eval1}
\end{figure}

\begin{figure}[htbp]
  \centering
  \includegraphics[width=\linewidth]{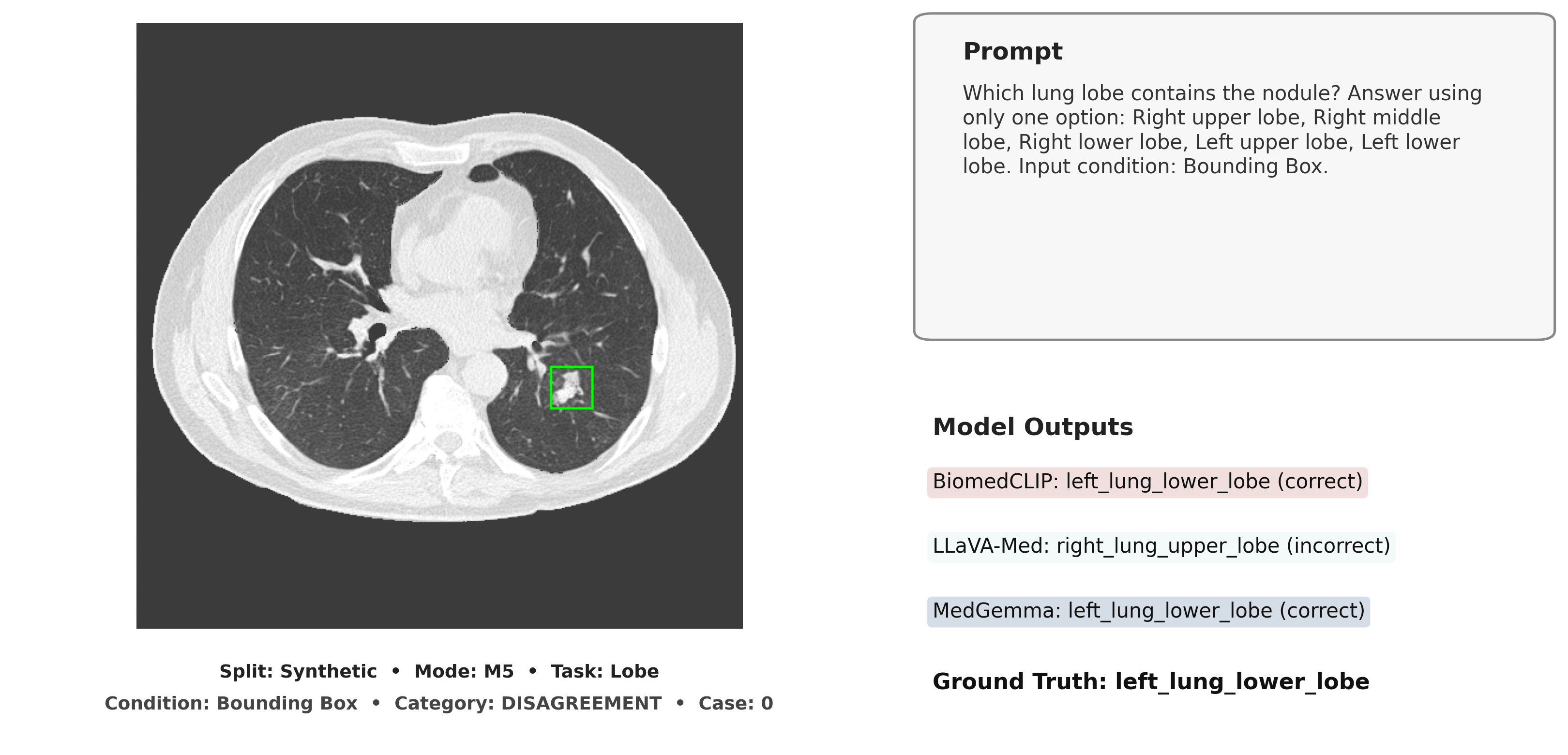}
  \caption{\textbf{Representative disagreement case for synthetic lobe localization under bounding-box guidance.} The ground-truth lobe is \textbf{left\_lung\_lower\_lobe}. \textbf{BiomedCLIP} and \textbf{MedGemma} predict the correct lobe, whereas \textbf{LLaVA-Med} incorrectly predicts \textbf{right\_lung\_upper\_lobe}.}
  \label{fig:val_eval2}
\end{figure}

\begin{figure}[htbp]
  \centering
  \includegraphics[width=\linewidth]{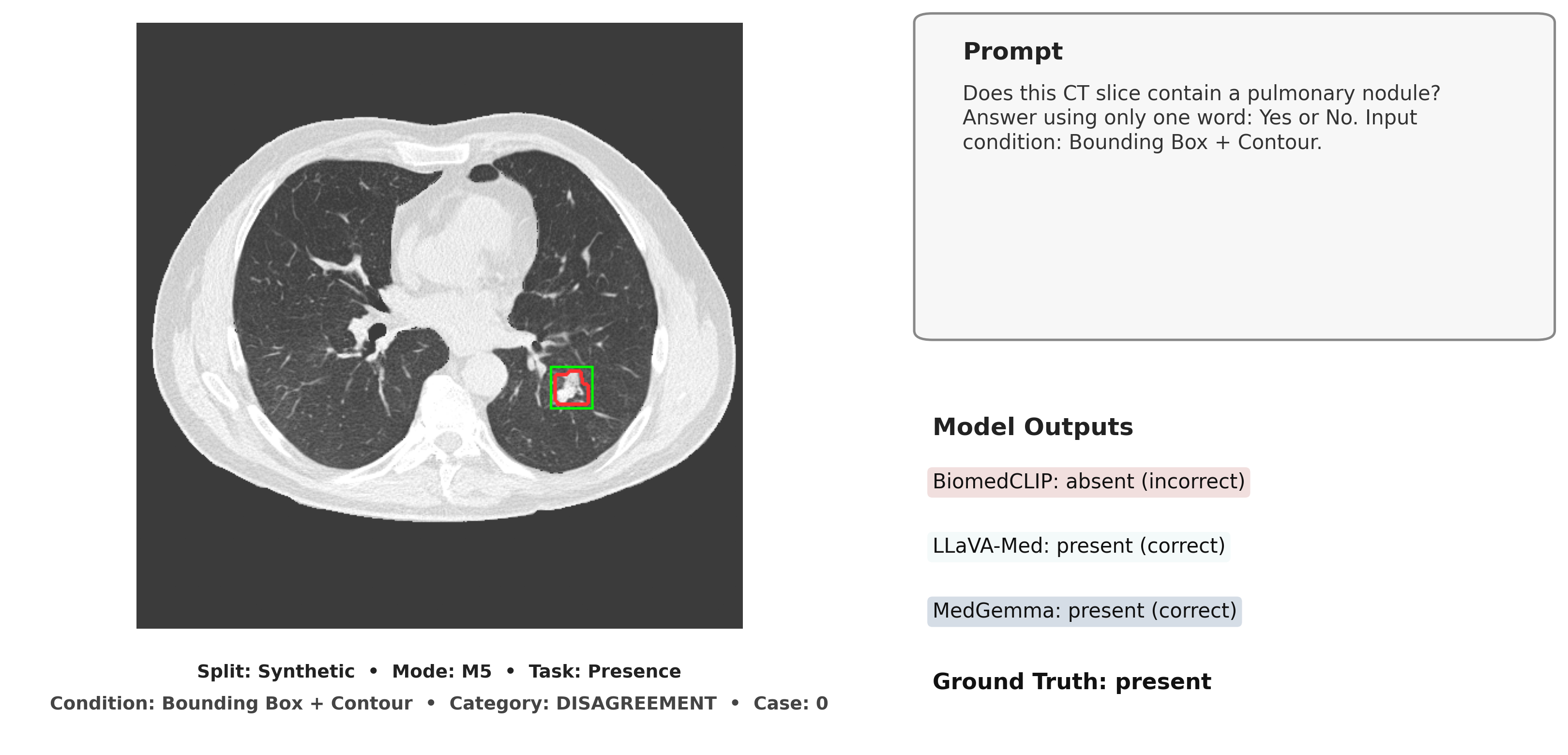}
  \caption{\textbf{Representative disagreement case for synthetic nodule presence detection under bounding-box + contour guidance.} The ground truth is \textbf{present}. \textbf{LLaVA-Med} and \textbf{MedGemma} correctly detect the nodule, whereas \textbf{BiomedCLIP} incorrectly predicts \textbf{absent}.}
  \label{fig:val_eval3}
\end{figure}

\begin{figure}[htbp]
  \centering
  \includegraphics[width=\linewidth]{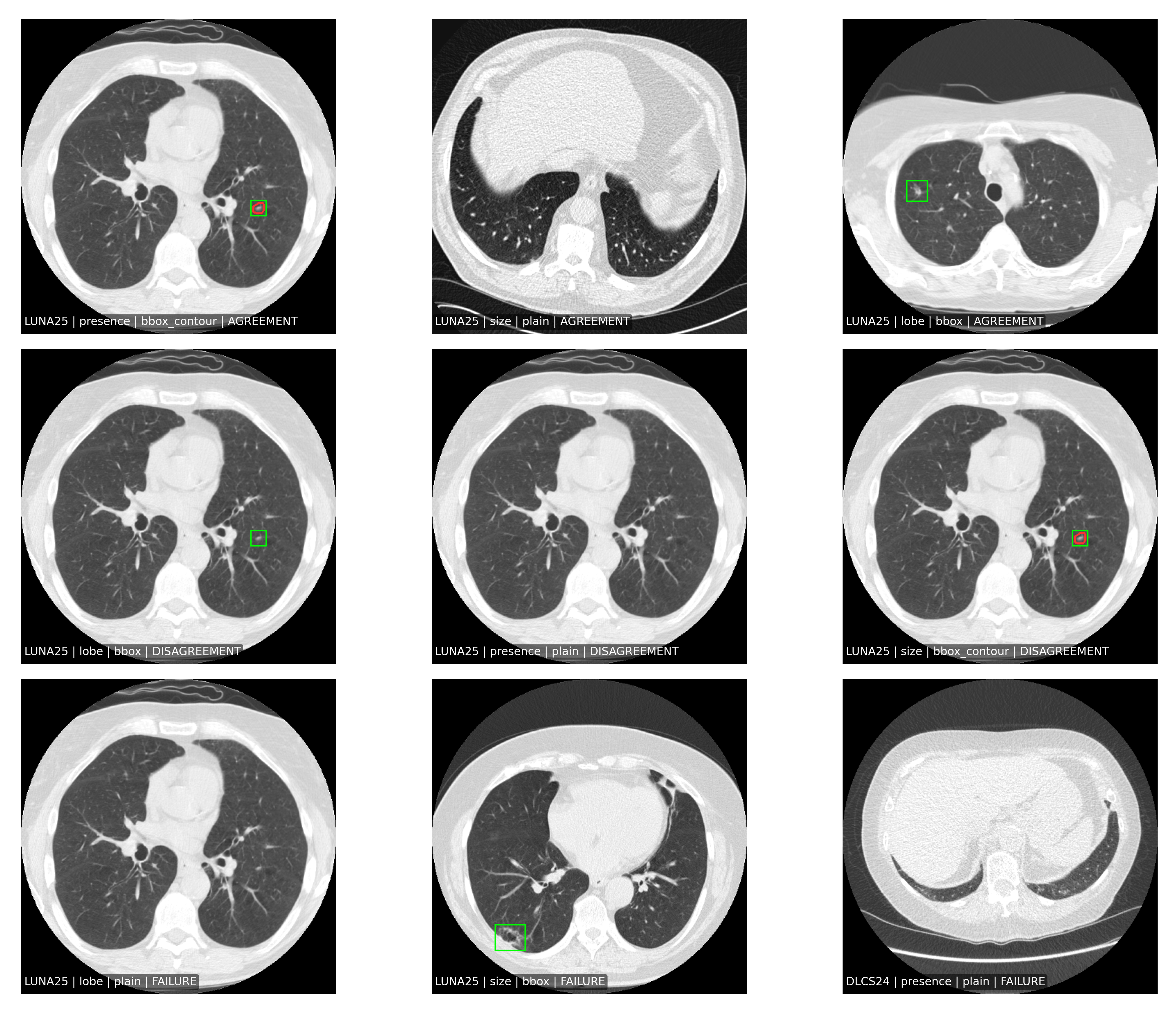}
  \caption{\textbf{A mixed panel of real clinical examples across agreement, disagreement, and failure outcomes.} The panel demonstrates where guidance conditions preserve consensus correctness and where model divergence appears under clinically diverse tasks.}
  \label{fig:val_eval4}
\end{figure}

\begin{figure}[htbp]
  \centering
  \includegraphics[width=\linewidth]{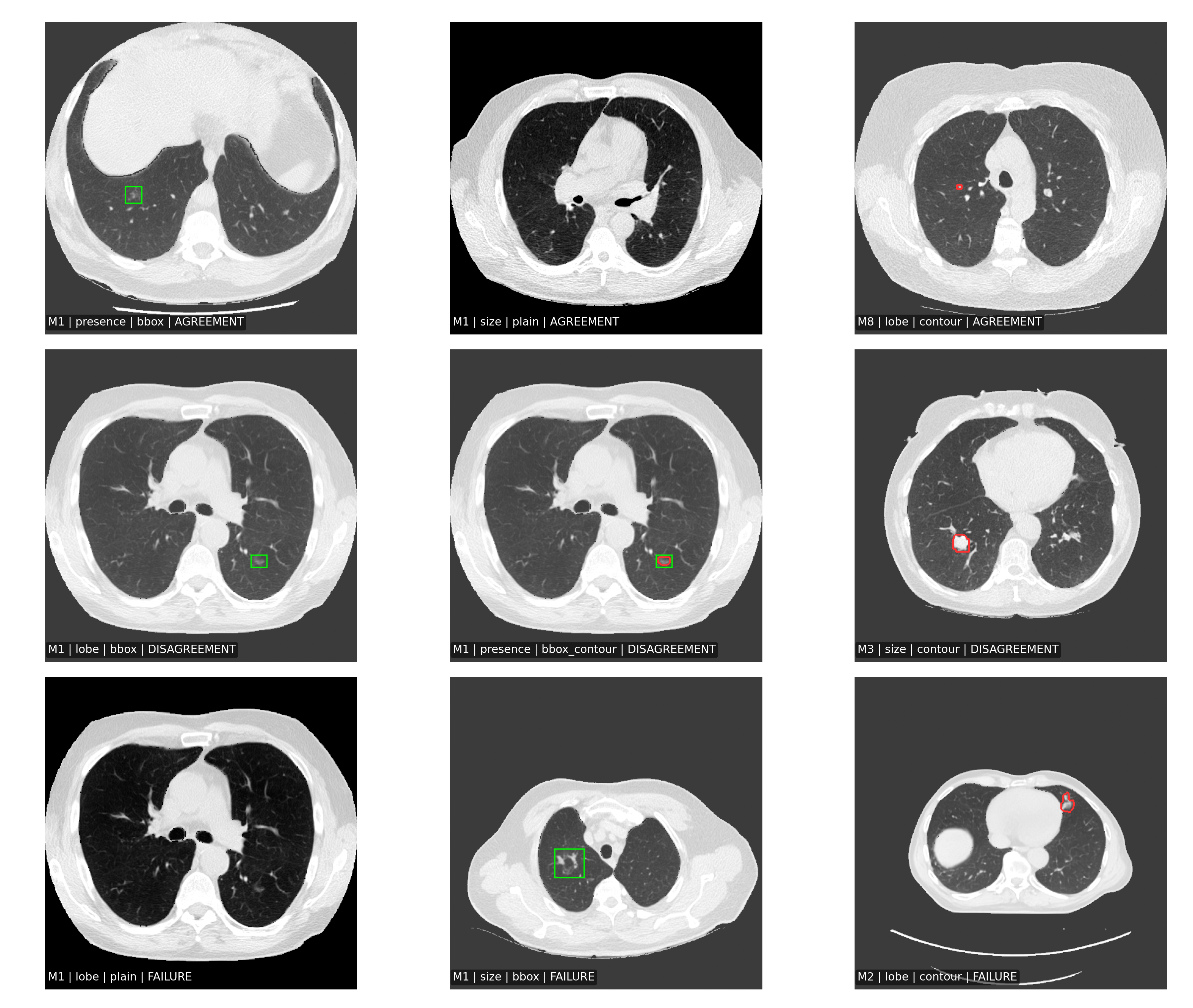}
  \caption{Representative synthetic examples across M1--M13 showing mode-dependent behavior under the same prompting protocol. This figure emphasizes controlled stress-testing of model robustness using trial-space variation.}
  \label{fig:val_eval5}
\end{figure}

\end{document}